\documentclass[10pt,twocolumn,letterpaper]{article}

\usepackage{iccv}
\usepackage{times}
\usepackage{epsfig}
\usepackage{graphicx}
\usepackage{amsmath}
\usepackage{amssymb}
\usepackage{booktabs,subfigure,color,threeparttable}
\graphicspath{{./Figure/}}
\usepackage{tabularx}
\usepackage{multirow}
\usepackage{tabulary}

\usepackage[breaklinks=true,bookmarks=false]{hyperref}

\iccvfinalcopy % *** Uncomment this line for the final submission

\def\iccvPaperID{2291} % *** Enter the ICCV Paper ID here

\ificcvfinal\fi

\begin{document}

%%%%%%%%% TITLE
\title{Cross-X Learning for Fine-Grained Visual Categorization}

\author{Wei Luo$^{1,2}$\quad Xitong Yang$^2$\quad Xianjie Mo$^1$\quad Yuheng Lu$^{2,5}$\quad Larry S. Davis$^2$\\ Jun Li$^3$\quad Jian Yang$^4$\quad Ser-Nam Lim$^5$\\
$^1$South China Agricultural University\quad $^2$University of Maryland, College Park\\
$^3$MIT\quad $^4$Nanjing University of Science and Technology\quad $^5$Facebook AI\\
{\tt\small \{cswluo,yangxitongbob,cedricmo.cs,junl.mldl,sernam\}@gmail.com}\\\
{\tt\small \{ylu,lsd\}@umiacs.umd.edu}\quad
{\tt\small csjyang@njust.edu.cn}
}

\maketitle
\ificcvfinal\thispagestyle{empty}\fi

%%%%%%%%% ABSTRACT
\begin{abstract}
Recognizing objects from subcategories with very subtle differences remains a challenging task due to the large intra-class and small inter-class variation.
Recent work tackles this problem in a weakly-supervised manner: object parts are first detected and the corresponding part-specific features are extracted for fine-grained classification.
However, these methods typically treat the part-specific features of each image in isolation while neglecting their relationships between different images.
In this paper, we propose Cross-X learning, a simple yet effective approach that exploits the relationships between different images and between different network layers for robust multi-scale feature learning.
Our approach involves two novel components: (i) a cross-category cross-semantic regularizer that guides the extracted features to represent semantic parts and, (ii) a cross-layer regularizer that improves the robustness of multi-scale features by matching the prediction distribution across multiple layers.
Our approach can be easily trained end-to-end and is scalable to large datasets like NABirds.
We empirically analyze the contributions of different components of our approach and  demonstrate  its  robustness, effectiveness  and state-of-the-art performance on five benchmark datasets.
Code is available at \url{https://github.com/cswluo/CrossX}.
\end{abstract}

%%%%%%%%% BODY TEXT
\section{Introduction}
\label{sec:intro}
Fine-grained visual categorization (FGVC) aims at classifying objects from very similar categories, \eg subcategories of birds~\cite{cubbirds11caltech,nabirds15Perona}, dogs~\cite{stdogs11feifei} and cars~\cite{stcars13feifei}. It has long been considered as a challenging task due to the large intra-class and small inter-class variation, as well as the deficiency of annotated data. Benefiting from the progress of deep neural networks~\cite{alexnet12hiton,vggnet15zisserman,googlenet15Szegedy,resnet16kaiming}, the recognition performance of FGVC has improved steadily in recent years and the community has more recently focused on weakly-supervised FGVC that obviates the need of labor-intensive part-based annotation. There are two main approaches to weakly-supervised FGVC, namely, exploiting relationships between fine-grained labels to regularize feature learning~\cite{multigranularity@iccv,hyperlabels@cvpr} and localizing discriminative parts for part-specific feature extraction~\cite{racnn@mei,ntscnn@eccv}. Compared to label-relationship based methods, the localization-based methods have the advantages of extracting fine-grained features from local regions where subtle differences between subcategories usually exist. %Our work falls into this line of research. 

Early work on localization-based methods typically adopts a multi-stage learning framework: part detectors are first obtained by training on DCNN features~\cite{augstronglysup@cvpr} or exploiting the hidden representations in DCNNs~\cite{twolevelattention@cvpr,neuralconstellations@iccv,pickfilter@tianqi}, and then used to extract part-specific features for fine-grained classification. 
More recent work merges these two stages into an end-to-end learning framework that utilizes the final objective to optimize both part localization and fine-grained classification at the same time ~\cite{racnn@mei,macnn@mei,ntscnn@eccv,dfbnet18larry}.
These methods localize semantic parts independently on each image while neglecting the relationships between the part-specific features from different images. \cite{mamc18eccv} explores the relationships between object parts by proposing a soft attention-based model. The model first generates attention region features of each input image via multiple excitation modules and then guides the attention features to have semantic meaning by adopting a metric learning framework.
However, the improvement from their model is limited as optimizing such a metric learning loss is challenging and involves a non-trivial sample selection procedure~\cite{wu2017sampling}.

We propose Cross-X learning, a simple but effective approach that leverages the relationships between different images and between different network layers for robust fine-grained recognition.
Similar to~\cite{mamc18eccv}, our approach first generates attention region features via multiple excitation modules, but it further involves two novel components: a \textit{cross}-category \textit{cross}-semantic regularizer ($C^3S$) and a \textit{cross}-layer regularizer ($CL$).
$C^3S$ is introduced to guide the attention features from different excitation modules to represent different semantic parts.
Ideally, the attention features for the same semantic parts, although coming from different images with different class labels, should be more correlated than those for different semantic parts (see Fig.~\ref{fig:c3s}).
Therefore, $C^3S$ regulates the feature learning by maximizing the correlation between attention features extracted by the same excitation module while decorrelating those extracted by different excitation modules.
Compared to the metric learning loss, $C^3S$ can be naturally integrated into the model and easily optimized without any sampling procedure.
Meanwhile, we exploit the relationships between different network layers for robust multi-scale feature learning. We first adapt FPN~\cite{fpn17kaiming} to generate merged features. The merged features enable our model to discover local discriminative structures with both fine spatial resolution and rich high-level semantic information. To further improve the robustness of the multi-scale features, we introduce a cross-layer regularizer ($CL$) that matches the prediction distribution of the mid-level features to that of the high-level features by minimizing their KL-divergence.
Experimental results on five benchmark datasets show that our approach outperforms or achieves comparable performance to the state-of-the-art methods. Moreover, our approach is easy to train and is scalable to large-scale datasets as it does not involve multi-stage or multi-crop mechanisms.
We make the following contributions:
\begin{itemize}
\item We propose a Cross-X learning approach for fine-grained feature learning. Cross-X learning explores relationships between features from different images and different network layers to learn semantic part features.
\item We address the issue of robust multi-scale feature learning through cross-layer regularization, which matches prediction distributions across layers, thus increasing the robustness of features in different layers.
% \item We conduct comprehensive experiments on five fine-grained benchmark datasets including one large-scale dataset with 555 categories and provide a detailed ablation study to clarify the role of every component of our approach. 
\end{itemize}
The remainder of the paper is organized as follows: Section~\ref{sec:rlwk} briefly reviews related work to our approach. Our approach is studied and detailed in Section~\ref{sec:models}. The model ablation studies and experimental results are analyzed and presented in Section~\ref{sec:exp}. We conclude our work in Section~\ref{sec:clus}.

\section{Related Work}
\label{sec:rlwk}
\textbf{Fine-grained categorization:} Benefiting from the development of DCNNs, \eg AlexNet~\cite{alexnet12hiton}, VGGNet~\cite{vggnet15zisserman}, InceptionNet~\cite{googlenet15Szegedy}, ResNet~\cite{resnet16kaiming}, the study of FGVC has been gradually shifted from strongly-supervised~\cite{pncnn@perona,dlac@cvpr,spdacnn@cvpr} to weakly-supervised~\cite{racnn@mei,dfbnet18larry,mamc18eccv} in recent years. In the weakly-supervised configuration, to induce models to learn features from the mostly discriminative regions, creating structural relationship between labels through either intermediate concepts~\cite{hyperlabels@cvpr,multigranularity@iccv} or shared attributes~\cite{bigraph@cvpr,describeattr@aaai}, often accompanied by data augmentation~\cite{kp@cvpr}, has been proposed. Multi-task learning is typically used to make the learning feasible~\cite{hyperlabels@cvpr,multigranularity@iccv,embeddinglabel@cvpr}. 
% However, optimizing metric loss involving in multi-task learning still poses a challenge and involves a non-trivial sample selection procedure~\cite{wu2017sampling}. 
Another line of research localizes semantic parts first and then learns feature from the localized parts in a multi-stage learning framework~\cite{twolevelattention@cvpr,neuralconstellations@iccv,pickfilter@tianqi}. Recently, this line of research combines part localization and feature learning in an end-to-end framework~\cite{racnn@mei,macnn@mei,ram@arxiv,dfbnet18larry}. Exploring relationships between objects in different images for part feature learning has also been investigated but with limited performance~\cite{mamc18eccv}, due to the non-trivial sample selection involved in optimizing the loss function. Our approach is a step towards improving the efficiency and effectiveness of robustly exploring relationships between different images. We explore correlations between objects from different images in regularization learning and learn robust multi-scale features.

\textbf{Multi-scale features:} Exploiting multi-scale features improves the performance of many visual tasks. Among them, a number of methods make predictions by combining results inferred from multiple individual layers~\cite{ssd16szegedy,mscnn16cai}, several other approaches first combine multiple layer features and then make a prediction~\cite{fcn15darrell,hypercolumn15girshick,hypernet16kong}. These approaches marry low-level features' spatial resolution with high-level features' semantic properties. More recent studies have constructed high-resolution multi-scale semantic features by building feature pyramids in DCNNs through lateral connections of bottom-up and top-down feature maps~\cite{fpn17kaiming}. Nonlinear and progressive connecting structures are studied in~\cite{dla@darrell} to enhance the exploitation of multi-scale features. Multi-scale features have also studied using multi-granularity labels~\cite{hyperlabels@cvpr,multigranularity@iccv}. These approaches learn multi-scale features by training networks with different granularity of labels. Our work also involves the utilization of multi-scale features but exploits the interactions between features at different scales by matching prediction distributions of different layer feature maps.

%-------------------------------------------------------------------------
\begin{figure*}[t]
\begin{center}
\includegraphics[width=0.9\linewidth]{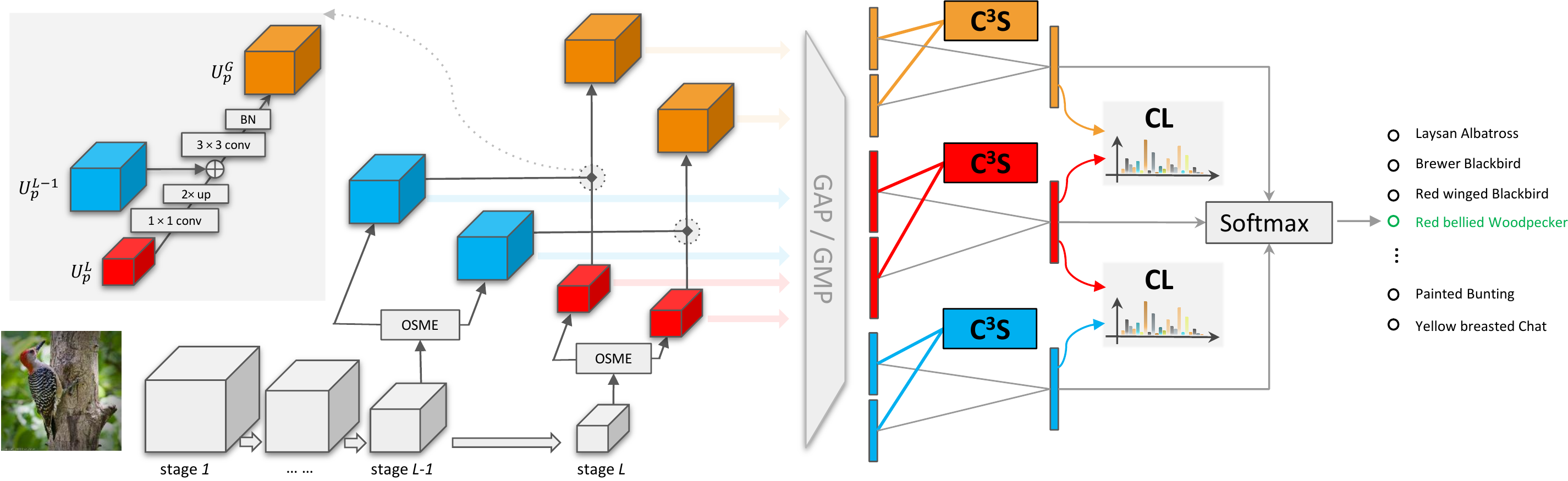}
\end{center}
\caption{Overview of our approach. Our network outputs multiple feature maps by employing the OSME block. Two OSME blocks, each with two excitations, are depicted in the last two stages to illustrate our approach. Feature maps from stage $L-1$ (blue) and $L$ (red) are combined to generate the merged feature maps (orange). Top-left corner is a zoomed in display of the merging process of the merged feature maps. Feature maps are then aggregated to obtain the corresponding pooling features through GAP or GMP. The pooled features from the same stage are mutually constrained by the $C^3S$ regularizer and are simultaneously concatenated to feed into a fully-connected layer to generate logits. The logits are constrained through the $CL$ regularizer after conversion into class probabilities and are combined for classification. Best viewed in color.}
\label{fig:diagram}
\end{figure*}

\section{Approach}
\label{sec:models}
% In this section we introduce the proposed Cross-X learning for robust fine-grained recognition.
Cross-X learning involves two main components: 
1) A \textit{cross}-category \textit{cross}-semantic regularizer ($C^3S$) that learns semantic part features by leveraging the correlations between different images (Sec.~\ref{sec:fsc}).
2) A \textit{cross}-layer regularizer ($CL$) that learns robust  features by matching prediction distributions between different layers (Sec.~\ref{sec:clr}).
An overview of our approach is depicted in Fig.~\ref{fig:diagram}.
%Note that Cross-X learning is independent of the backbone architectures; we use ResNet-50~\cite{resnet16kaiming} with SE blocks~\cite{senet17cvpr} as an instantiation.

\subsection{Preliminaries}
\label{sec:pre}
We begin by briefly reviewing the one-squeeze multi-excitation (OSME) block~\cite{mamc18eccv} that learns multiple attention region features for each input image.
Let $\mathbf{U}=[u_1,\cdots,u_C] \in \mathbb{R}^{W\times H\times C}$ denote the output feature map of a residual block $\tau$. 
In order to generate multiple attention-specific feature maps, the OSME block extends the original residual block by performing one-squeeze and multiple-excitation operations.

Formally, OSME first performs global average pooling to squeeze $\mathbf{U}$ and produce a channel-wise descriptor $\mathbf{z}=[z_1,\cdots,z_C] \in \mathbb{R}^{C}$.
Then a gating mechanism is independently employed on $\mathbf{z}$ for each excitation module, $p=1, \cdots, P$, to output: % feature maps $\mathbf{m}^p$:
\begin{equation}
\label{eq:osme-mp}
\mathbf{m}^p=\sigma(\mathbf{W}_2^p\delta(\mathbf{W}_1^p\mathbf{z}))=[m_1^p,\cdots,m_C^p]\in\mathbb{R}^C,
\end{equation}
where $\sigma$ and $\delta$ refer to the Sigmoid and ReLU functions. Finally, the attention-specific features $\mathbf{U}_p$ are generated by re-weighting the channels of the original feature maps $\mathbf{U}$:
\begin{equation}
\label{osme-map}
\mathbf{U}_p = [m_1^p\mathbf{u}_1, \cdots, m_2^p\mathbf{u}_C]\in \mathbb{R}^{W\times H\times C}.
\end{equation}

Although OSME can generate attention-specific features, guiding these features to have semantic meanings is challenging. 
{\cite{mamc18eccv}} tackles this by optimizing a metric learning loss which pulls features from the same excitation closer and pushes features from different excitations away. However, optimizing such a loss still poses a challenge and involves a non-trivial sample selection procedure~\cite{wu2017sampling}.

\subsection{Cross-Category Cross-Semantic Regularizer}
\label{sec:fsc}
\begin{figure}[t]
\begin{center}
   \includegraphics[width=0.9\linewidth]{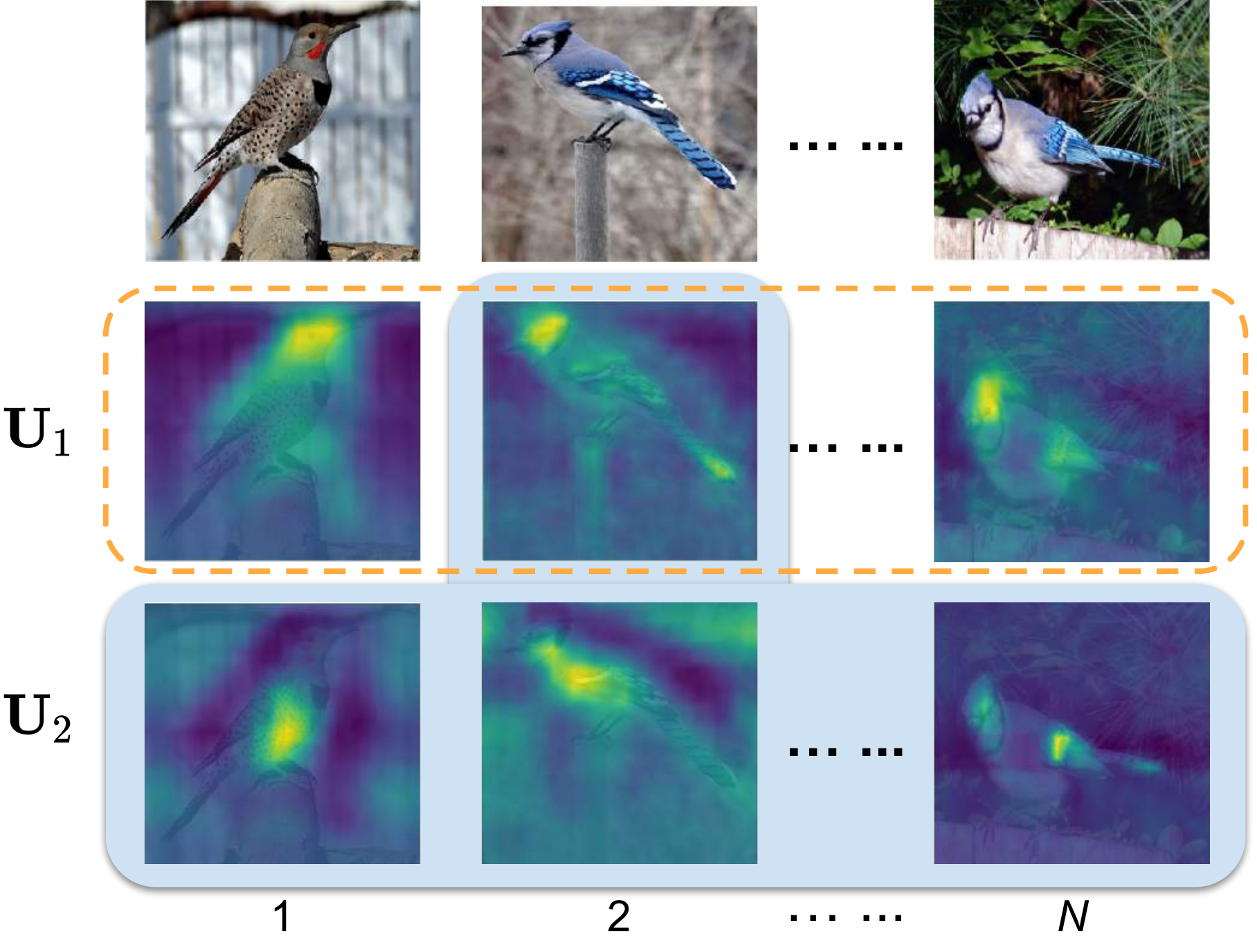}
\end{center}
   \caption{An illustration of the $C^3S$ learning. Take the center image as an example, $C^3S$ encourages the excitation modules, $\mathbf{U}_1$ and $\mathbf{U}_2$, to be activated on different semantic parts by exploiting relationships between features from different images (orange dash box) and features from different excitation modules (blue shaded box). Best viewed in color.}
\label{fig:c3s}
\end{figure}

Instead of optimizing a metric loss as in ~\cite{mamc18eccv}, we propose to learn semantic features by exploring the correlations between feature maps from different images and different excitation modules.
Ideally, we want the extracted features from the same excitation module to have the same semantic meaning, even though they come from different images with different class labels. And the extracted features from different excitation modules should have different semantic meanings, even though they come from the same image (see Fig.~\ref{fig:c3s} for an illustration).
To achieve this goal, we introduce the cross-category cross-semantic regularizer ($C^3S$) that maximizes the correlation of features from the same excitation module while minimizes the correlation of features from different excitation modules.

Formally, we first perform global average pooling (GAP) on $\mathbf{U}_p$ to obtain the corresponding pooled features $\mathbf{f}_p\in\mathbb{R}^C$, followed by $\ell_2$ normalization ($\mathbf{f}_p\leftarrow\mathbf{f}_p/\Vert\mathbf{f}_p\Vert$). Then the correlations between all pairs of excitation modules $p$ and $p'$ form a matrix $S$:
\begin{equation}
\label{eq:cs}
S_{p,p'} = \frac{1}{N^2}\sum {\mathbf{F}_p}^T \mathbf{F}_{p'}, 
\end{equation}
where $T$ is the transpose operator, $N$ is the batch size and $\mathbf{F}_p = [\mathbf{f}_{p,1}, \cdots, \mathbf{f}_{p,N}] \in \mathbb{R}^{C\times N}$ is a matrix storing the pooled features from excitation module $p$ for all samples in the batch. 

The $C^3S$ regularization loss is then constructed from two parts: 1) maximizing the diagonal of $S$ to maximize the correlation within the same excitation module and, 2) penalizing the norm of $S$ to minimize the correlation between different excitation modules:
\begin{equation}
\label{seq:c3sreg}
\mathcal{L}_{C^3S}(S) = \frac{1}{2}\left (\|S\|_F^2 - 2 \|diag(S)\|_2^2\right ),
\end{equation}
where $\|\cdot\|$ is the Frobenius norm, and the $diag(\cdot)$ operator extracts the main diagonal of a matrix into a vector. 
Compared to the triplet based metric learning loss, $C^3S$ loss can be naturally integrated into the OSME block and is easily optimized without any sampling procedure.

\subsection{Cross-Layer Regularizer}
\label{sec:clr}
Exploiting  semantic features from different layers of CNNs has been shown to be beneficial to many vision tasks~\cite{fcn15darrell,hypercolumn15girshick,hypernet16kong,ssd16szegedy,mscnn16cai}. A simple extension of this idea to fine-grained recognition is to combine the prediction outputs of different layers for the final prediction.
However, we observe in our experiments that this simple strategy usually leads to inferior performance (see Sec.~\ref{sec:ablation}). 
We hypothesize that the problem is due to two reasons: 1) mid-level features are more sensitive to input changes ~\cite{understanding_CNNs14lecun} which makes them less robust for fine-grained recognition where the intra-class variation is large,
2) relationships between the predictions of  features are not exploited.
To alleviate these problems, we adapt the feature pyramid network (FPN)~\cite{fpn17kaiming} to integrate features from different layers and propose a novel cross-layer regularizer ($CL$) that learns robust  features by matching the prediction distribution between different layers.

Formally, let $\mathbf{U}^{L}=\{\mathbf{U}_p^L\}_{p=1}^P$, $\mathbf{U}^{L-1}=\{\mathbf{U}_p^{L-1}\}_{p=1}^P$ be the feature maps at stage $L$ and $L-1$ (here a stage refers to a group of layers that produce feature maps with the same size ~\cite{resnet16kaiming}).
We generate the merged feature maps $\mathbf{U}_p^G$ in a similar way to FPN~\cite{fpn17kaiming} but with two differences.
First, the dimensionality reduction of $\mathbf{U}_p^L$ is performed before up-sampling.
%This operation can \xyang{polish} preserve information better while reducing computations. 
Second, batch normalization (BN)~\cite{bn15Szegedy} is used after the anti-aliasing operation on the merged feature maps. The procedure can be summarized as:
\begin{equation}
\mathbf{U}_p^G = \mathbf{BN}\left (\mathbf{K}_2 \ast \left (\mathbf{U}_p^{L-1}+\text{Bilinear}(\mathbf{K}_1 \ast \mathbf{U}_p^L) \right ) \right ),
\end{equation}
where $\ast$ is convolutional operation, Bilinear($\cdot$) denotes bilinear interpolation, $\mathbf{K}_1$, $\mathbf{K}_2$ are $1\times 1$ and $3\times 3$ filters, respectively. 
$\mathbf{U}^G$ integrates the property of fine spatial resolution in the mid-level layers and the rich high-level semantic in the top-level layers.

To further exploit the relationships between the predictions of features, we propose the $CL$ regularizer that matches the prediction distribution between different layers. Let $\mathbf{Pr}^L = \sigma(f(U^L))$ and $\mathbf{Pr}^{L-1}=\sigma(f(U^{L-1}))$ be the prediction outputs of stage $L$ and $L-1$, where $\sigma(\cdot)$ is the softmax function and $f(\cdot)$ denotes the output layer. 
The $CL$ regularizer encourages $\mathbf{Pr}^{L-1}$ to match $\mathbf{Pr}^{L}$ by minimizing their KL-divergence:
\begin{align}
\label{eq:klreg}
\begin{split}
\mathcal{L}_{CL}(\mathbf{Pr}^{L}, \mathbf{Pr}^{L-1})&= \text{KL}(\mathbf{Pr}^{L}\ ||\  \mathbf{Pr}^{L-1})\\
& =\frac{1}{N}\sum_{n=1}^{N}\sum_{k=1}^K p_{nk}^{L}\log \frac{p_{nk}^{L}}{p_{nk}^{L-1}},
\end{split}
\end{align}
where $K$ is the number of classes.
A similar regularizer can be added to constrain the feature maps $U^L$ and $U^G$ as well.
The $CL$ regularizer can be viewed as knowledge distillation ~\cite{distill15hinton} that uses ``soft targets" from $U^L$ with rich structure information to guide the feature learning of $U^{L-1}$ and $U^G$.

\subsection{Optimization}
\label{sec:fpe}
Given the feature maps $\mathbf{U}^L$, $\mathbf{U}^{L-1}$ and $\mathbf{U}^G$, our final prediction can be obtained by combining their prediction outputs:
\begin{equation}
\label{eq:softmax}
\mathbf{Pr} = \sigma\left ( f(\mathbf{U}^L) + f(\mathbf{U}^{L-1}) + f(\mathbf{U}^G)\right ).
\end{equation}
Putting this all together, the full objective function of Cross-X learning is:
\begin{equation}
\label{eq:obj}
\mathcal{L} = \mathcal{L}_{data} + \gamma \mathcal{L}_{C^3S} + \lambda \mathcal{L}_{CL},
\end{equation}
\begin{equation}
\label{eq:datafit}
\mathcal{L}_{data}=-\frac{1}{N}\sum_{n=1}^N\sum_{k=1}^K c_{nk} \log p_{nk},
\end{equation}
\begin{equation}
\label{eq:c3s_loss}
\mathcal{L}_{C^3S}=\gamma_1\mathcal{L}_{C^3S}(S^L) + \gamma_{2}\mathcal{L}_{C^3S}(S^{L-1}) + \gamma_3\mathcal{L}_{C^3S}(S^G),
\end{equation}
\begin{equation}
\label{eq:kl_loss}
\mathcal{L}_{CL}=\lambda_{1}\mathcal{L}_{CL}(\mathbf{Pr}^L, \mathbf{Pr}^{L-1}) + \lambda_2\mathcal{L}_{CL}(\mathbf{Pr}^L, \mathbf{Pr}^G),
\end{equation}
where $L_{data}$ is the classification loss, $\gamma$ and $\lambda$ are hyper-parameters that balance the contribution of different costs.
Our model can be trained end-to-end using stochastic gradient descent (SGD) and does \textit{not} require other optimization tricks such as multiple crops~\cite{ntscnn@eccv}, data augmentation~\cite{kp@cvpr}, model ensemble~\cite{macnn@mei}, and separate initialization~\cite{dfbnet18larry}

\section{Experiments}
\label{sec:exp}

\subsection{Datasets and Baselines}
\label{sec:datasets}

\begin{table}
\small
\begin{center}
\begin{tabular}{|l|c|c|c|}
\hline
Datasets & $\#$category & $\#$training & $\#$testing \\
\hline\hline
NABirds\cite{nabirds15Perona} & 555 & 23,929 &24,633\\
\hline
CUB-Birds~\cite{cubbirds11caltech} & $200$ & $5,994$ & $5,794$ \\
\hline
Stanford Cars~\cite{stcars13feifei} & 196 & 8,144 &8,041 \\
\hline
Stanford Dogs~\cite{stdogs11feifei} & 120 & 12,000 &8,580\\
\hline
FGVC-Aircraft~\cite{vggaircraft13Vedaldi} & 100 & 6,667 &3,333\\
\hline
\end{tabular}
\end{center}
\caption{The statistics of fine-grained datasets in this paper}
\label{tab:datasets}
\end{table}

\textbf{Datasets:} We conduct experiments on five fine-grained visual categorization datasets, including NABirds\cite{nabirds15Perona}, Caltech-UCSD Birds (CUB-Birds)~\cite{cubbirds11caltech}, Stanford Cars~\cite{stcars13feifei}, Stanford Dogs~\cite{stdogs11feifei} and  FGVC-Aircraft~\cite{vggaircraft13Vedaldi}. Note that NABirds is a recently released dataset with much larger scale and many more  fine-grained categories. The detailed statistics such as category numbers and data splits are summarized in Tab.~\ref{tab:datasets}. We report top-1 accuracy in this study. %We use top-1 accuracy as the evaluation metric.

\textbf{Baselines:} We compare our approach with various state-of-the-art methods using weakly-supervised learning for fine-grained recognition. For fair comparison, we mainly compare to the results with ResNet-50 as their backbone network and include the best results of VGG based methods for completeness in the following, unless otherwise stated. 
In addition, an ablation study of Cross-X learning is analyzed based on the SENet backbone \cite{senet17cvpr}, since OSME is a direct extension of the SE block. Moreover, we also report results of our approach on the ResNet-50 backbone \cite{resnet16kaiming}.
%
%In addition, due to OSME is an extention of the SE block, we present the ablation study of Cross-X learning on the backbone of SENet \cite{senet17cvpr} for consistency. We also report results of our approach on the backbone of ResNet-50 \cite{resnet16kaiming}, since it
All the baselines are listed as follows:
%\textbf{Baselines:} We compare our approach with various state-of-the-art methods using weakly-supervised learning for fine-grained recognition. For fair comparison, we mainly compare with the results with ResNet-50 as their backbone network and include the best results of VGG based methods for completeness. Unless otherwise explained, all results reported in the following are from models implemented with ResNet-50. We also re-implement SE-ResNet-50~\cite{senet17cvpr} to better illustrate the effectiveness of our proposed Cross-X learning approach. All the baselines are listed as follows:
\begin{itemize}
\itemsep0em 
\item \textbf{FCAN}~\cite{fcan@lin}: fully convolutional attention network that adaptively selects multiple task-driven visual attentions by reinforcement learning.
\item \textbf{RA-CNN}~\cite{racnn@mei}: recurrent attention  convolutional neural network that localizes discriminative areas and extracts features from coarse to fine scale.
\item \textbf{DT-RAM}~\cite{ram@arxiv}: recurrent visual attention model that selects a sequence of regions through a dynamic continue/stop gating mechanism. 
\item \textbf{MA-CNN}~\cite{macnn@mei}: multi-attention convolutional neural network that generates multiple parts from spatially-correlated channels via multi-task learning.
\item \textbf{DFB-CNN}~\cite{dfbnet18larry}: discriminative filter bank approach that learns a bank of convolutional filters that capture class-specific discriminative patches.
\item \textbf{NTS-Net}~\cite{ntscnn@eccv}: navigator-teacher-scrutinizer network finds consistent informative regions through multi-agent cooperation. 
\item \textbf{MAMC-CNN}~\cite{mamc18eccv}: multi-attention multi-class constraint approach that learns soft attention masks by regularizing features from different images.
\item \textbf{MaxEnt-CNN}~\cite{maxent@nips}: maximum entropy approach provides a training routine to maximize the entropy of the output probability distributions for FGVC.
\end{itemize}

\subsection{Implementation Details}
\label{sec:iplt}
% \begin{table}[t]
% \small
% \begin{center}
% %\begin{tabularx}{\linewidth}{|c|c|c|c|c|c|}
% \begin{tabular}{@{}@{\extracolsep{\fill}}|c|c|c|c|c|c|@{}}
% \hline
% params	&CUB-Birds 	& Dogs 	& Cars 	& Aircraft & NABirds \\
% \hline\hline
% $\#$P 		& 2 	& 3 	& 2 	&2 		& 2 \\
% \hline
% $\gamma^L$ 		& 1 	& 1 	&1		&0.5	&0.1\\
% \hline
% $\gamma^{L-1}$ 	& 0.25 	& 0.5 	&0.25 	&0.1	&0.25\\
% \hline
% $\gamma^G$ 		& 1 	& 1 	&1		&0.1	&0.5\\
% \hline
% \end{tabular}
% % \begin{tablenotes}
% % $\#$parts is an alternative to $\#$excitations in this paper.
% % \end{tablenotes}
% \end{center}
% \caption{The hyper-parameters involved in our approach.}
% \label{tab:hyperparams}
% \end{table}
We develop our model in PyTorch, on top of the implementation of SENet/ResNet-50.
Specifically, we place the OSME block after $\verb'conv5_3'$ and $\verb'conv4_6'$ in SENet/ResNet-50.
The size of the output feature maps of the two blocks are $14\times 14 \times 2048$ and $28\times 28 \times 1024$, respectively.
Therefore, the channel sizes of $\mathbf{U}^L$, $\mathbf{U}^{L-1}$ and $\mathbf{U}^G$ are 4096, 2048 and 2048 when $P=2$. 
We initialize most of our network using the weights pretrained on ImageNet and initialize the newly introduced layers (OSME blocks, FPN blocks) from scratch.
No part or bounding box annotations are used during training.

Our network is trained using SGD on a single NVIDIA P6000 GPU with momentum 0.9 and a mini-batch size of 32. The initial learning rate is set to be 0.01 except for the experiments on Stanford Dogs where 0.001 is used. We train the network for 30 epochs and decay the learning rate by 0.1 every 15 epochs. For datasets that do not provide a validation set, we randomly take $10\%$ out of the training samples from each category for validation. Input images are cropped to $448\times 448$ and flipped horizontally with a probability of $0.5$. We report our results on a single scale of $448\times 448$ from a single model. More details can be found in the supplementary material.

\subsection{Ablation Studies}
\label{sec:ablation}
\begin{figure}[t]
\begin{center}
   \includegraphics[width=\linewidth]{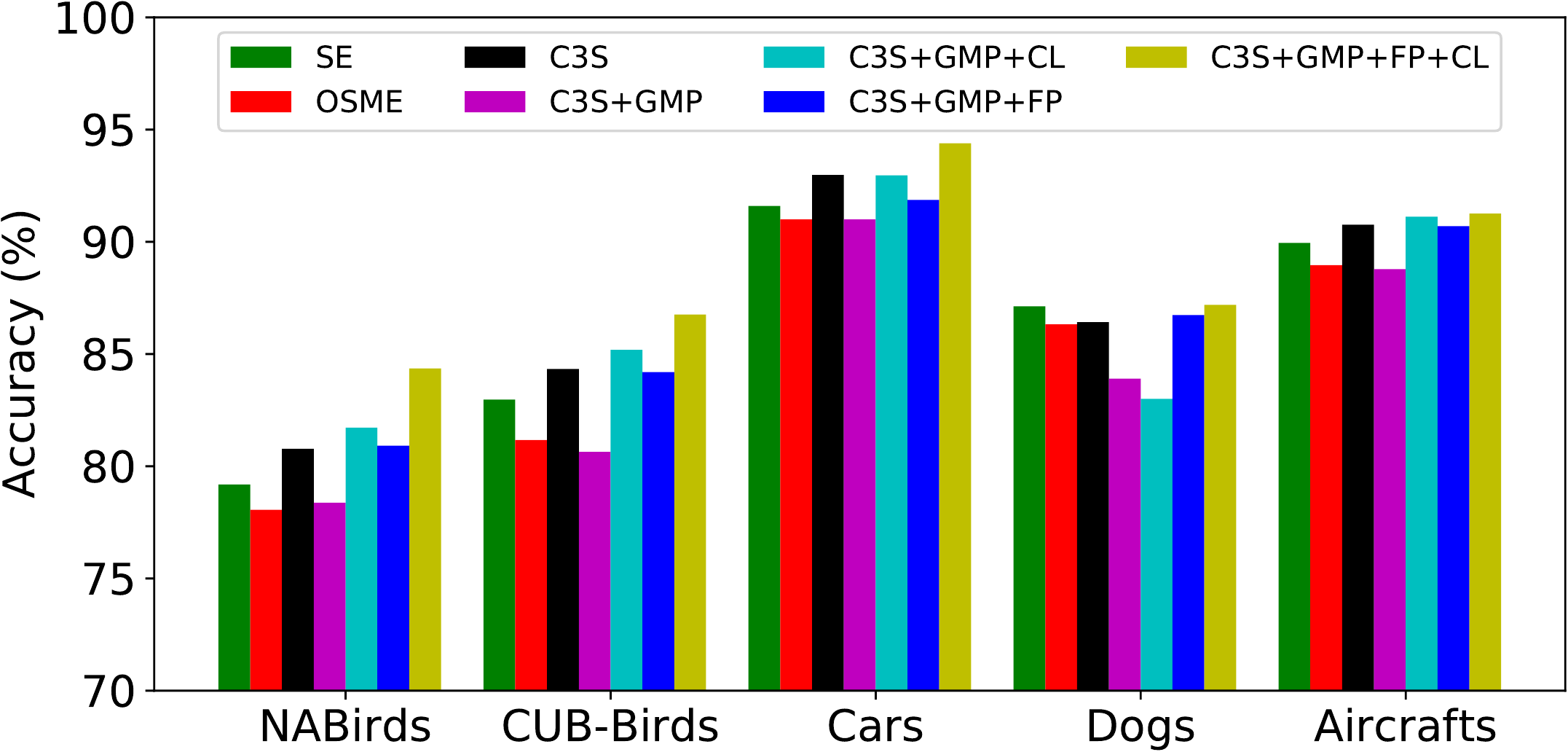}
\end{center}
   \caption{Ablation performance on 5 benchmark datasets with GMP employed on $\mathbf{U}_p^{L-1}$. The legend only shows the added block/regularizer names with the default ResNet-50 backbone omitted, \eg SE means SENet-50. Best viewed in color.} %CC-CS, FP and CLR mean cross-category cross-semantic regularizer, feature pyramid and cross-layer regularizer, respectively.}
\label{fig:gmpresults}
\end{figure}

\textbf{Effectiveness of $C^3S$ and $CL$:} The effectiveness of our regularization is studied in Fig.~\ref{fig:gmpresults}. We find the performance of our base network (OSME, putting the OSME block after $\verb'conv5_3'$ in SENet-50.) is lower than that of the SENet-50 on almost all datasets (SE vs. OSME), this indicates the training difficulties when employing the OSME block for multiple outputs. As we expected, $C^3S$ can effectively regularize the learning of our network to force excitations in the OSME block to be activated on different semantic parts, thus resulting in better features for classification (C3S vs. OSME). In addition, we find combining mid-level (stage $L-1$) and high-level (stage $L$) features without a constraint between them results in a performance drop (C3S vs. C3S+GMP). 
% This demonstrates the weakness of sensitive to variation of the mid-level feature. 
However, $CL$ can effectively increase the robustness of the mid-level feature and thus boosting the performance (C3S+GMP vs. C3S+GMP+CL).

\begin{figure}[t]
\begin{center}
   \includegraphics[width=\linewidth]{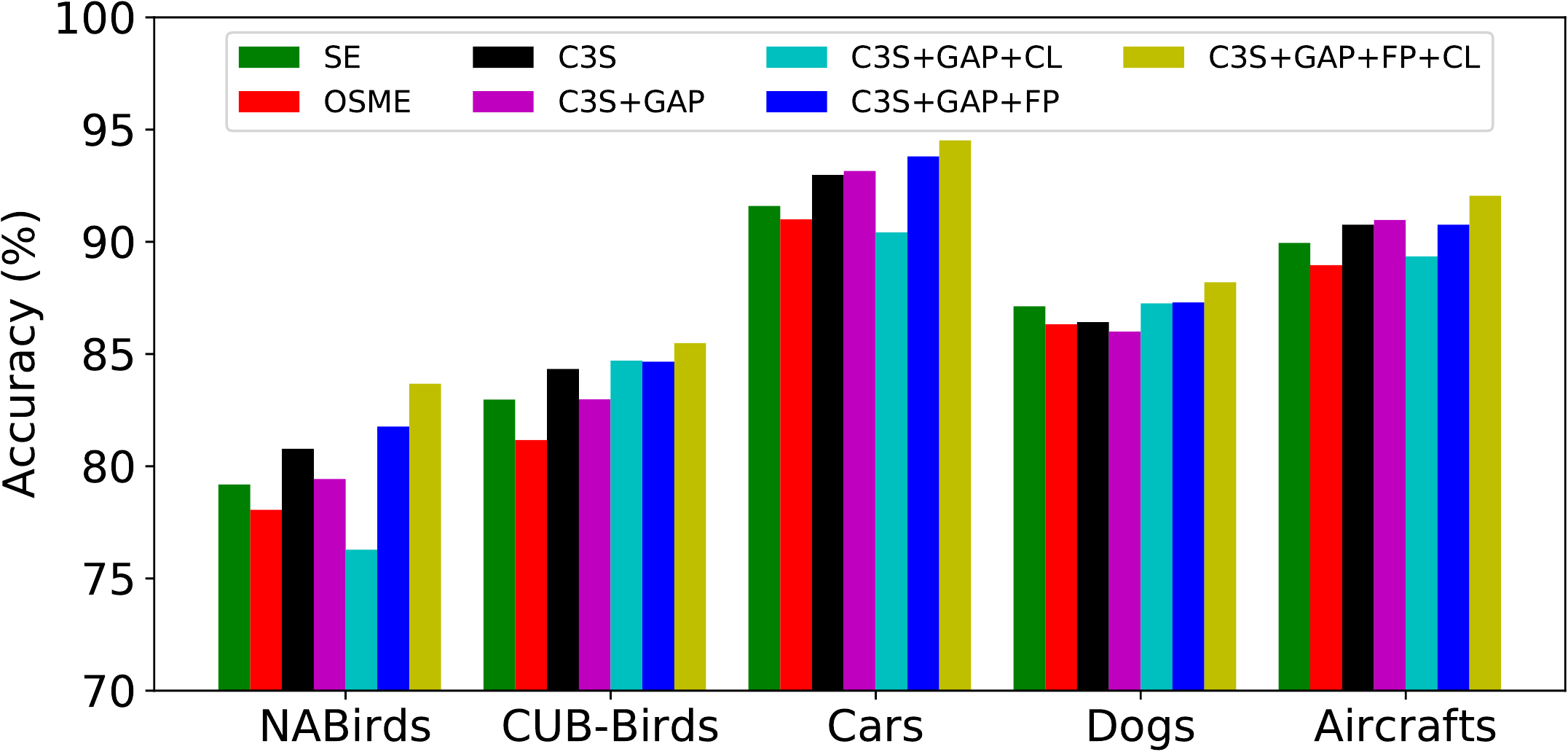}
\end{center}
   \caption{Ablation performance on 5 benchmark datasets with GAP employed on $\mathbf{U}_p^{L-1}$. C3S, CL and FP represent $C^3S$, $CL$ and merged feature maps, respectively. Best viewed in color.}
\label{fig:gapresults}
\end{figure}

\textbf{Benefits from the merged feature maps:} Employing the merged feature maps can bring systematic performance improvement on all datasets, whether $CL$ is used or not (C3S+GxP vs. C3S+GxP+FP and C3S+GxP+CL vs. C3S+GxP+FP+CL in Fig.~\ref{fig:gmpresults}--\ref{fig:gapresults}). This validates the effectiveness of our proposal that extra semantic features can be introduced to improve the FGVC performance and, the correctness of our operation that generates the merged feature maps. An interesting observation is that the performance of C3S+GxP+FP is systematically lower than that of C3S+GxP+FP+CL in Fig.~\ref{fig:gmpresults}--\ref{fig:gapresults}; this signifies that increasing the robustness of the newly introduced merged feature maps is also necessary and it further demonstrates that $CL$ has the capability to improve the robustness of mid-level features.    

\begin{table}
\small
\begin{center}
%\begin{tabularx}{0.5\textwidth}{|X|X|X|X|X|X|}
%\begin{tabular}[width=0.8\linewidth]{@{}@{\extracolsep{\fill}}|c|c|c|c|c|c|@{}}
\begin{tabular*}{\linewidth}{@{}@{\extracolsep{\fill}}|c|c|c|c|c|c|@{}}
%\begin{tabulary}{20cm}{|c|c|c|c|c|c|}
\hline
		            &NABirds	&CUB-Birds 	&Cars	&Dogs	&Aircraft\\
\hline\hline
GMP{\scriptsize -}	&$\mathbf{81.7}$   &$\mathbf{85.2}$	&93.0	&83.0	&91.1\\
GAP{\scriptsize -}	&76.3   &84.7	&90.4	&87.3	&89.4\\
\hline
GMP{\scriptsize +} 	&80.9   &84.2	&91.9 	&86.7 	&90.7\\
GAP{\scriptsize +} 	&$\mathbf{81.7}$   &84.7   &$\mathbf{93.8}$   &$\mathbf{87.3}$   &$\mathbf{91.3}$\\
\hline
\end{tabular*}
\end{center}
\caption{Performance of our approach on five benchmark datasets with GAP and GMP alternatively employed on $\mathbf{U}_p^{L-1}$. The top group compares results from the approach with $CL$ but without merged feature maps. The bottom group shows results from the approach with merged feature maps but without $CL$.}
\label{tab:rslt-gmpgap}
\end{table}

\begin{figure*}[t]
\begin{center}
\begin{minipage}{0.095\linewidth}
\includegraphics[width=0.99\linewidth]{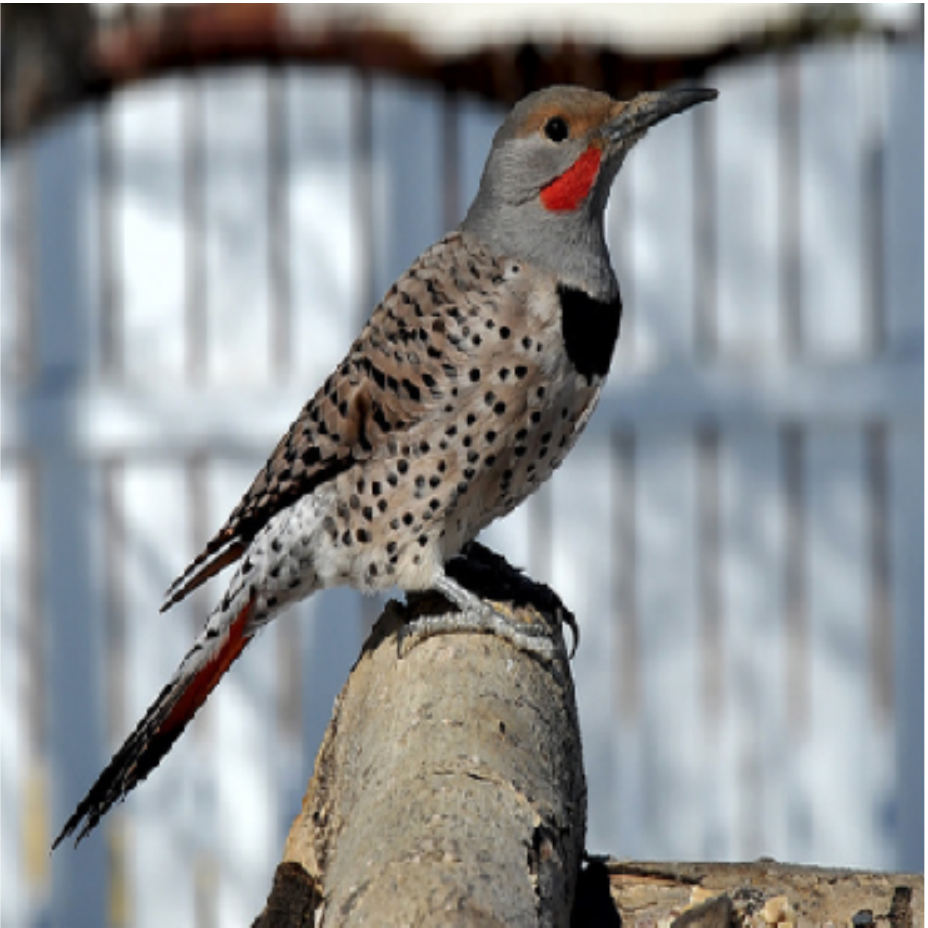}\\
\includegraphics[width=0.99\linewidth]{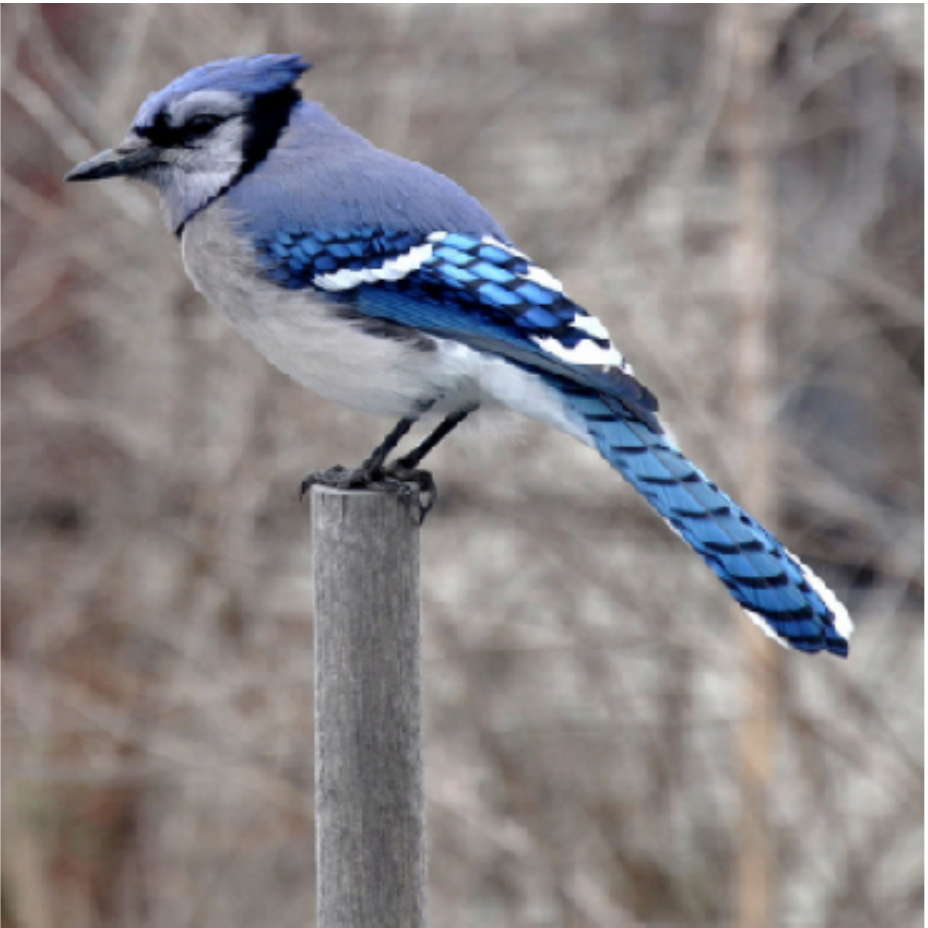}\\
\includegraphics[width=0.99\linewidth]{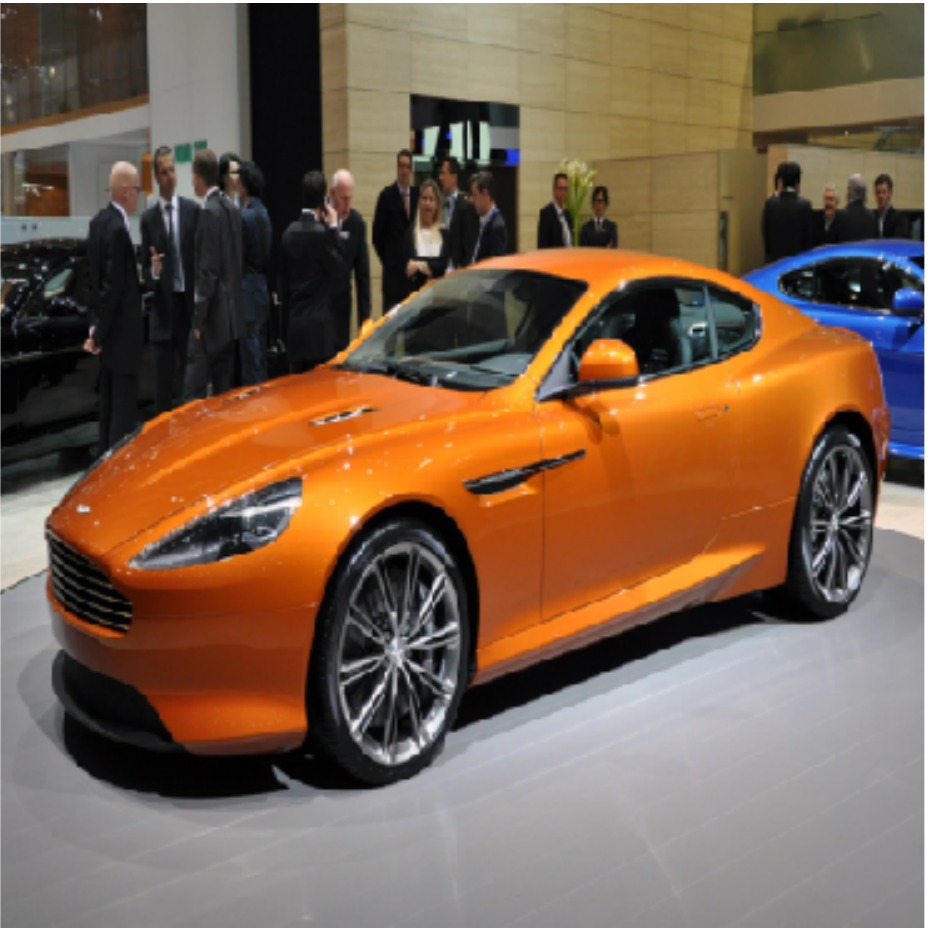}\\
\includegraphics[width=0.99\linewidth]{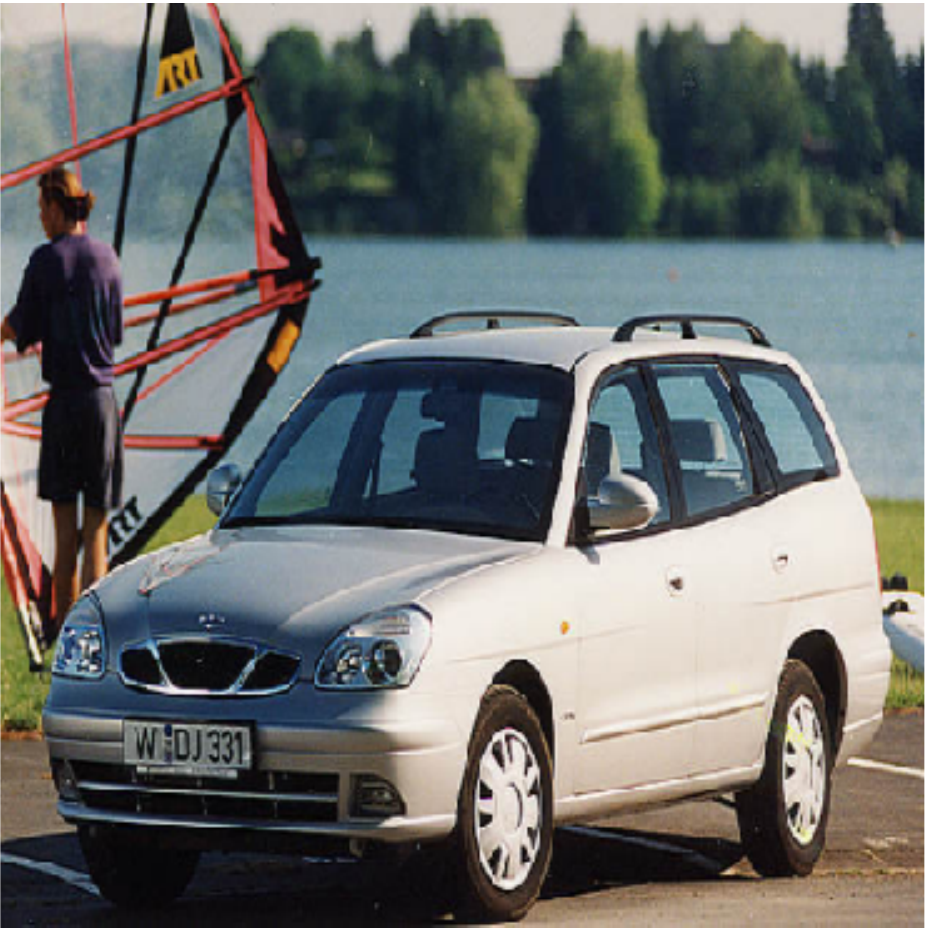}\\
\includegraphics[width=0.99\linewidth]{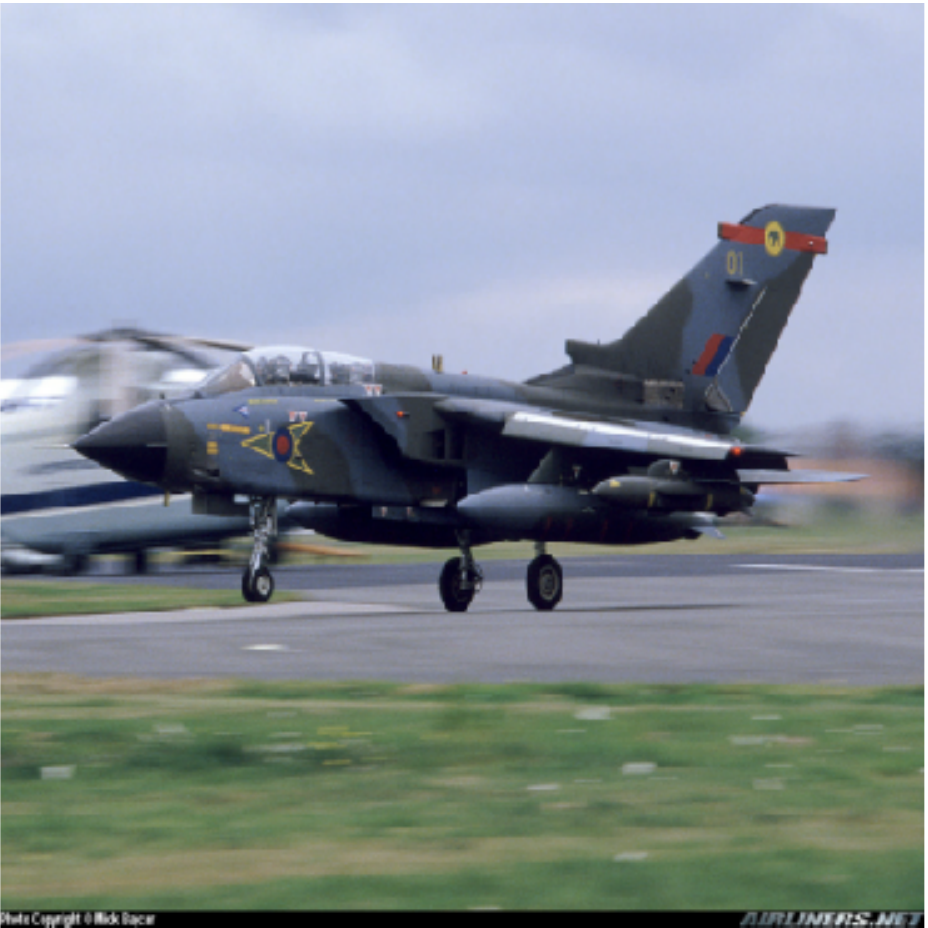}\\
\includegraphics[width=0.99\linewidth]{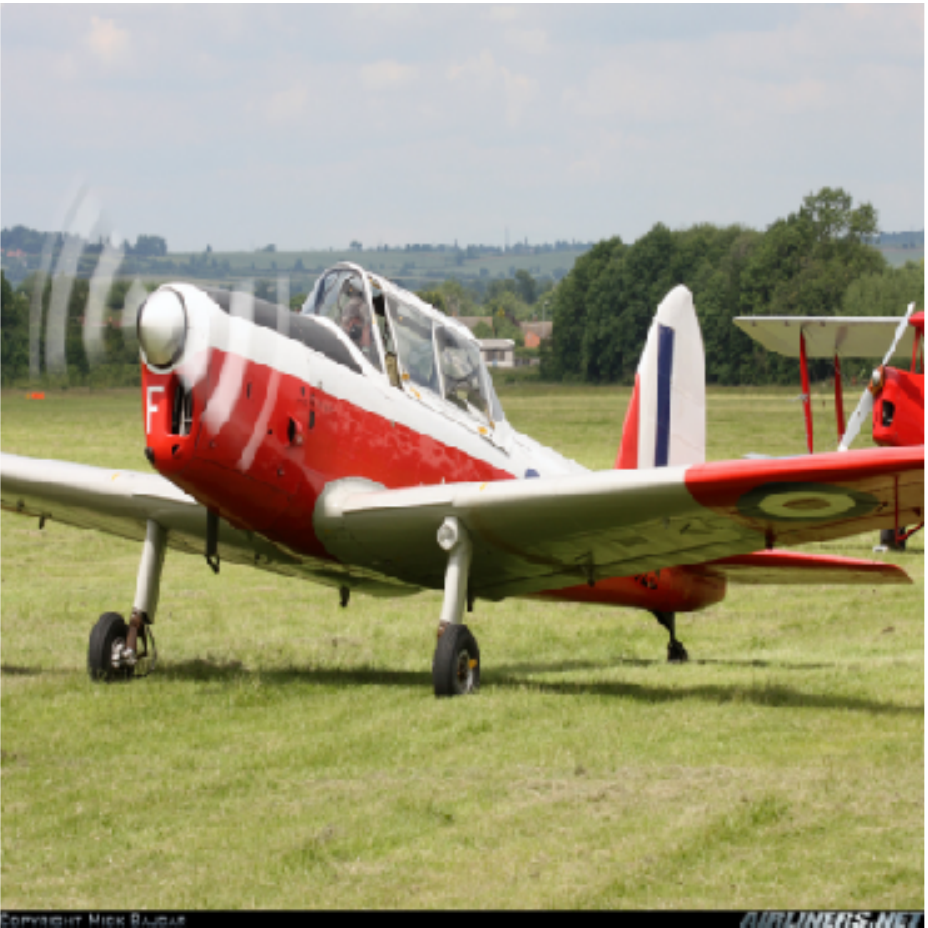}\\
\centering (a)
\end{minipage}
\begin{minipage}{0.21\linewidth}
\includegraphics[width=0.45\linewidth]{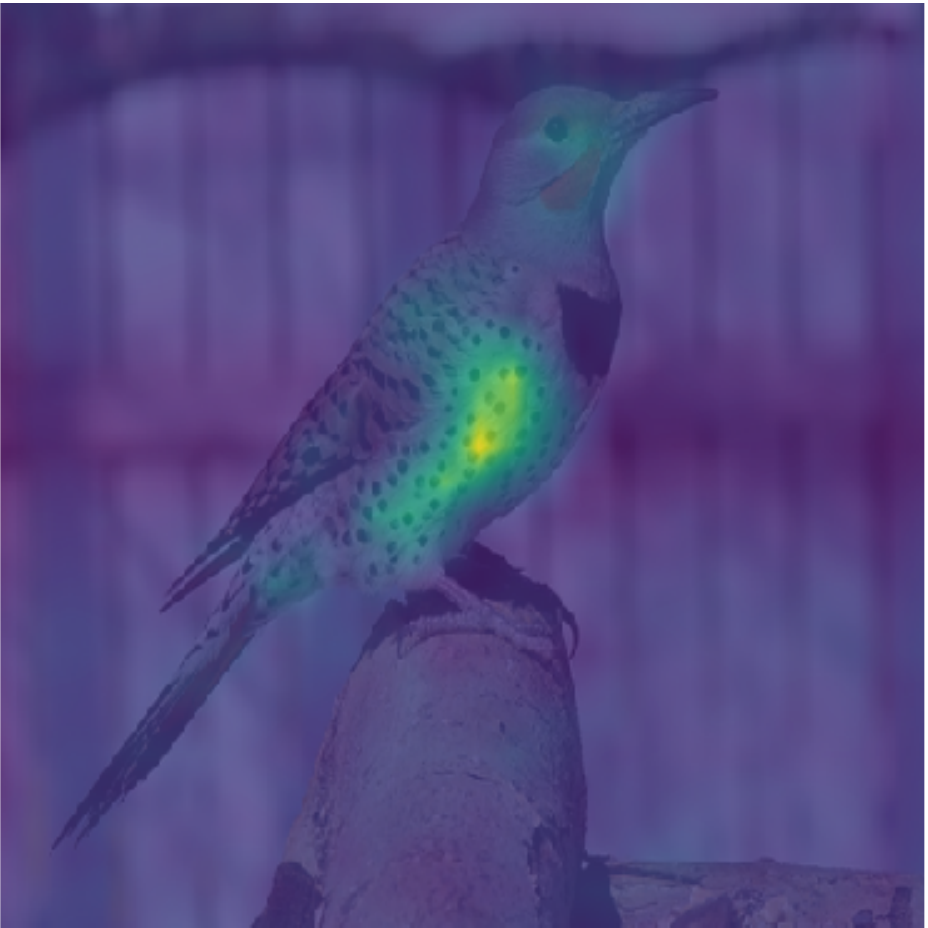}
\includegraphics[width=0.45\linewidth]{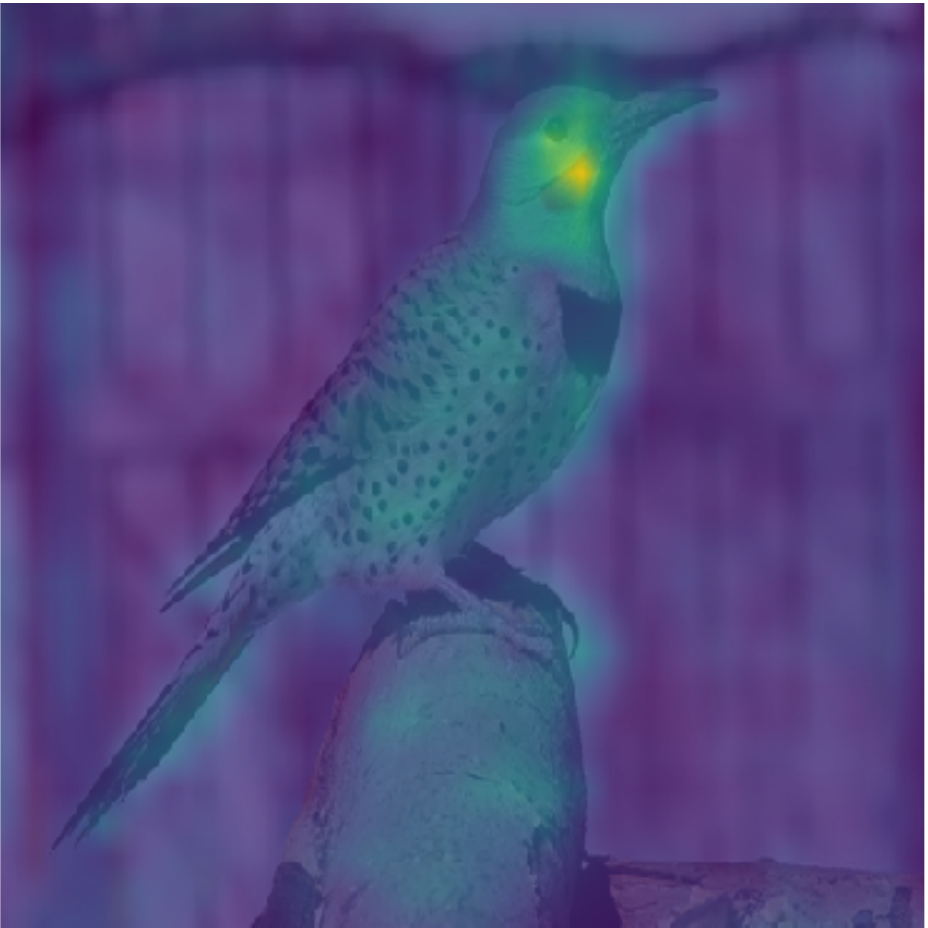}\\
\includegraphics[width=0.45\linewidth]{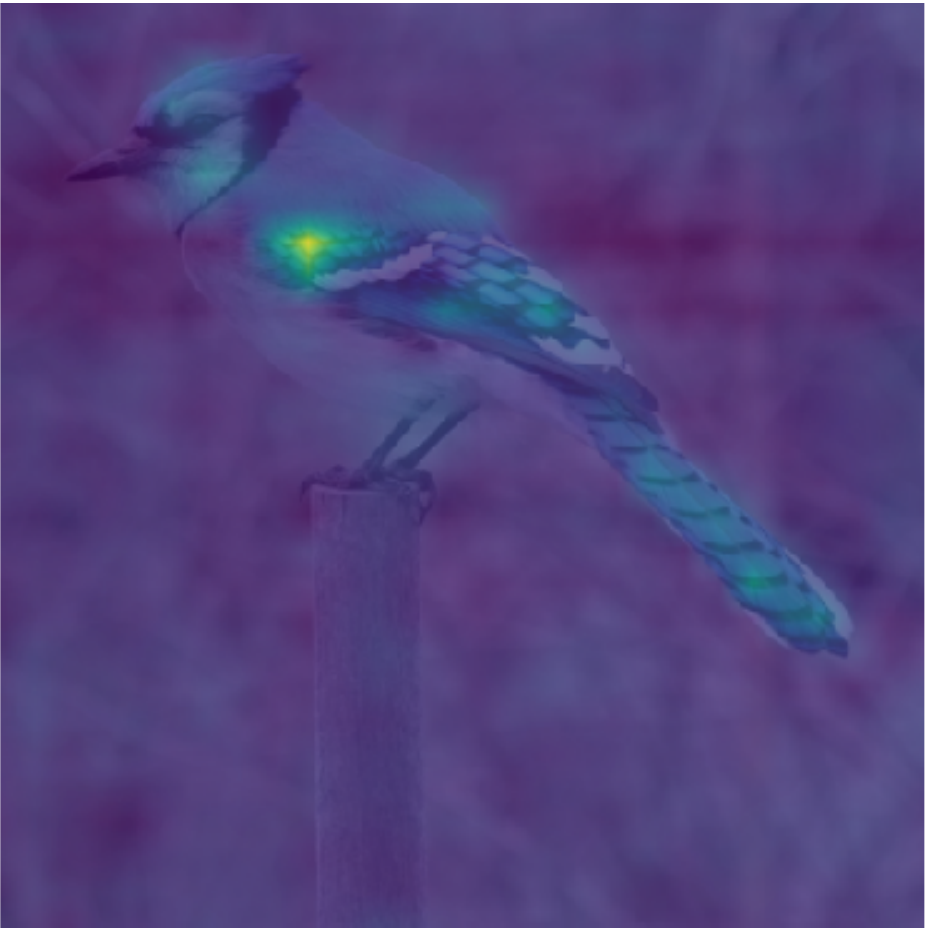}
\includegraphics[width=0.45\linewidth]{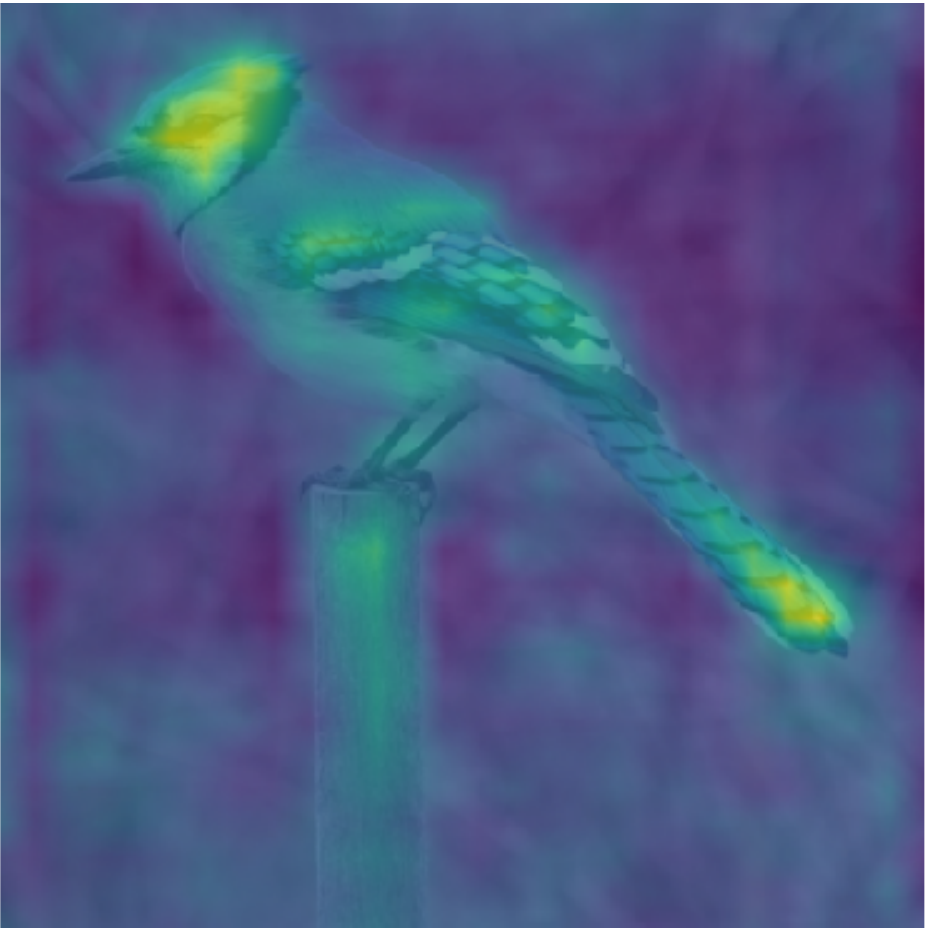}\\
\includegraphics[width=0.45\linewidth]{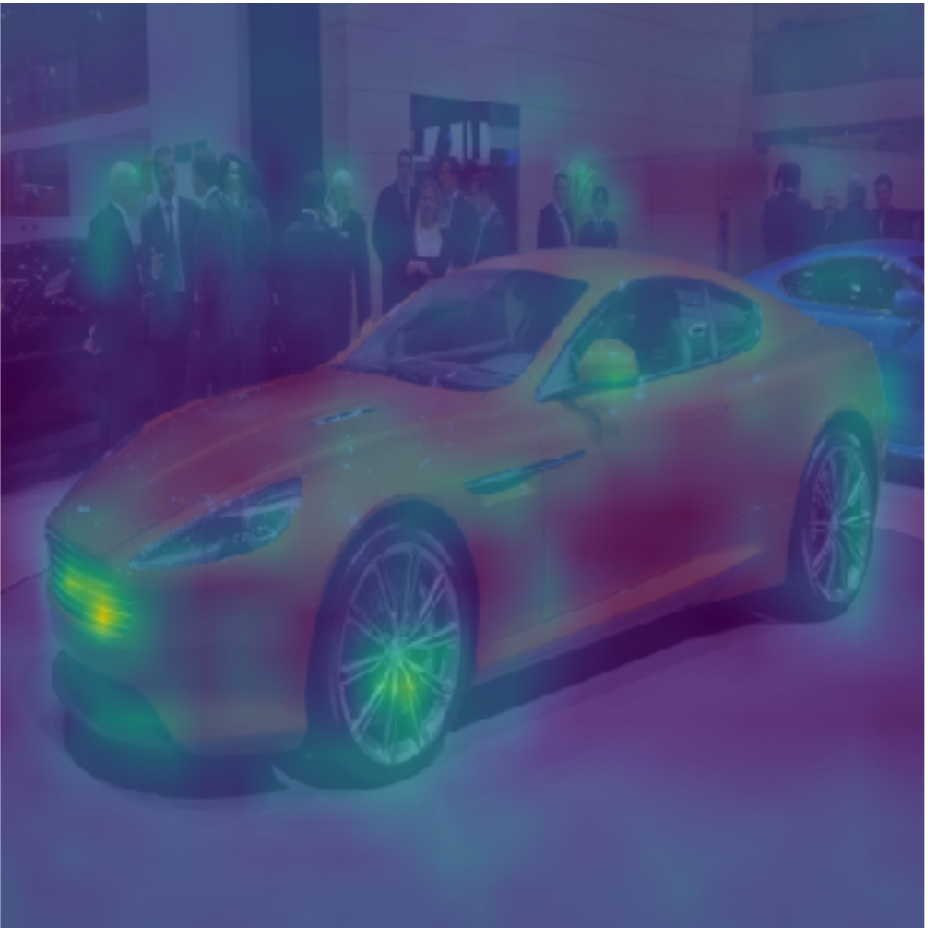}
\includegraphics[width=0.45\linewidth]{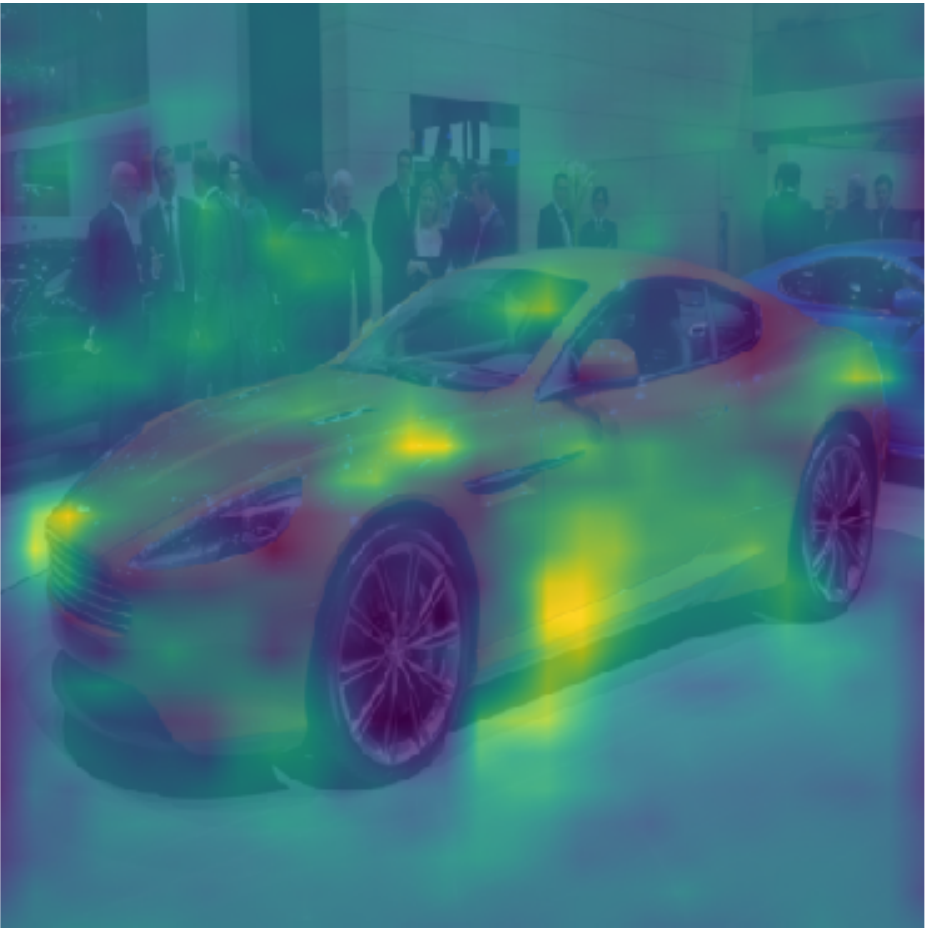}\\
\includegraphics[width=0.45\linewidth]{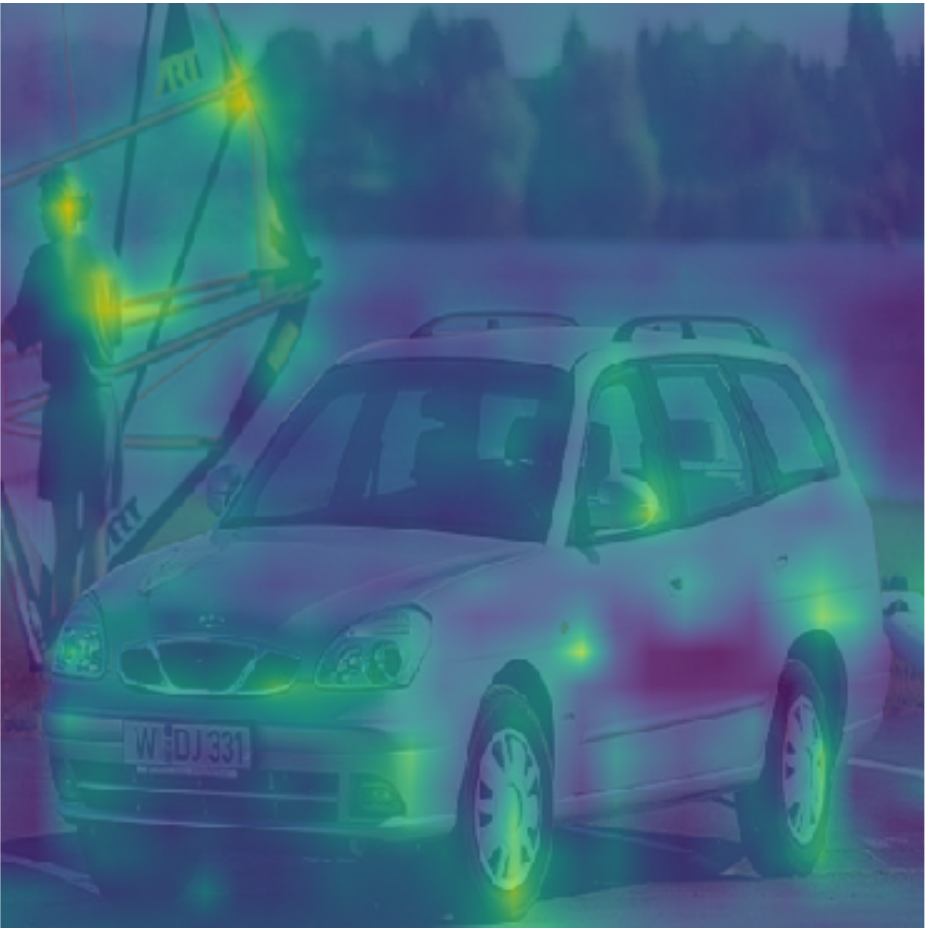}
\includegraphics[width=0.45\linewidth]{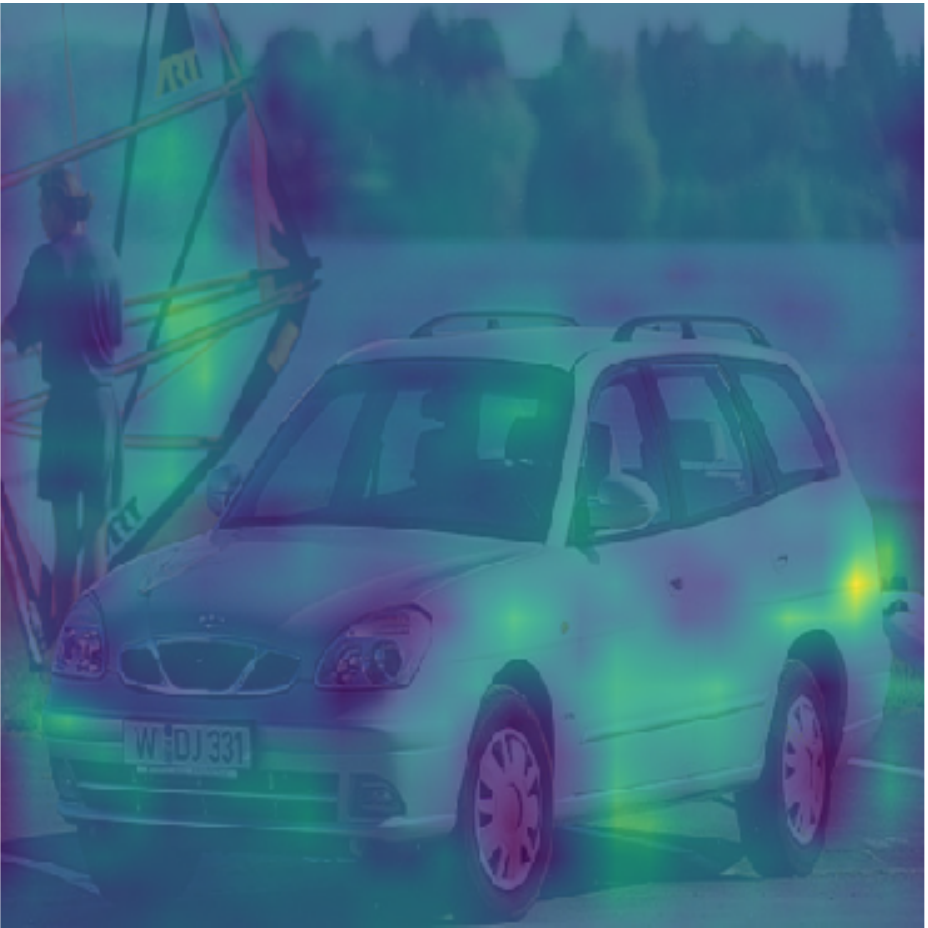}\\
\includegraphics[width=0.45\linewidth]{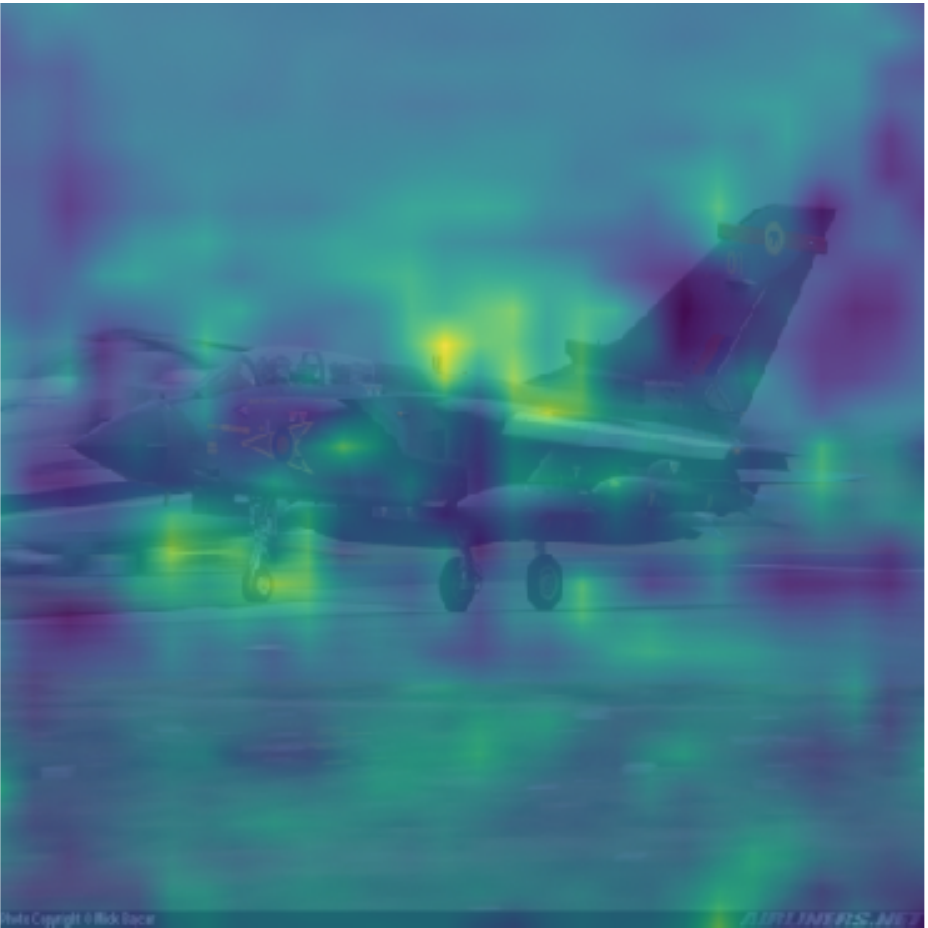}
\includegraphics[width=0.45\linewidth]{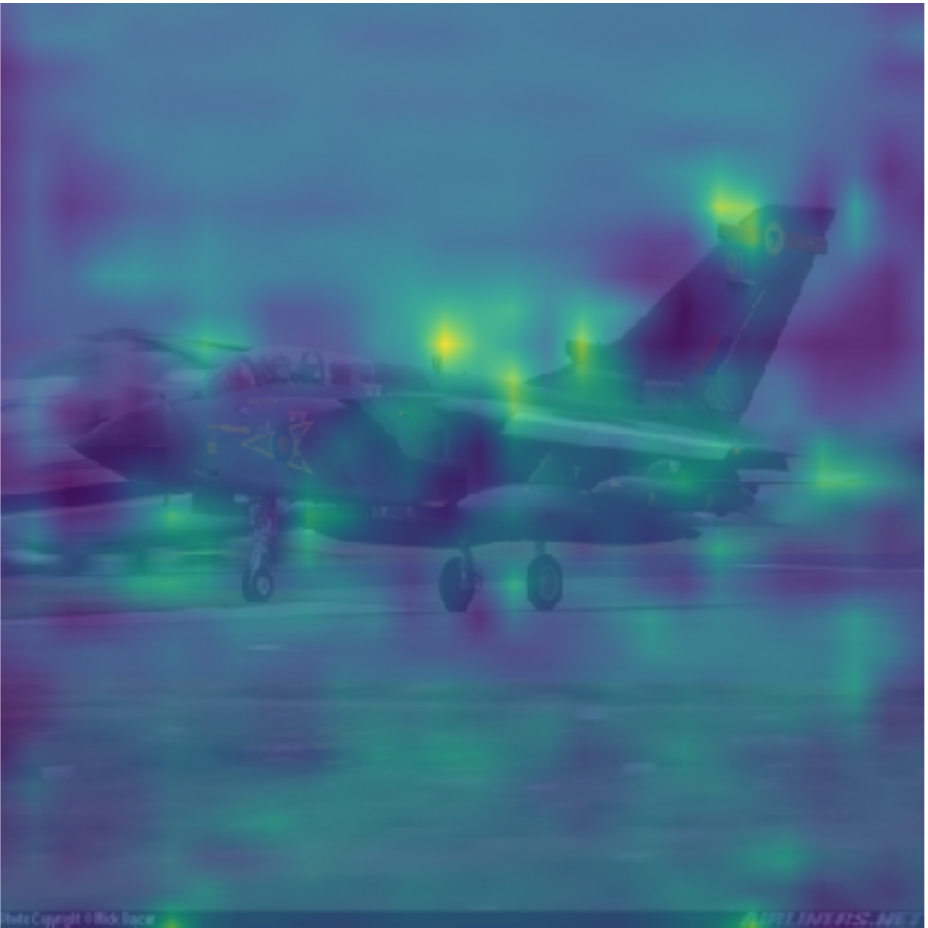}\\
\includegraphics[width=0.45\linewidth]{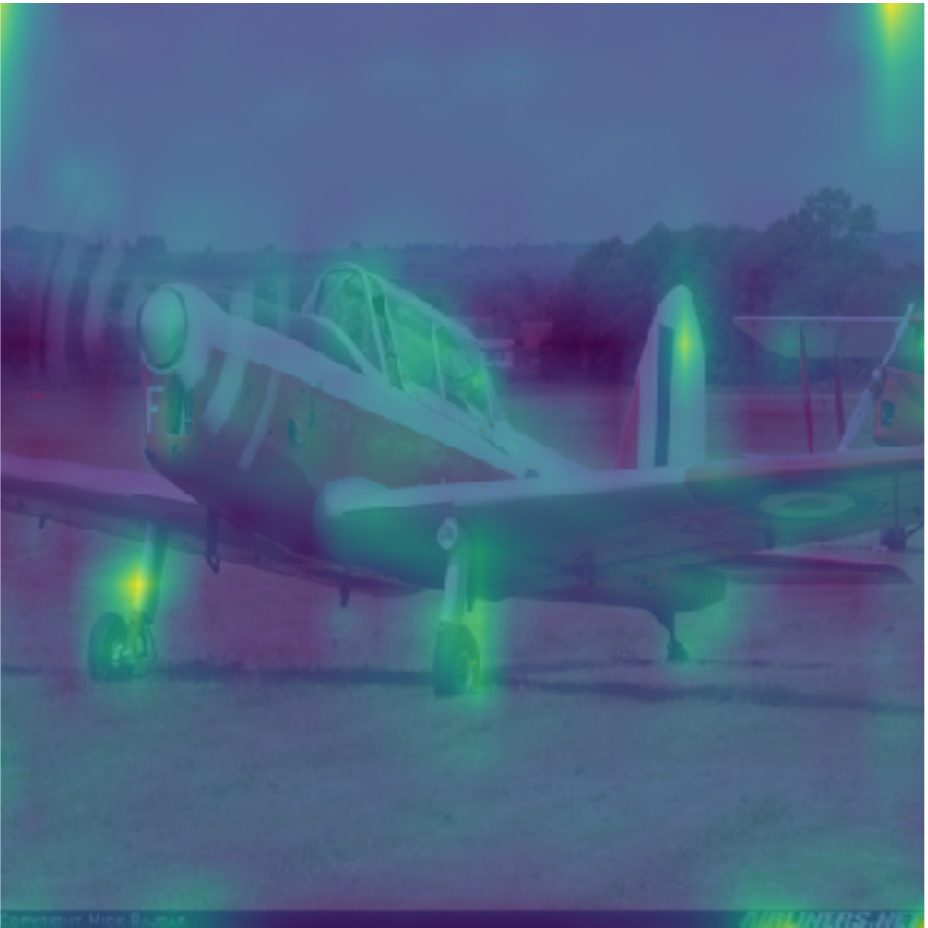}
\includegraphics[width=0.45\linewidth]{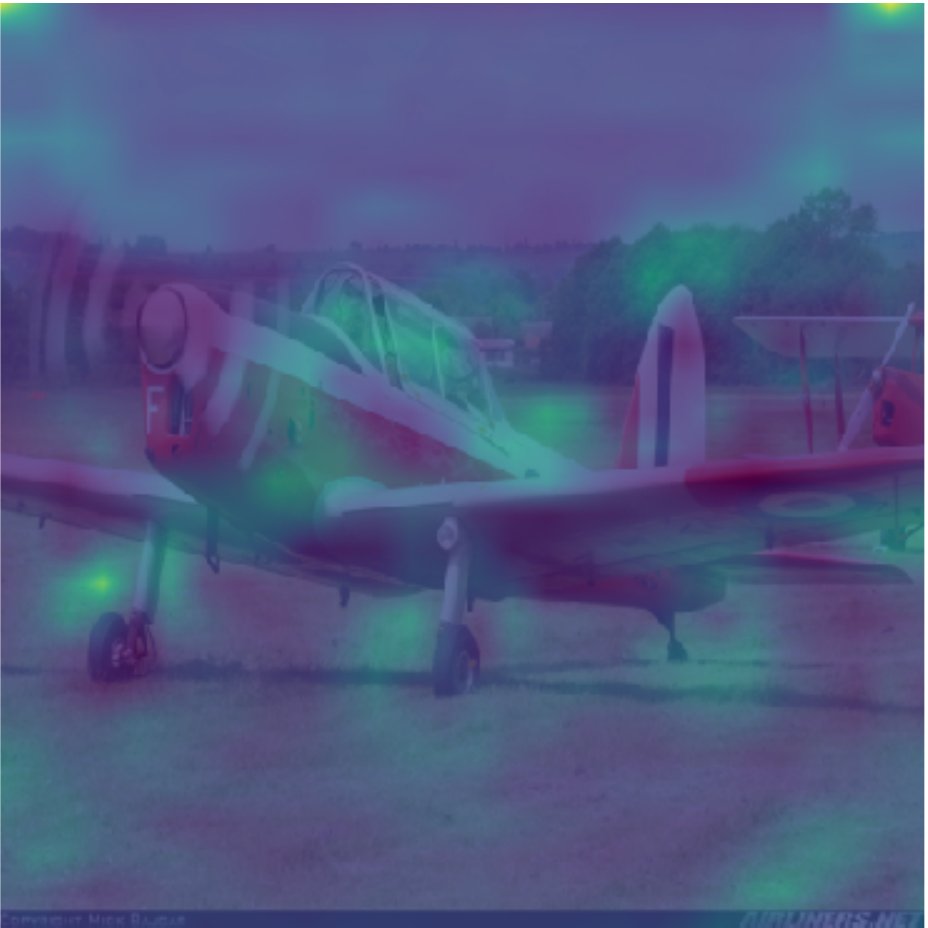}\\
\centering (b)
\end{minipage}
\begin{minipage}{0.21\linewidth}
\includegraphics[width=0.45\linewidth]{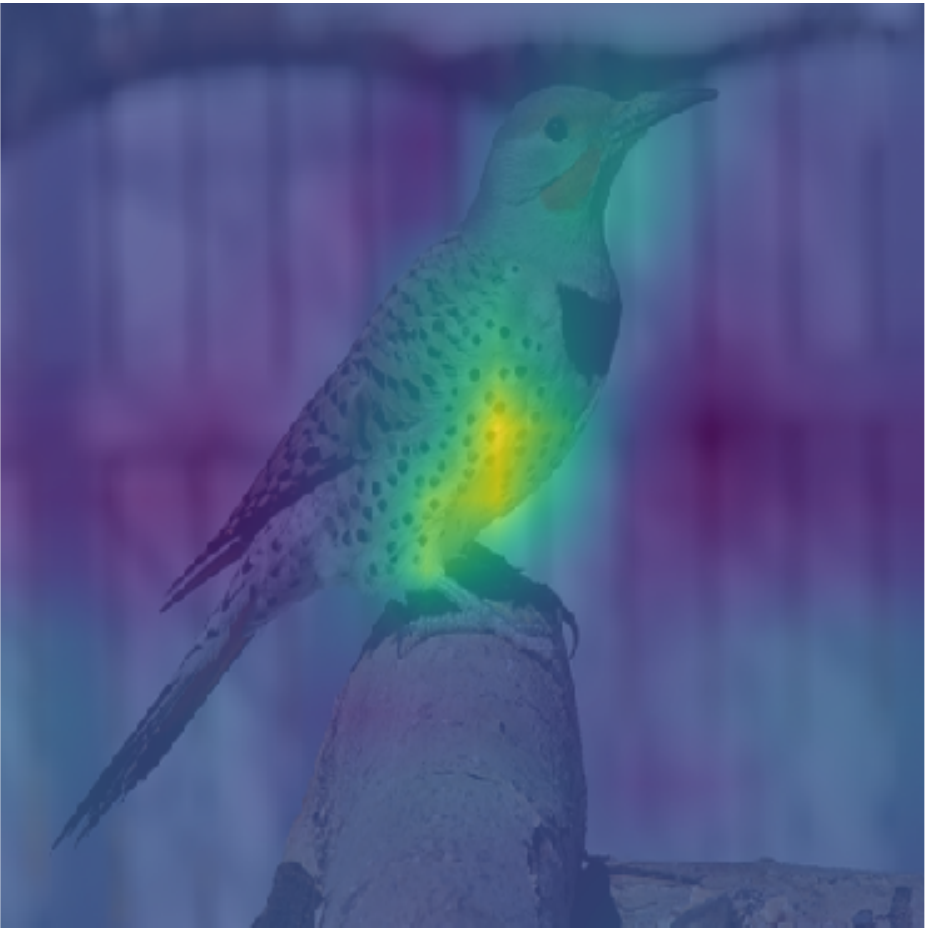}
\includegraphics[width=0.45\linewidth]{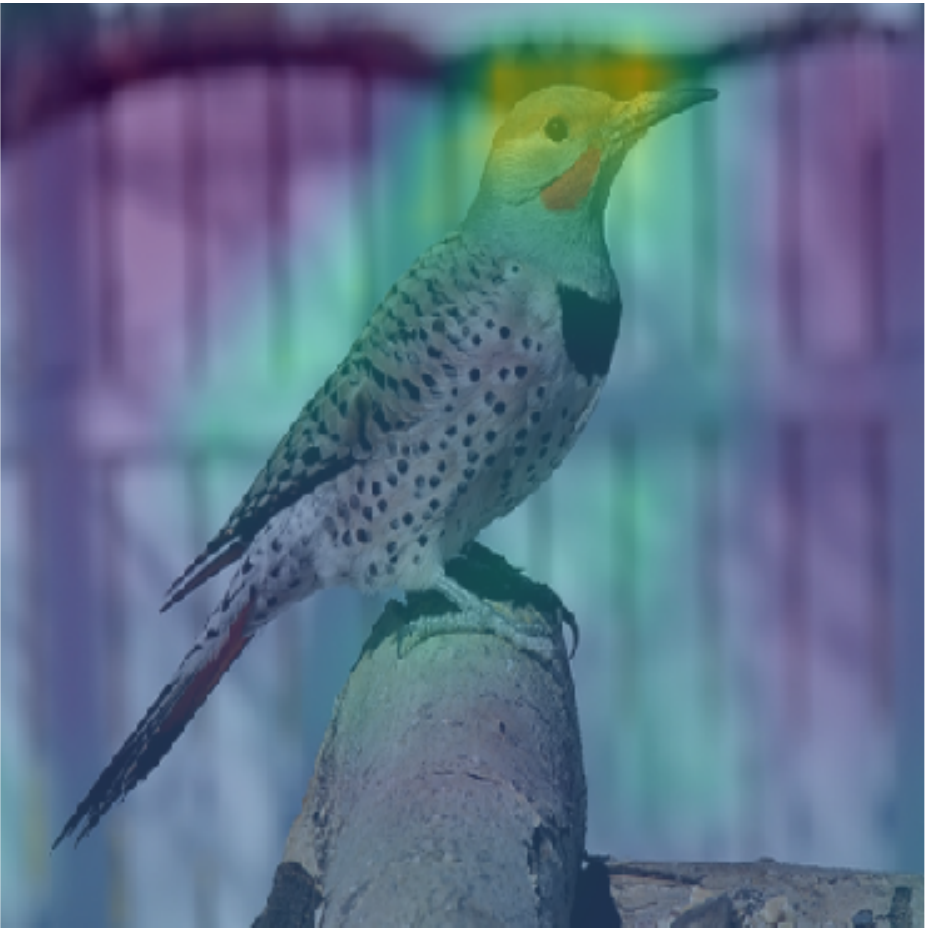}\\
\includegraphics[width=0.45\linewidth]{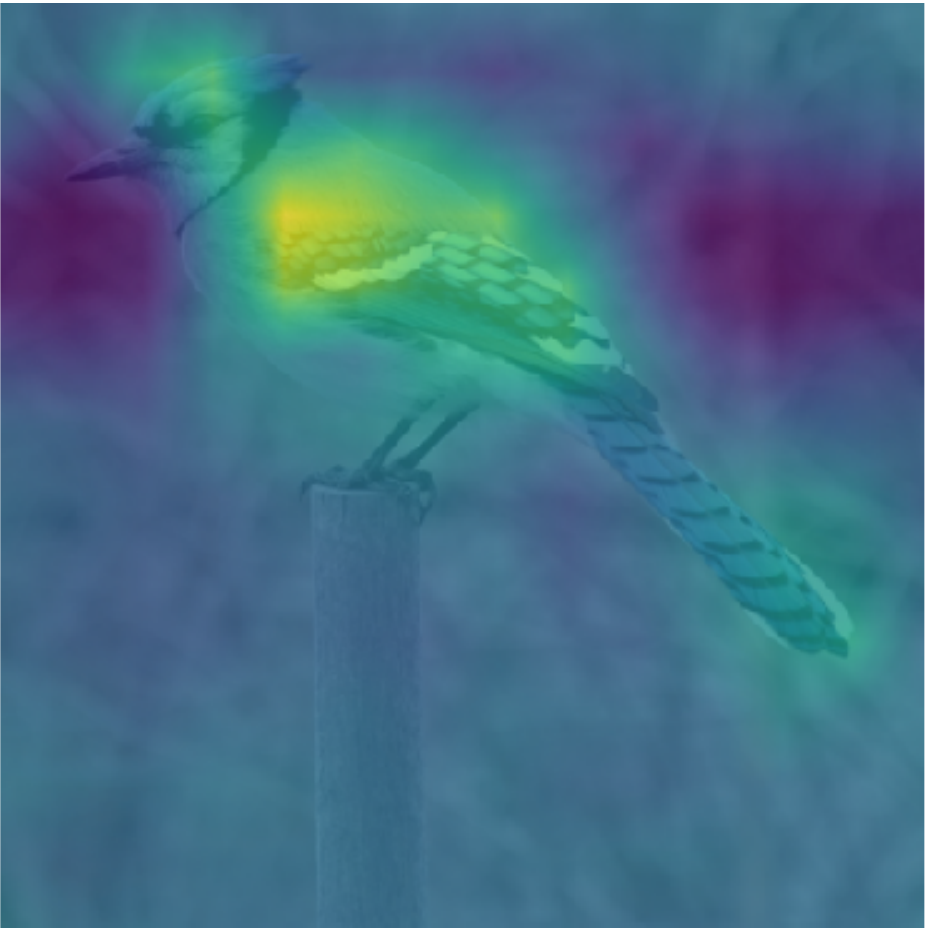}
\includegraphics[width=0.45\linewidth]{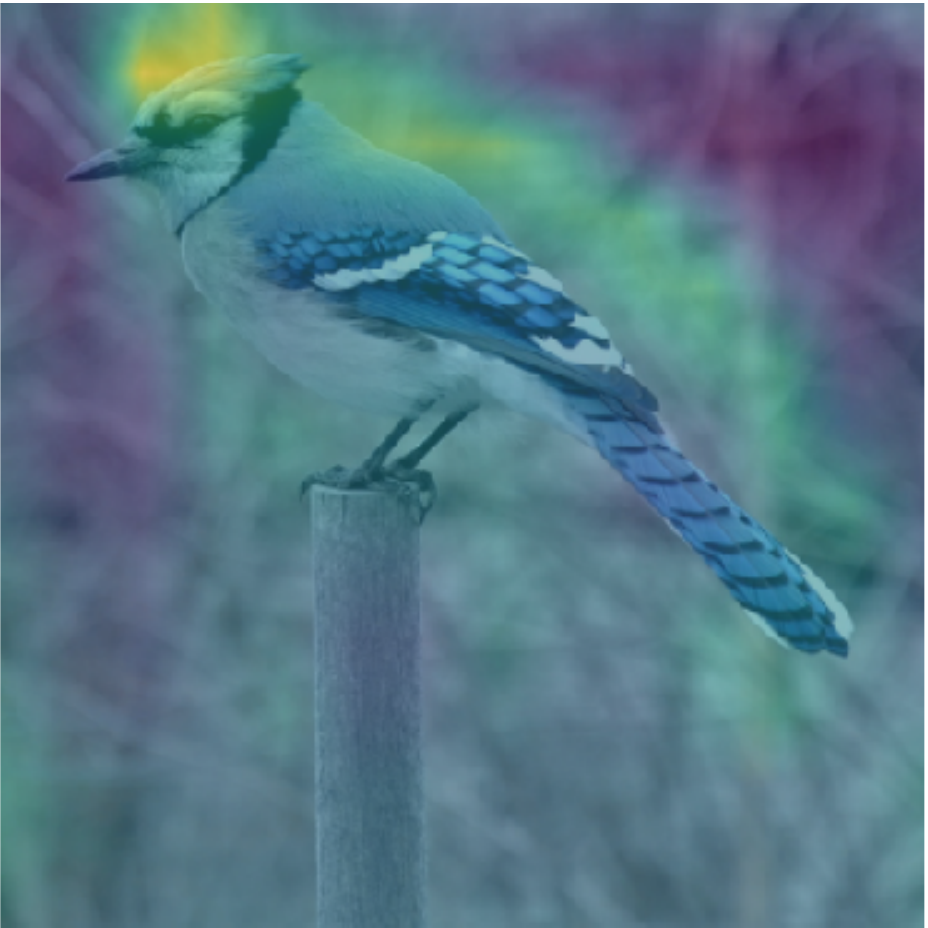}\\
\includegraphics[width=0.45\linewidth]{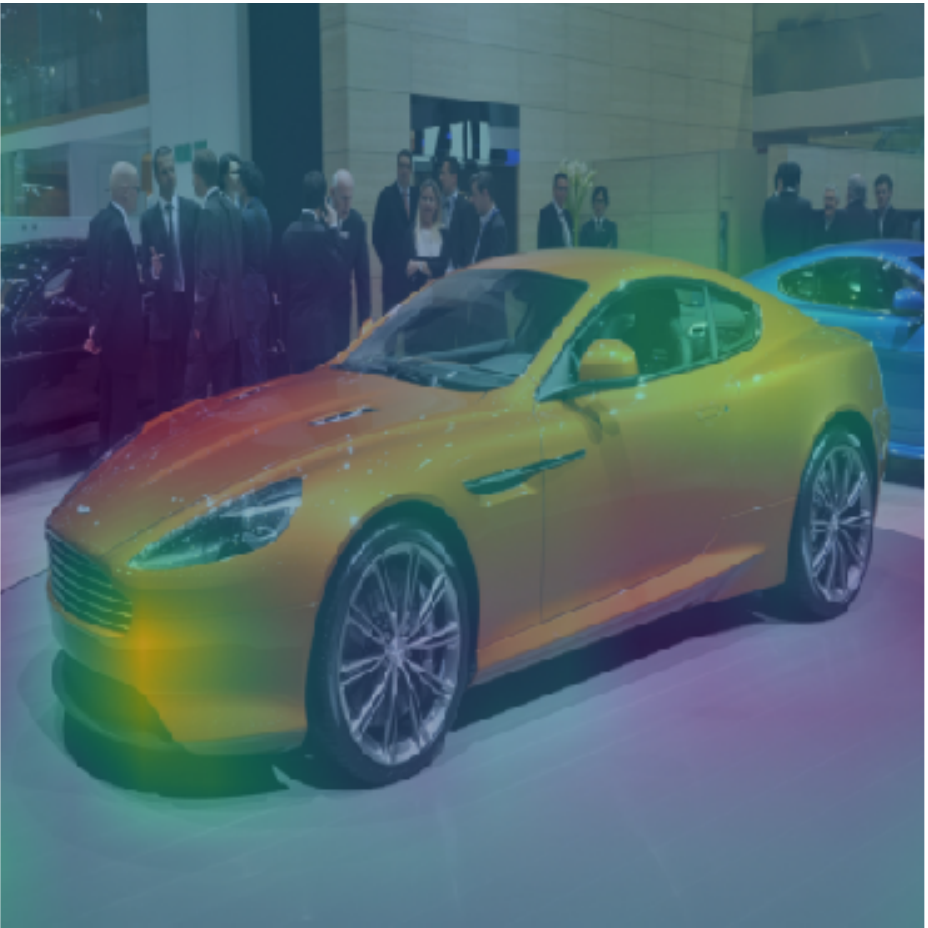}
\includegraphics[width=0.45\linewidth]{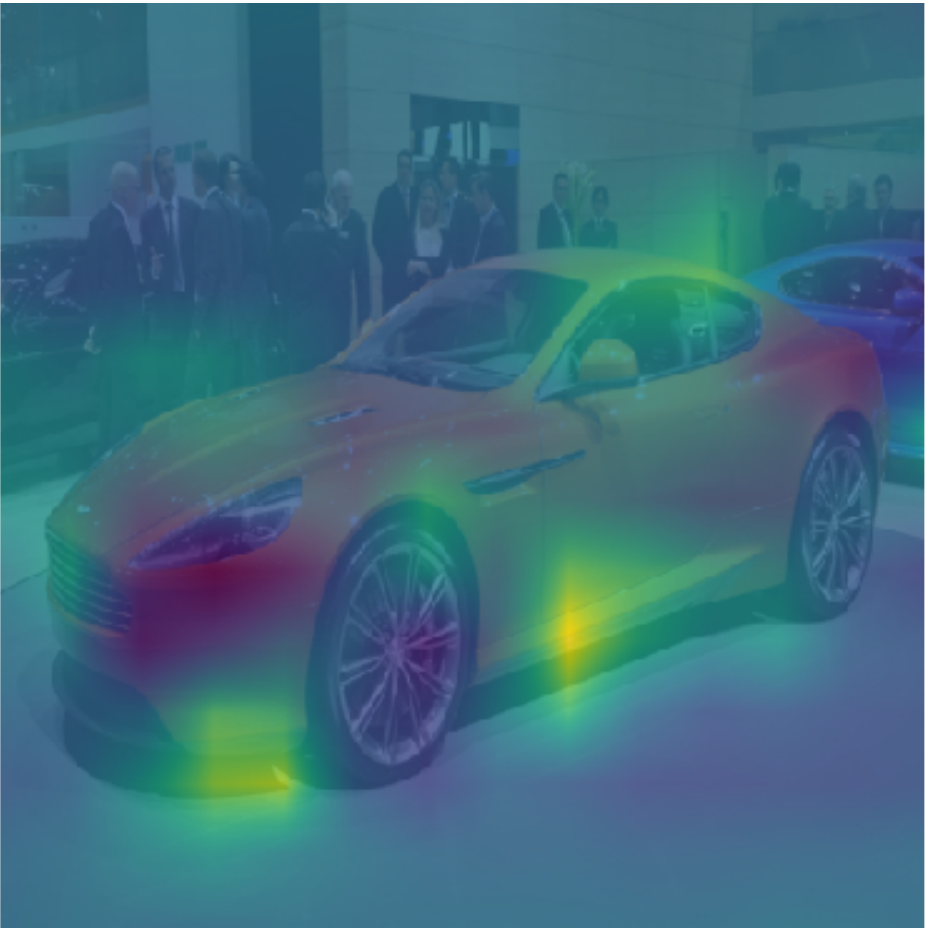}\\
\includegraphics[width=0.45\linewidth]{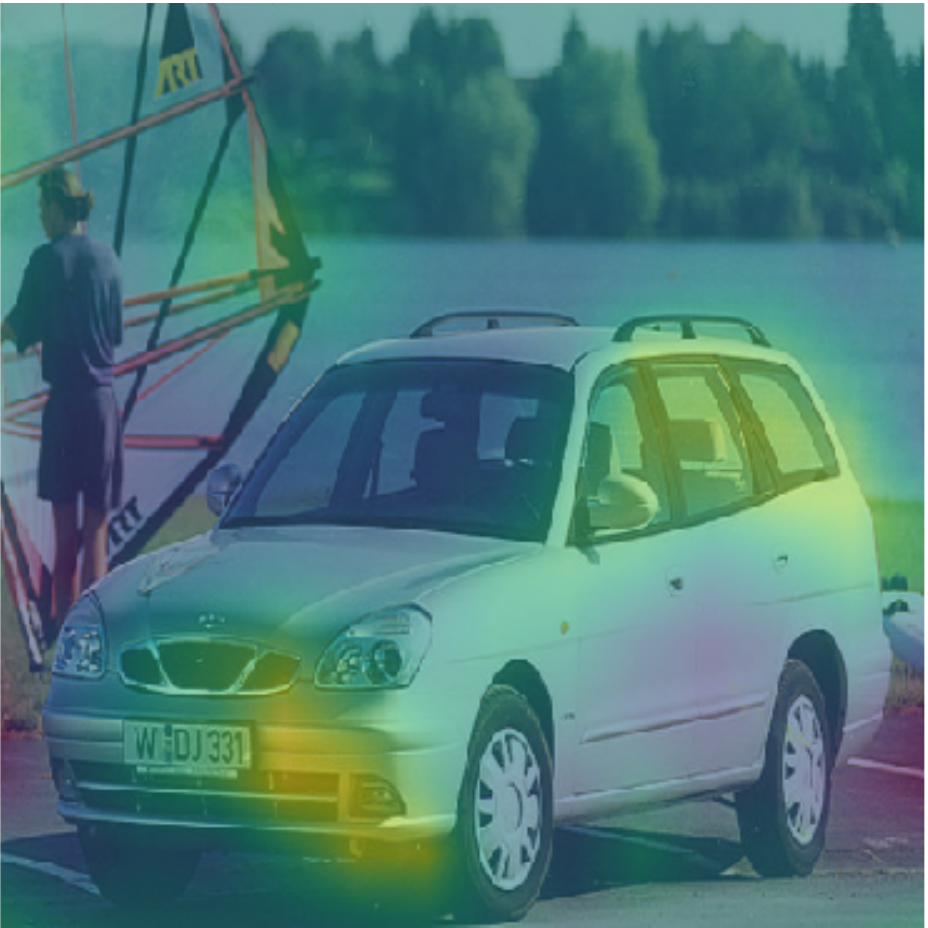}
\includegraphics[width=0.45\linewidth]{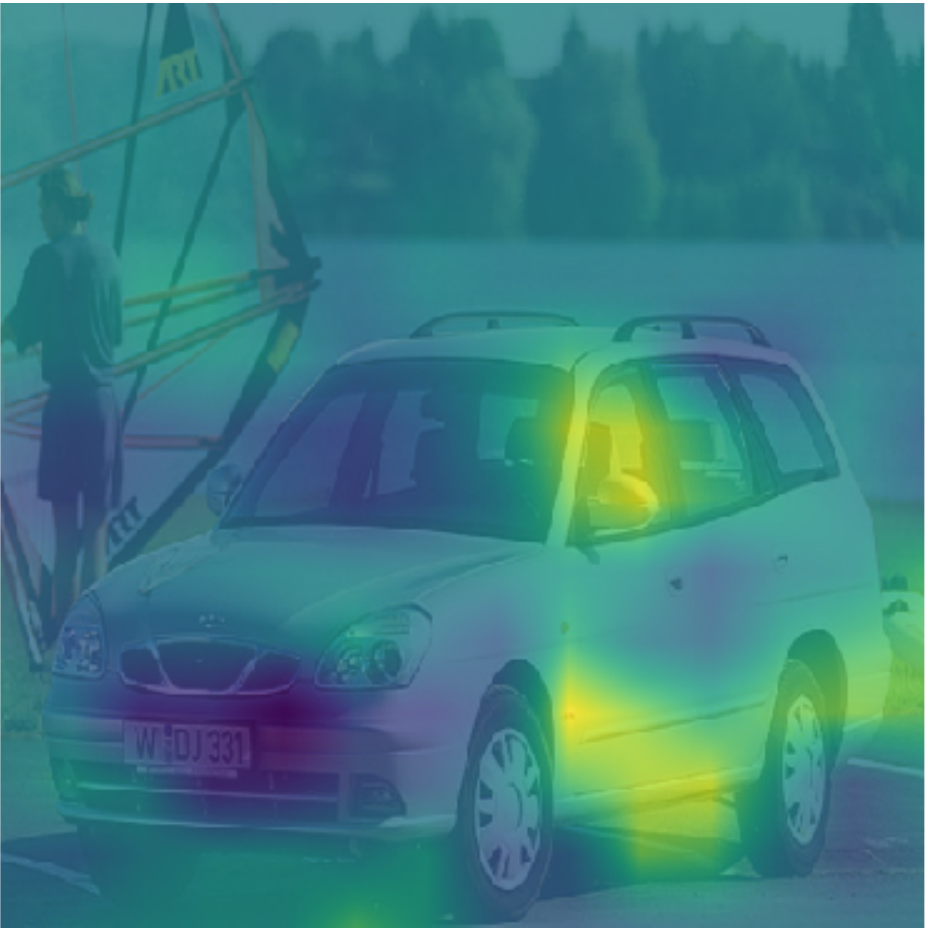}\\
\includegraphics[width=0.45\linewidth]{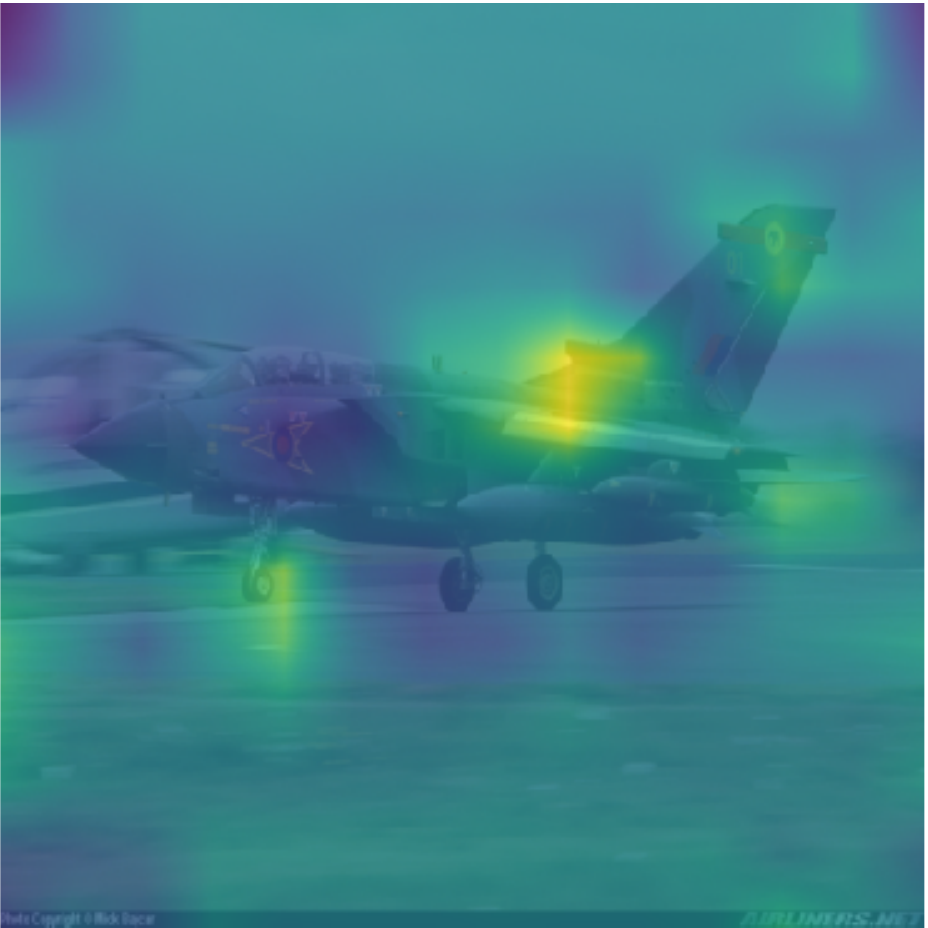}
\includegraphics[width=0.45\linewidth]{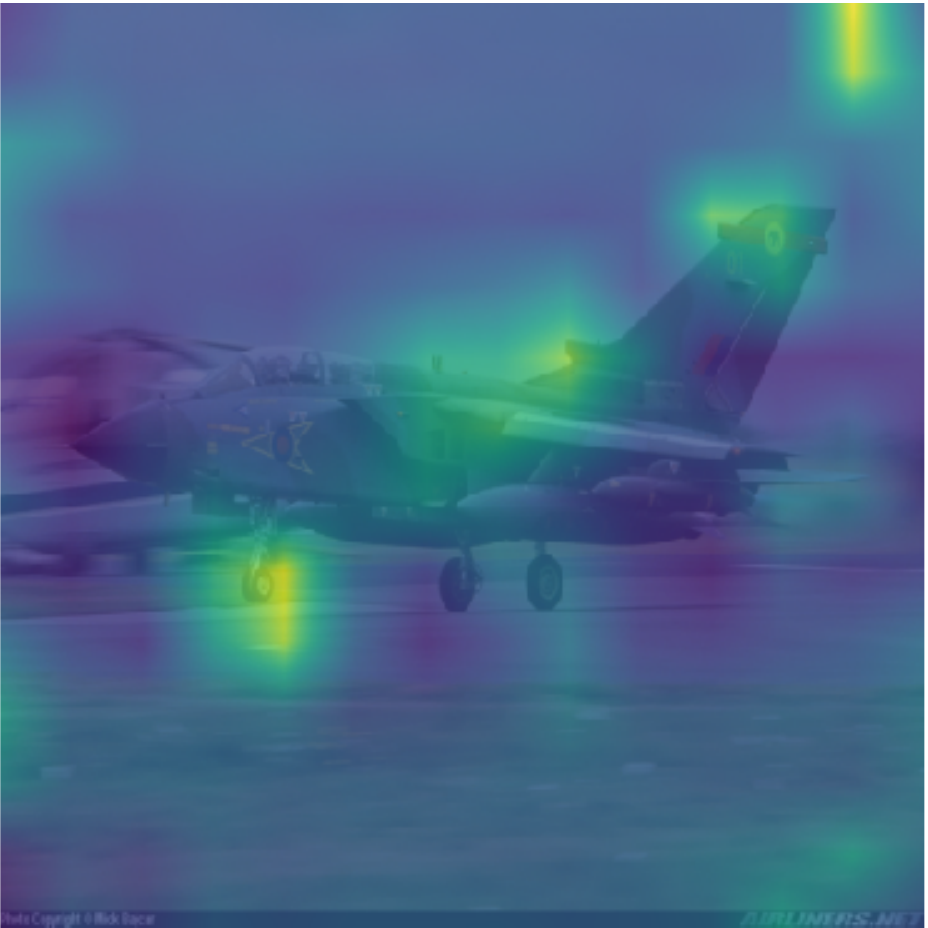}\\
\includegraphics[width=0.45\linewidth]{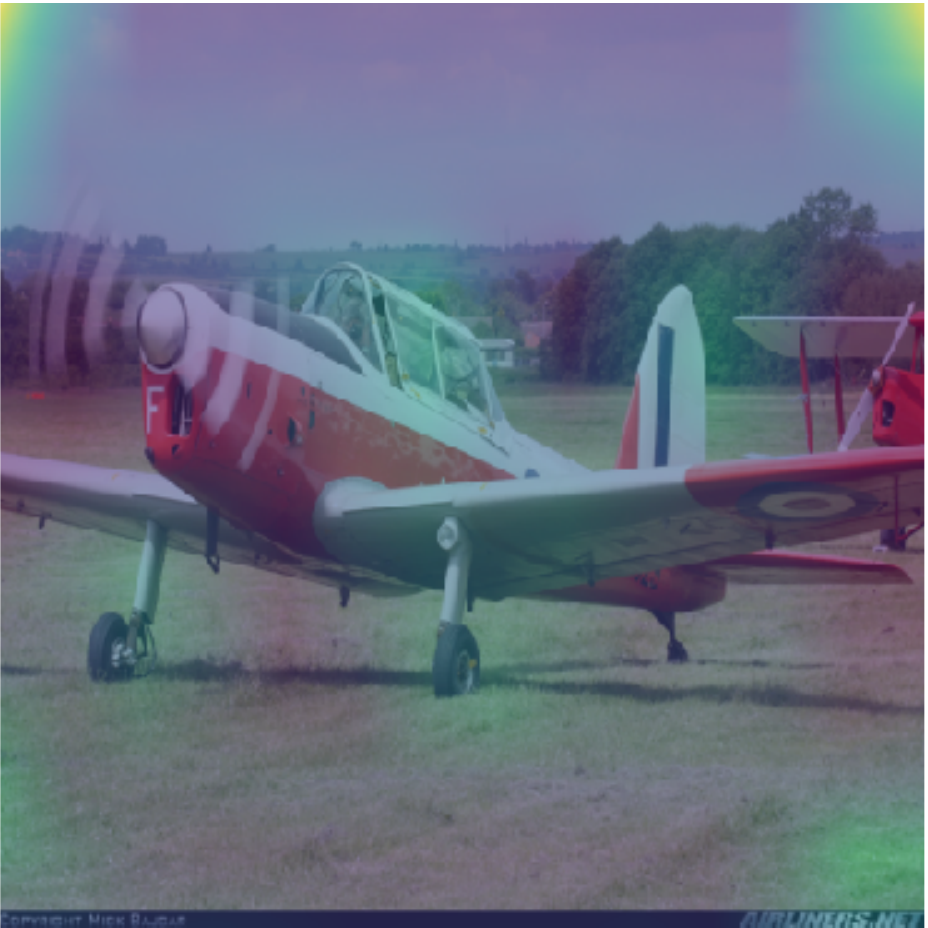}
\includegraphics[width=0.45\linewidth]{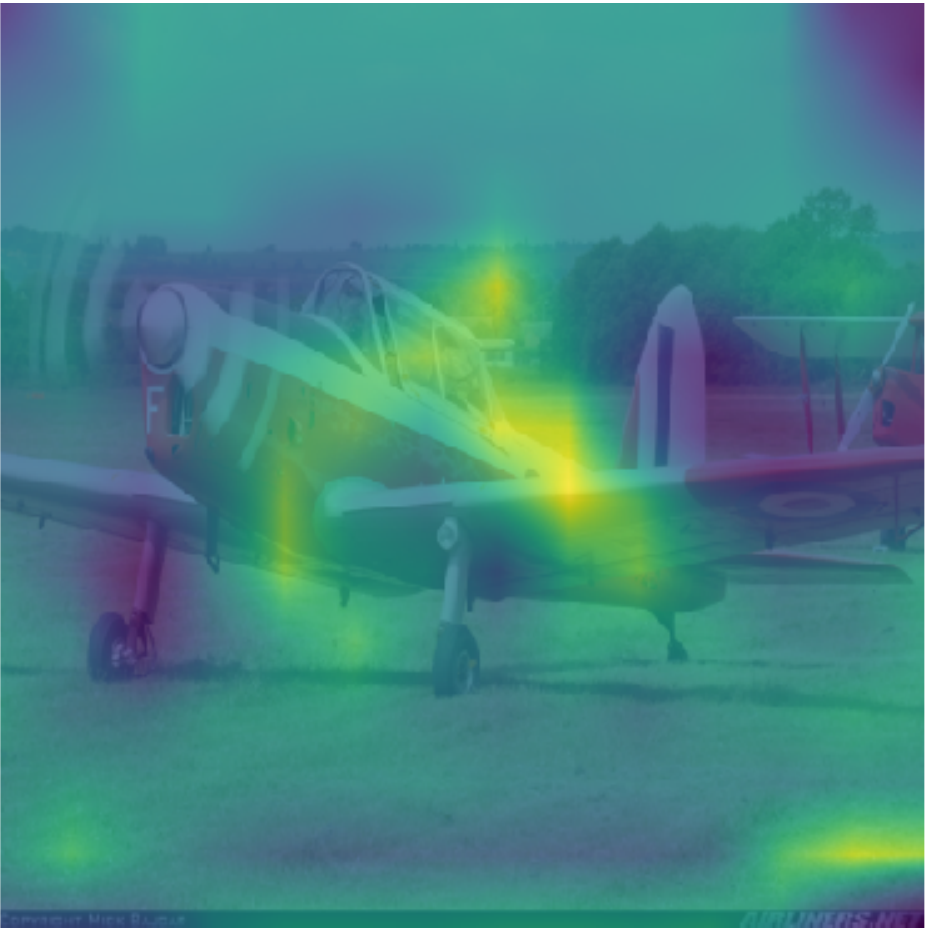}\\
\centering (c)
\end{minipage}
\begin{minipage}{0.21\linewidth}
\includegraphics[width=0.45\linewidth]{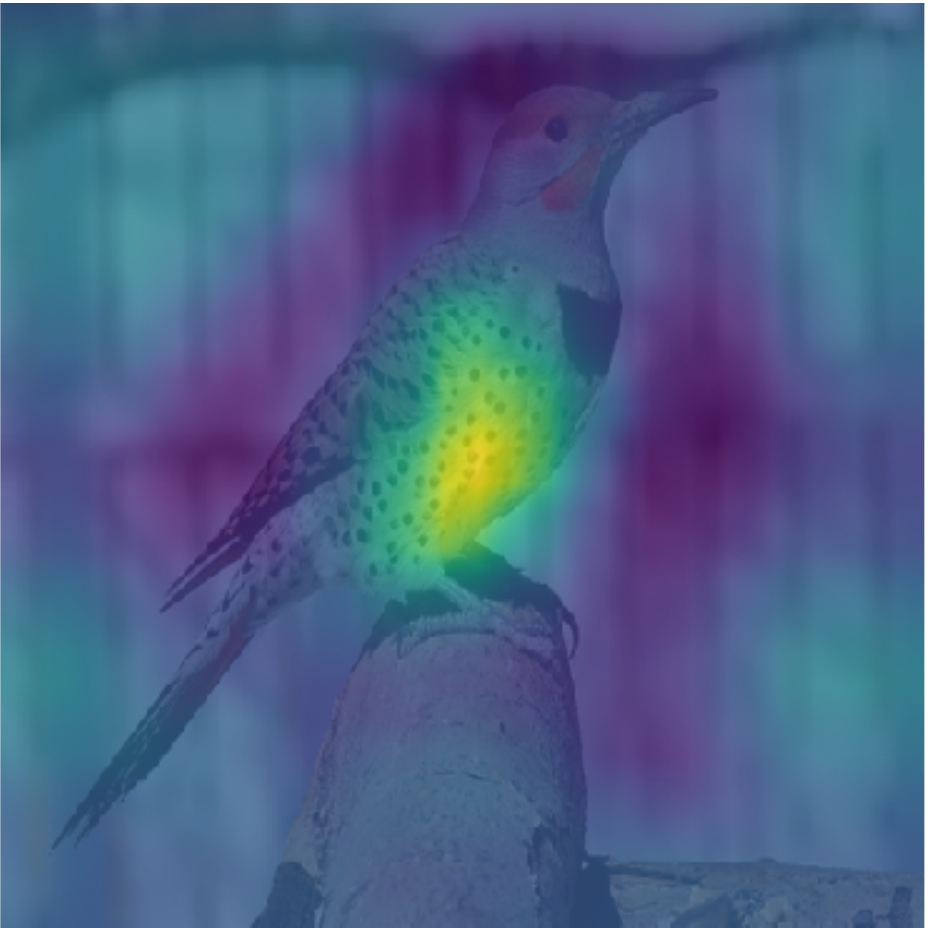}
\includegraphics[width=0.45\linewidth]{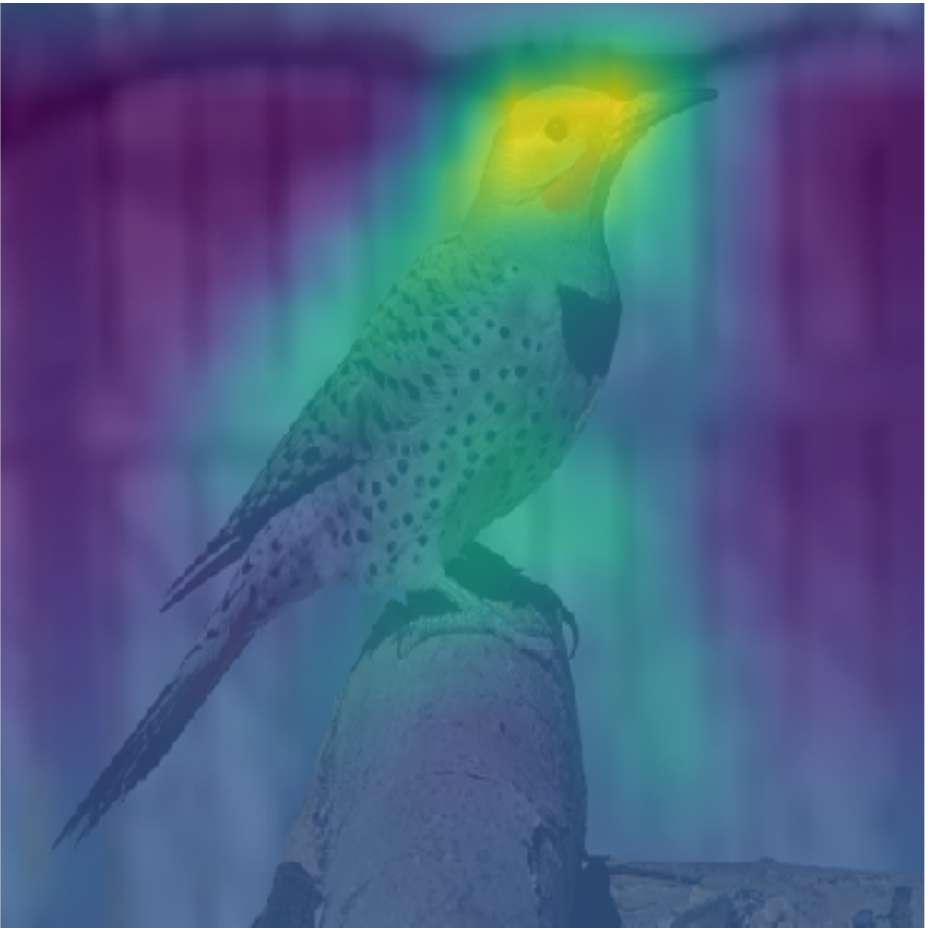}\\
\includegraphics[width=0.45\linewidth]{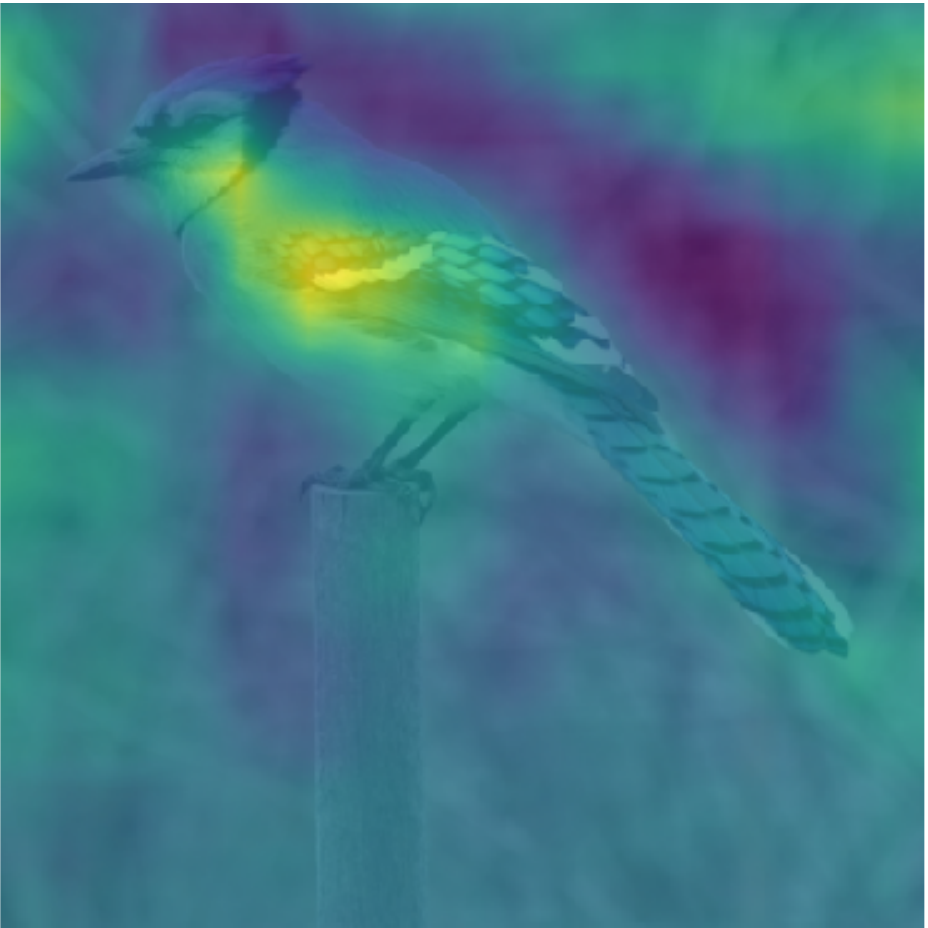}
\includegraphics[width=0.45\linewidth]{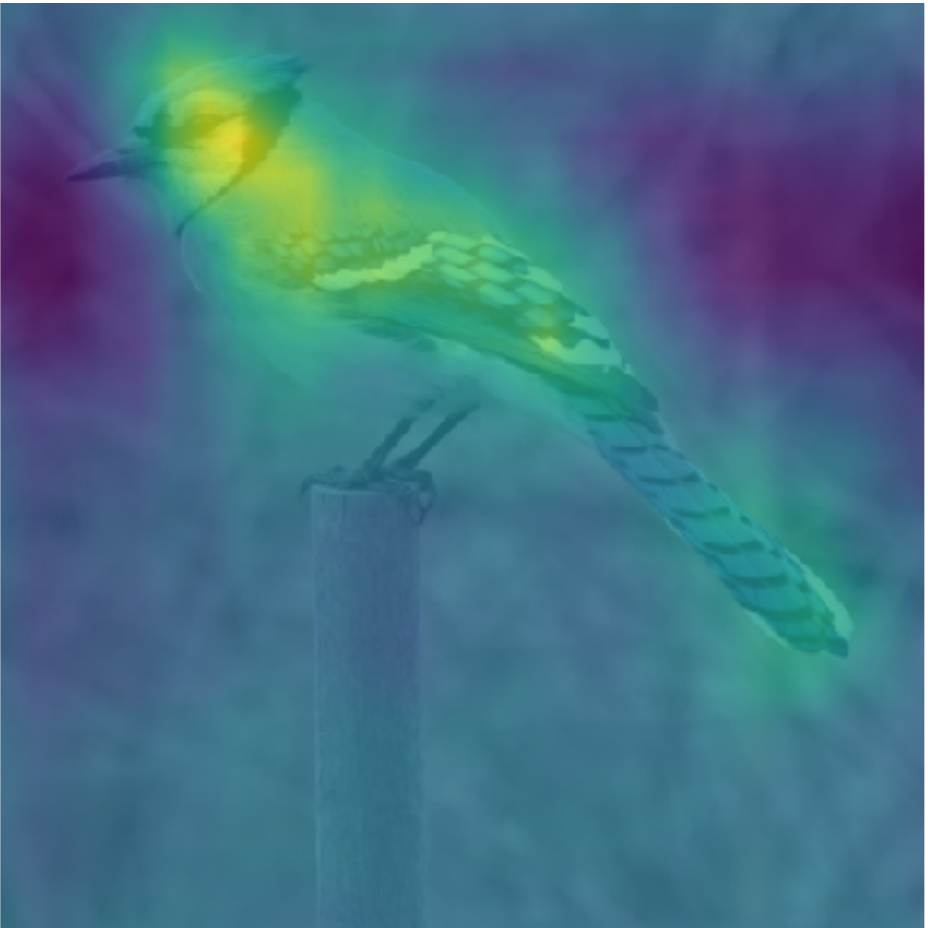}\\
\includegraphics[width=0.45\linewidth]{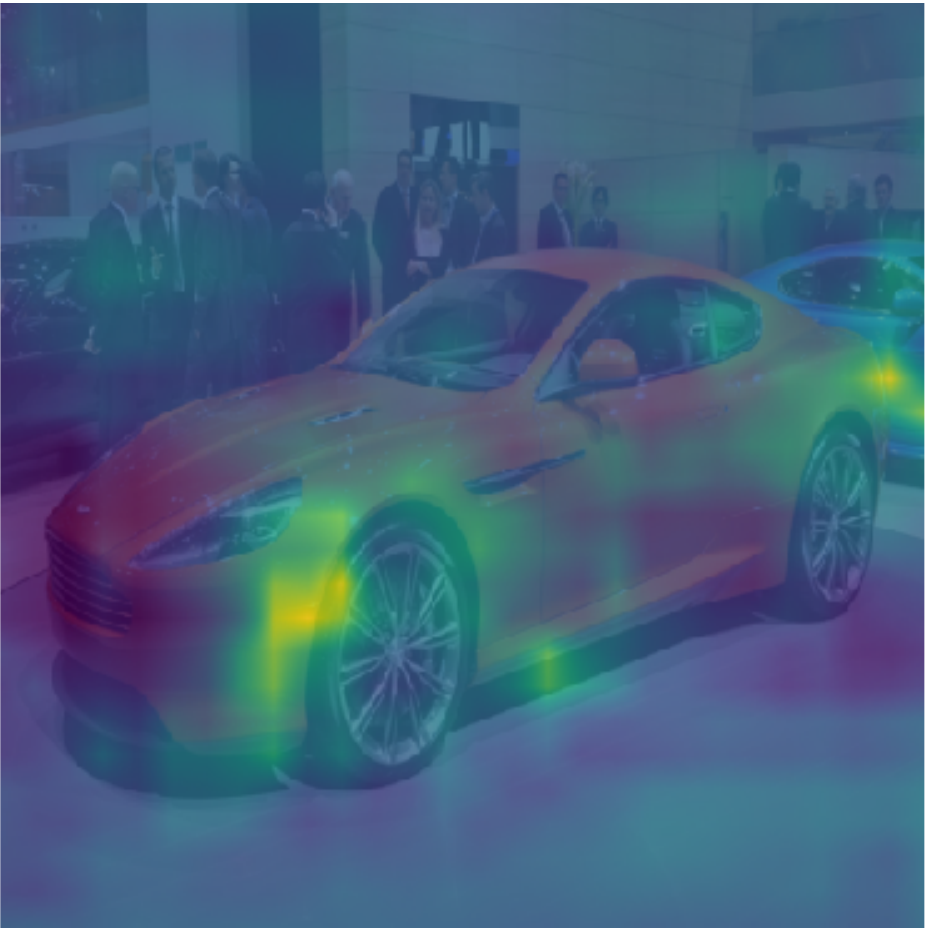}
\includegraphics[width=0.45\linewidth]{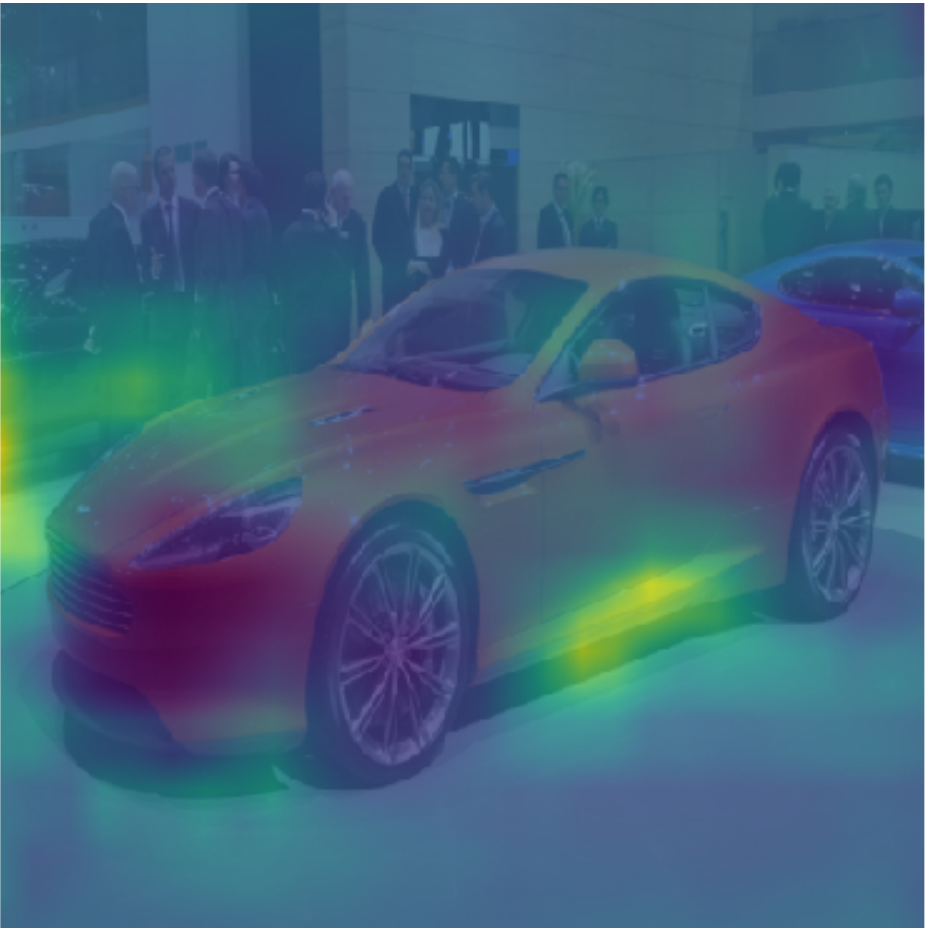}\\
\includegraphics[width=0.45\linewidth]{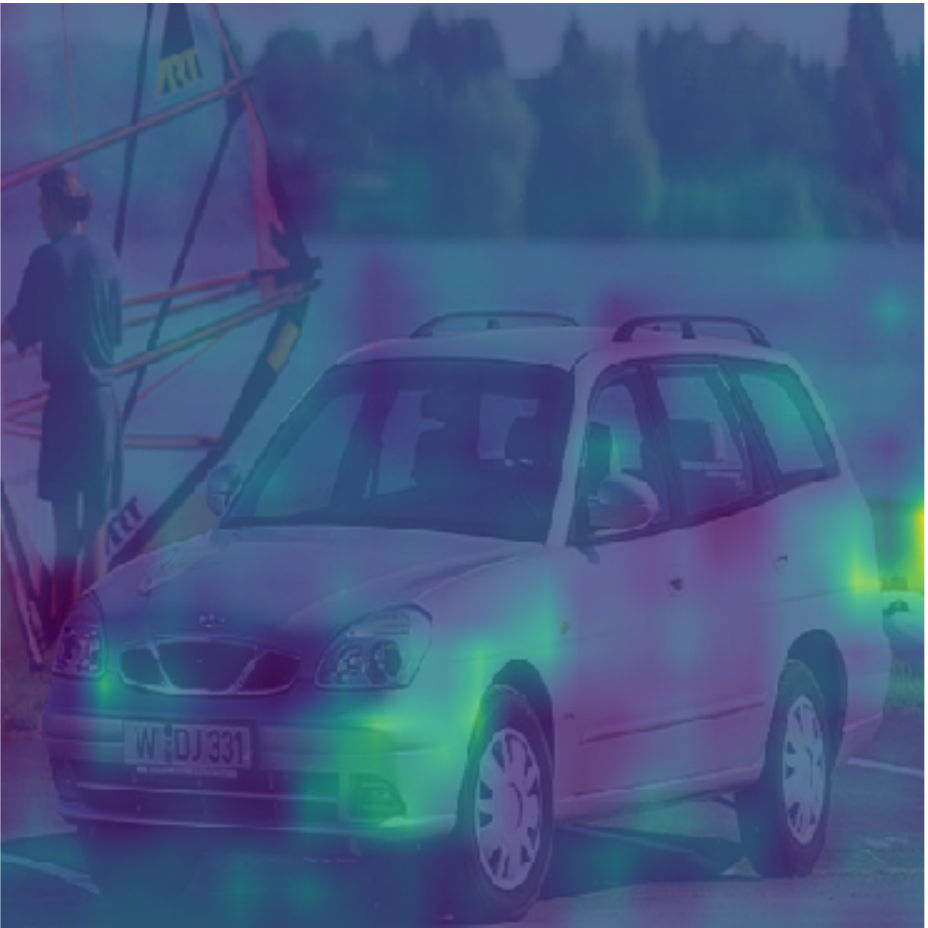}
\includegraphics[width=0.45\linewidth]{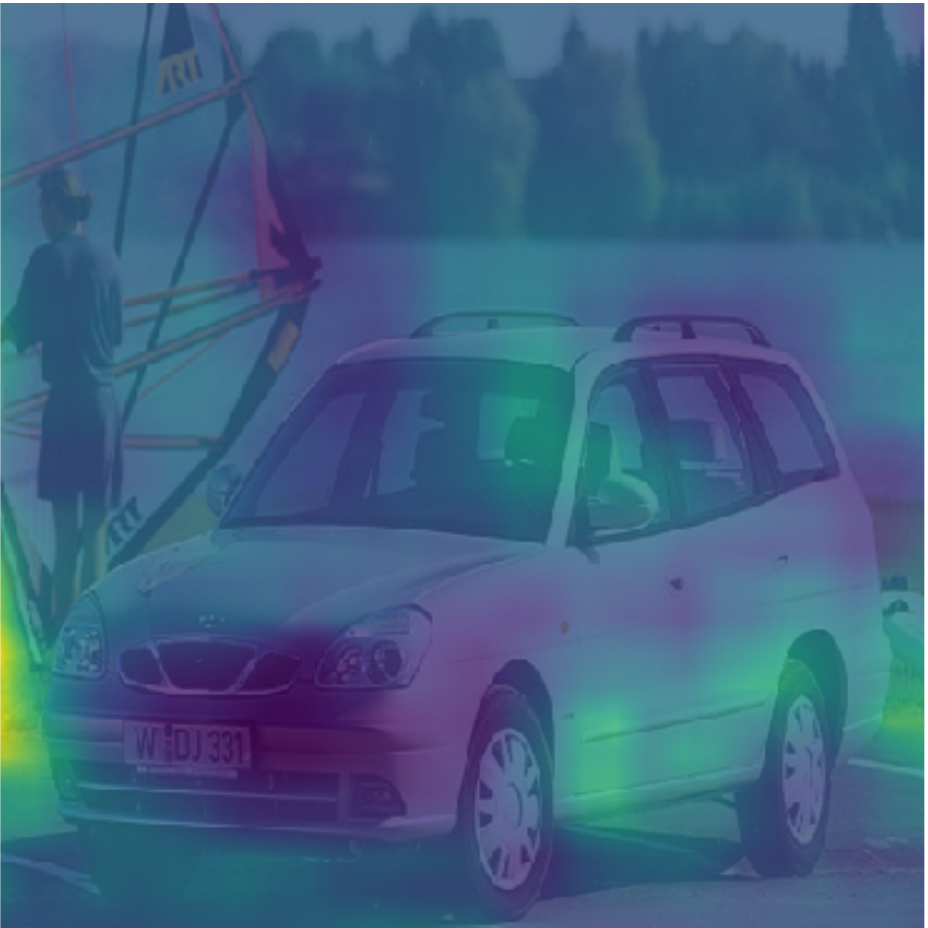}\\
\includegraphics[width=0.45\linewidth]{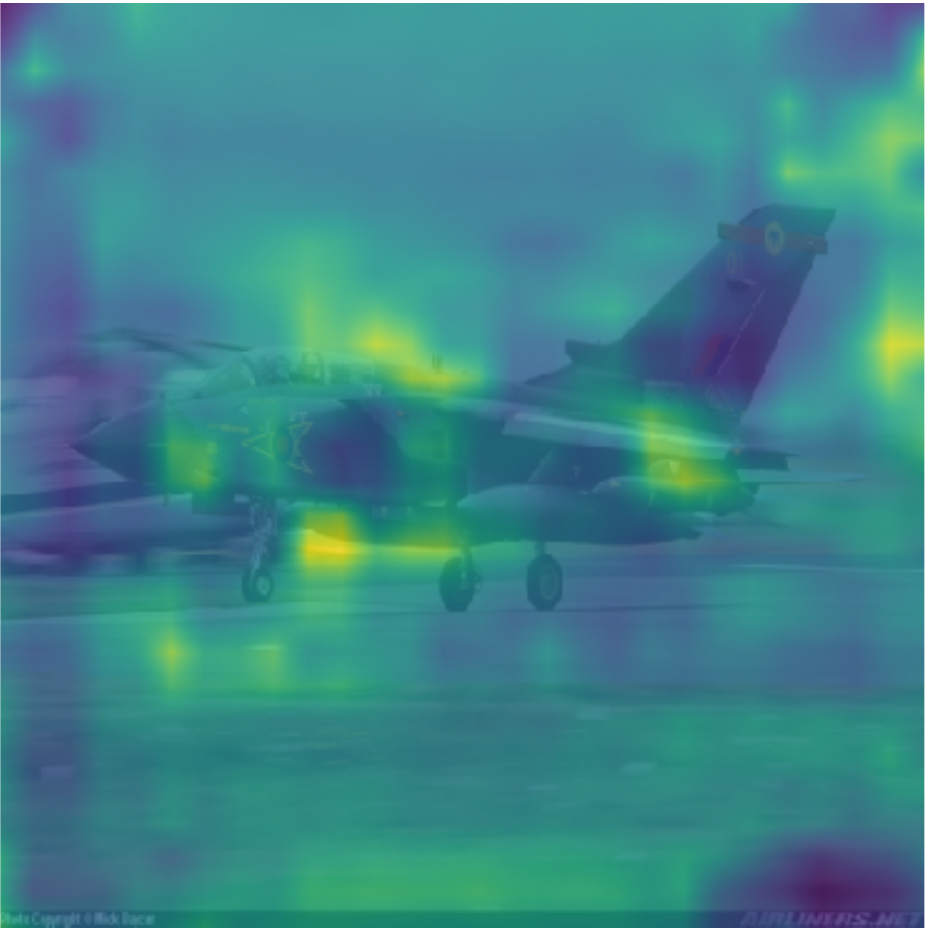}
\includegraphics[width=0.45\linewidth]{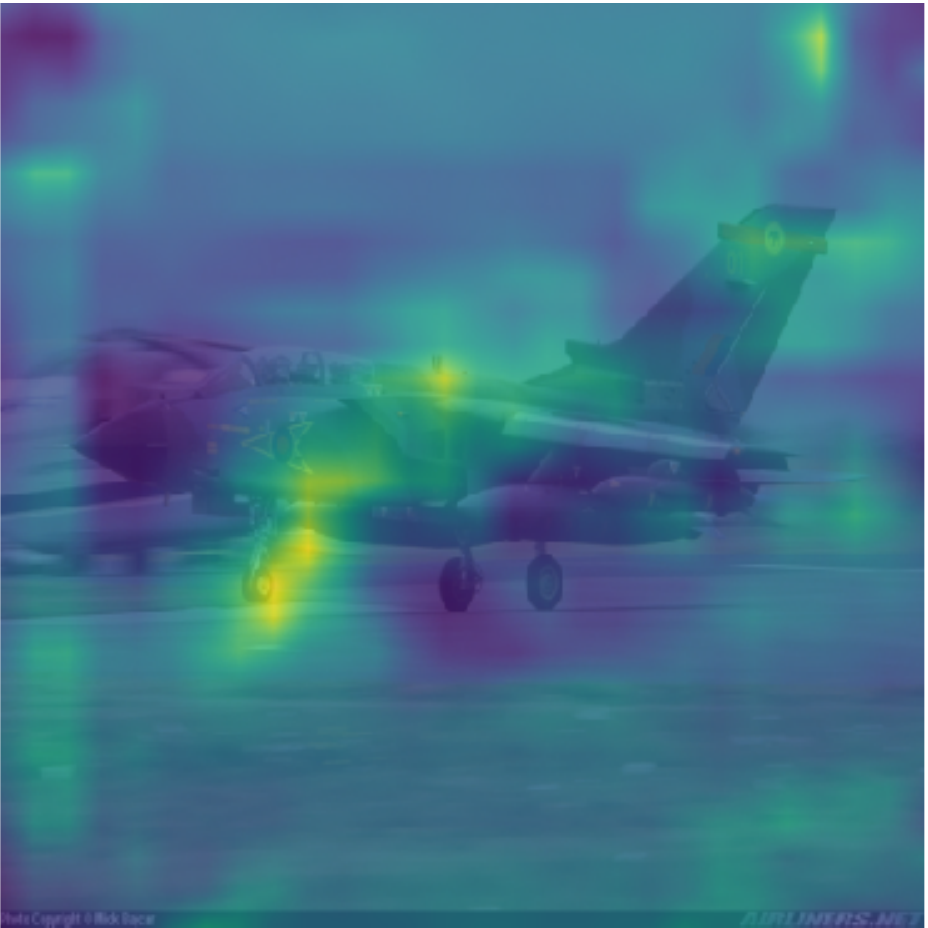}\\
\includegraphics[width=0.45\linewidth]{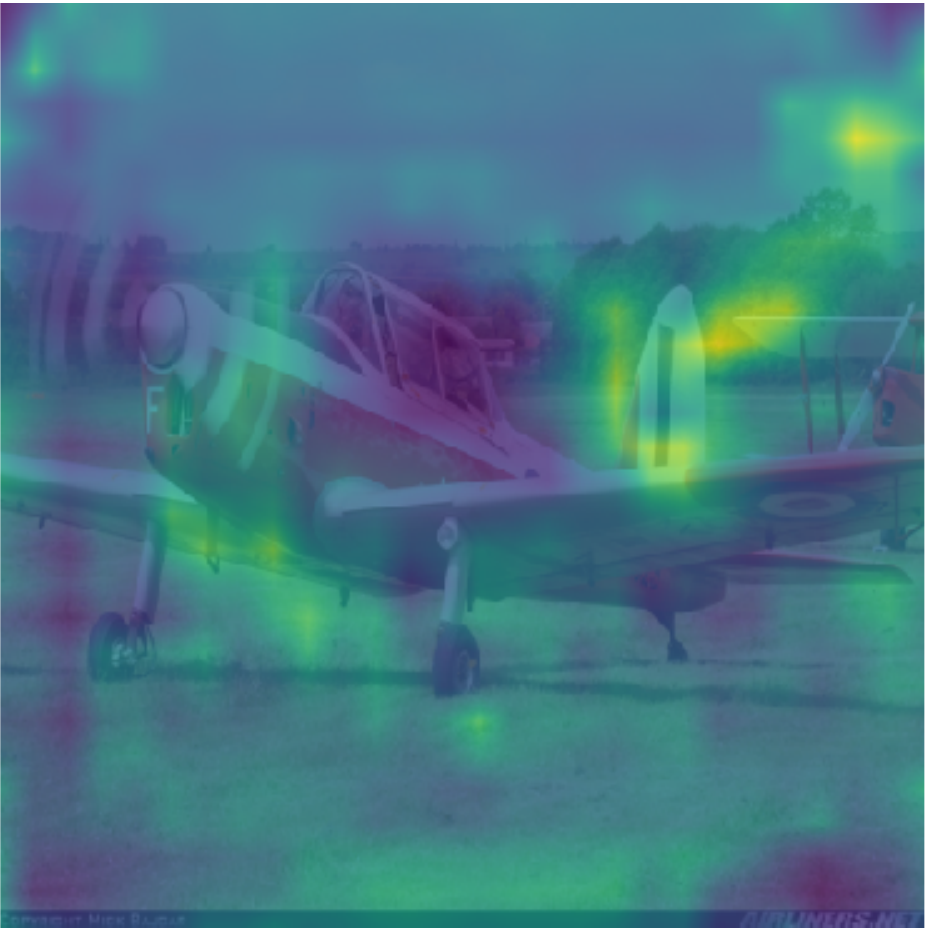}
\includegraphics[width=0.45\linewidth]{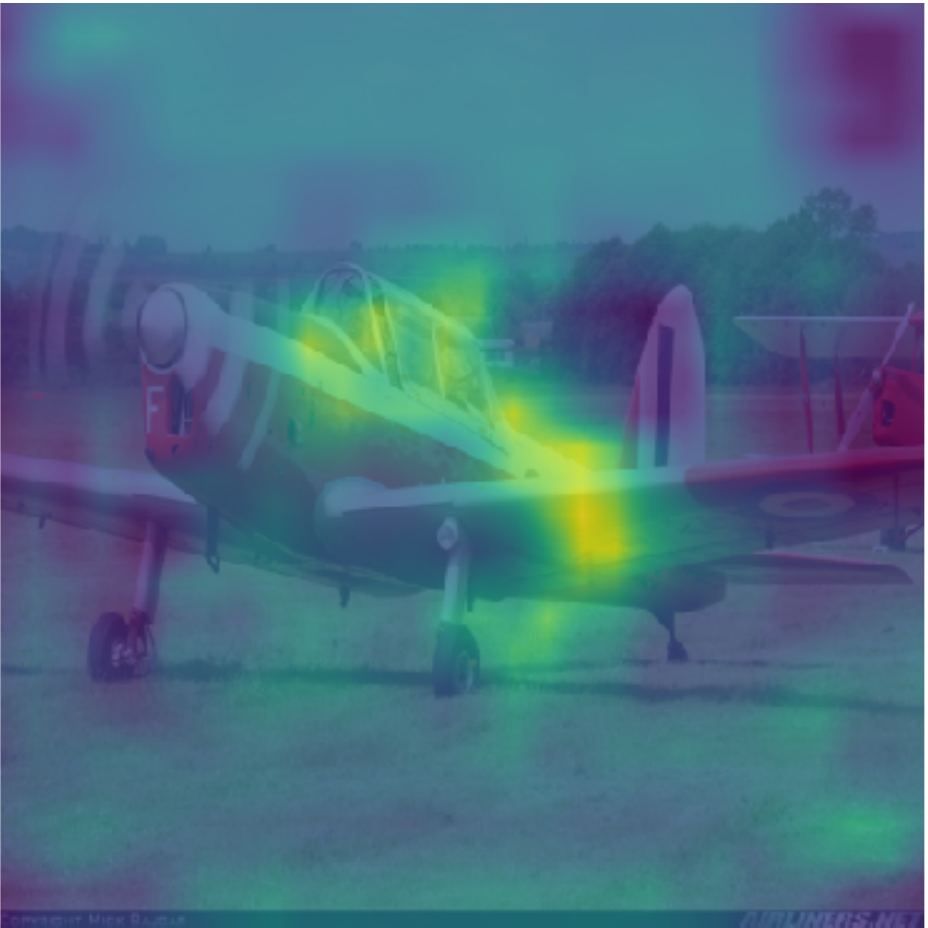}\\
\centering (d)
\end{minipage}
\begin{minipage}{0.21\linewidth}
\includegraphics[width=0.45\linewidth]{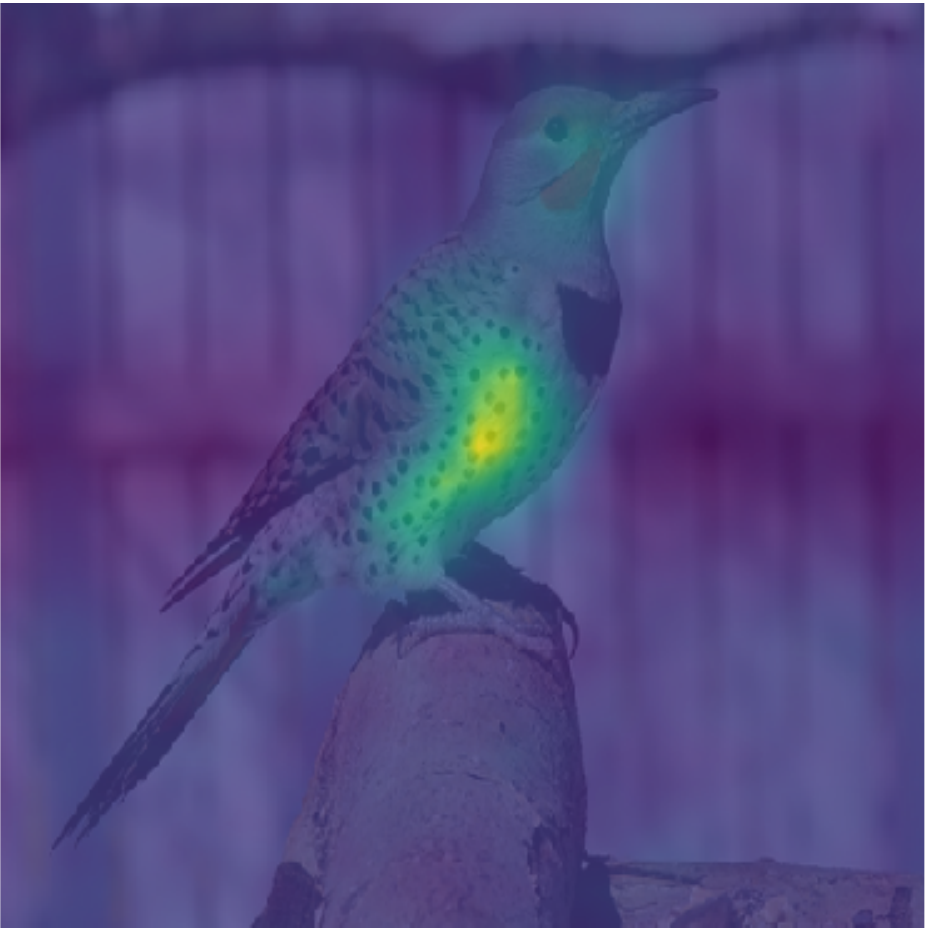}
\includegraphics[width=0.45\linewidth]{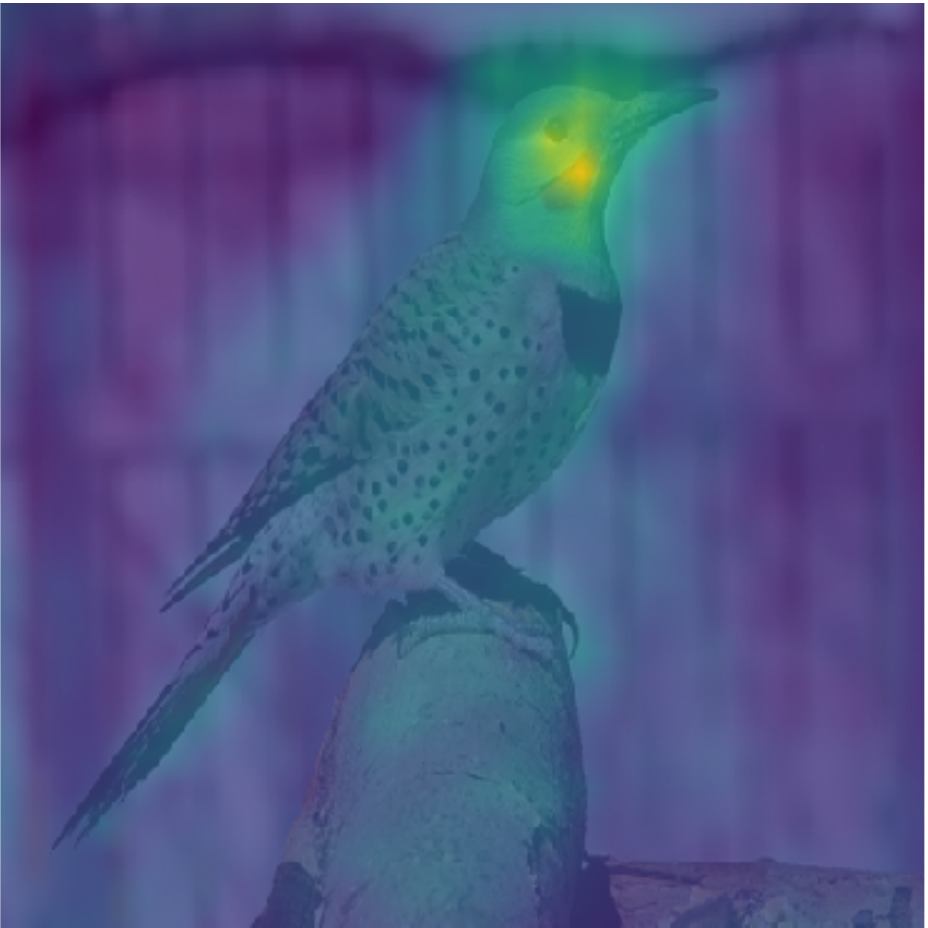}\\
\includegraphics[width=0.45\linewidth]{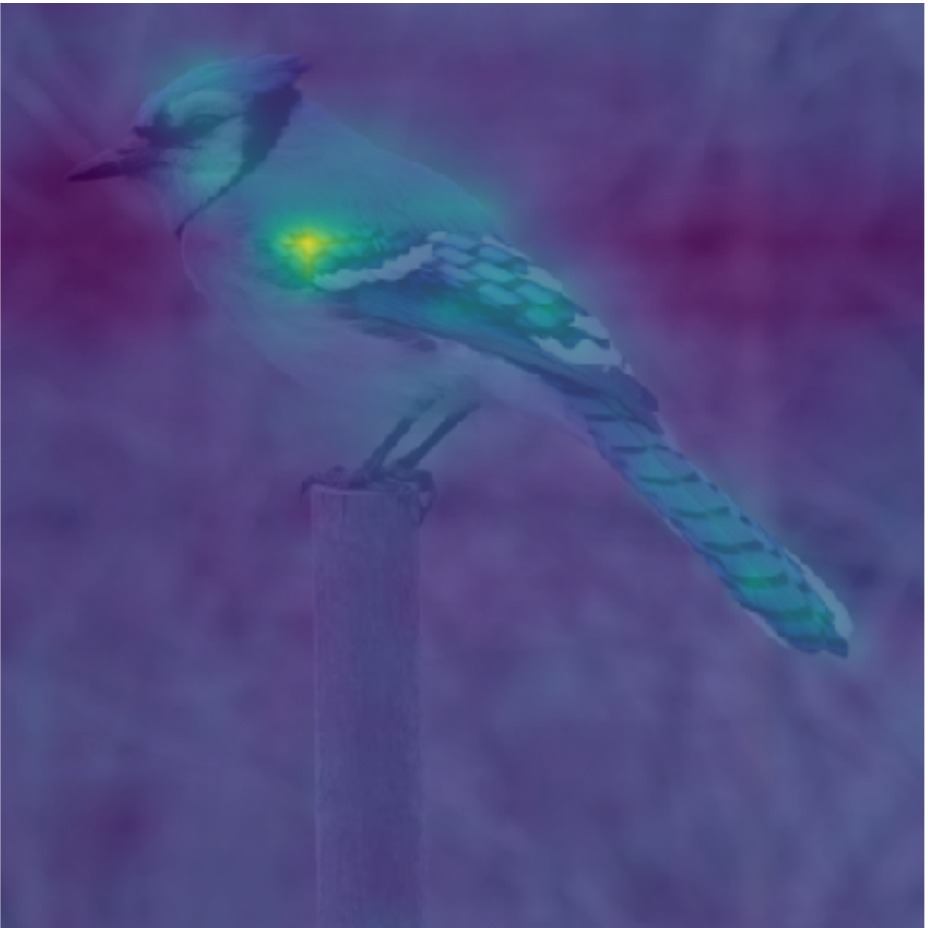}
\includegraphics[width=0.45\linewidth]{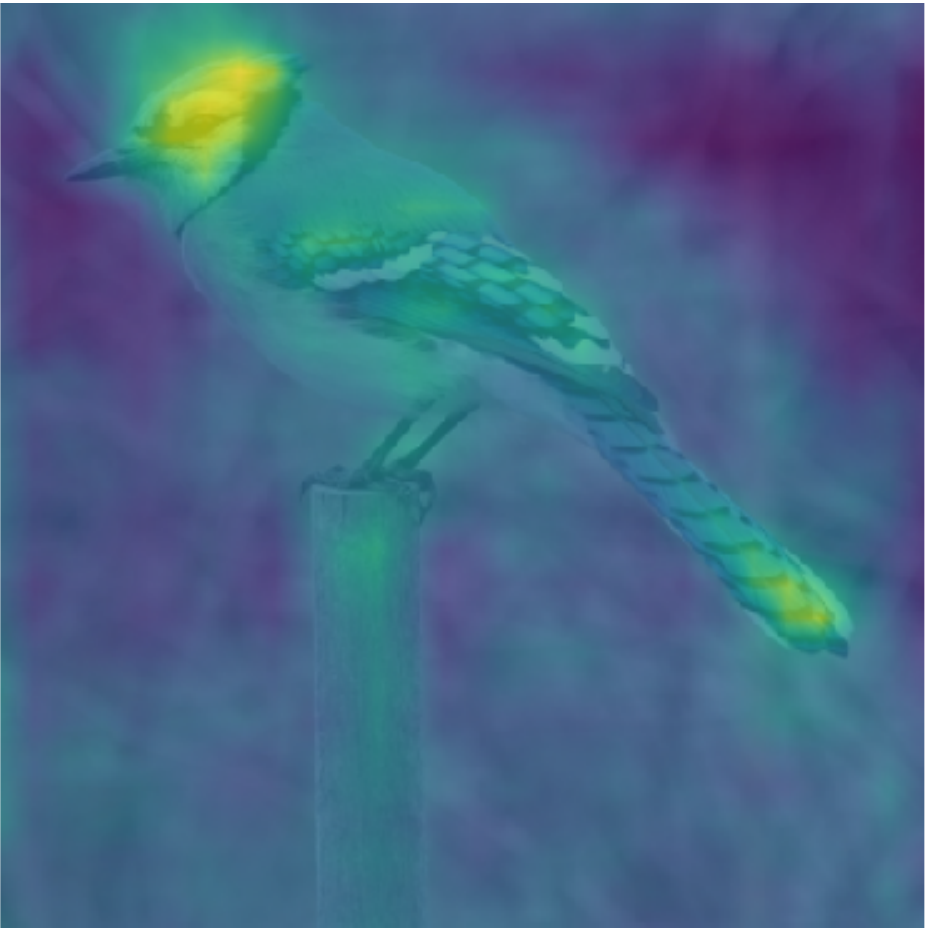}\\
\includegraphics[width=0.45\linewidth]{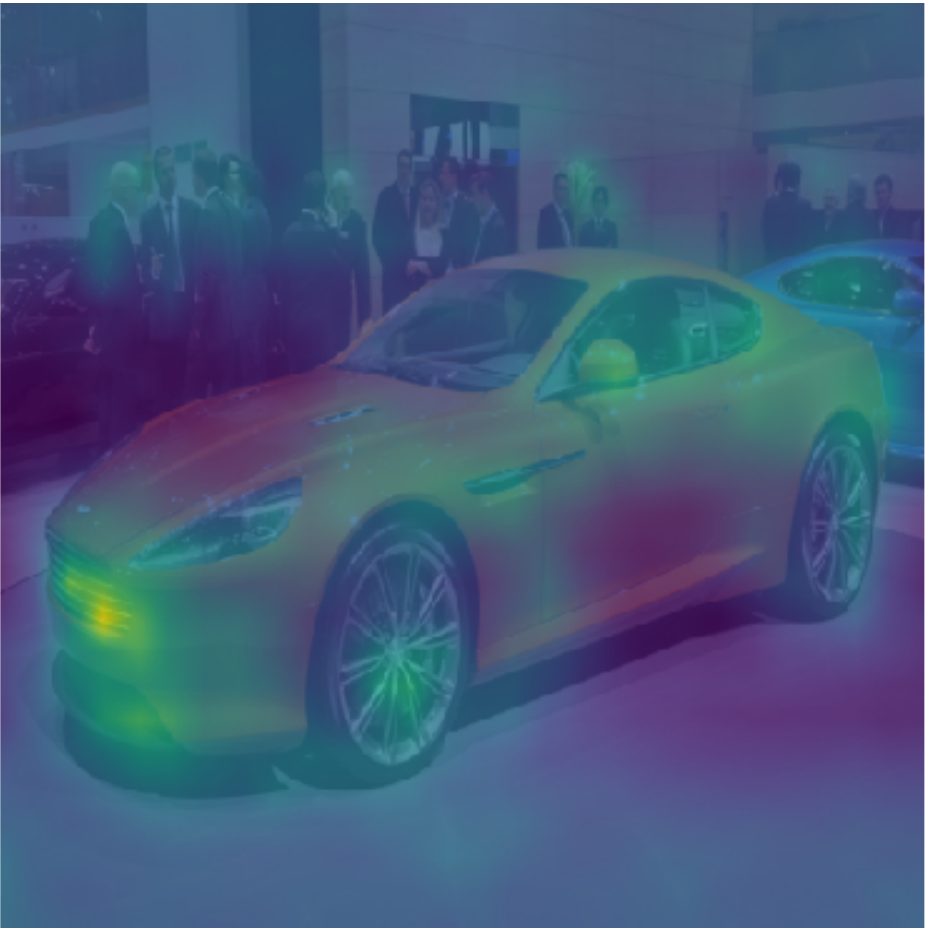}
\includegraphics[width=0.45\linewidth]{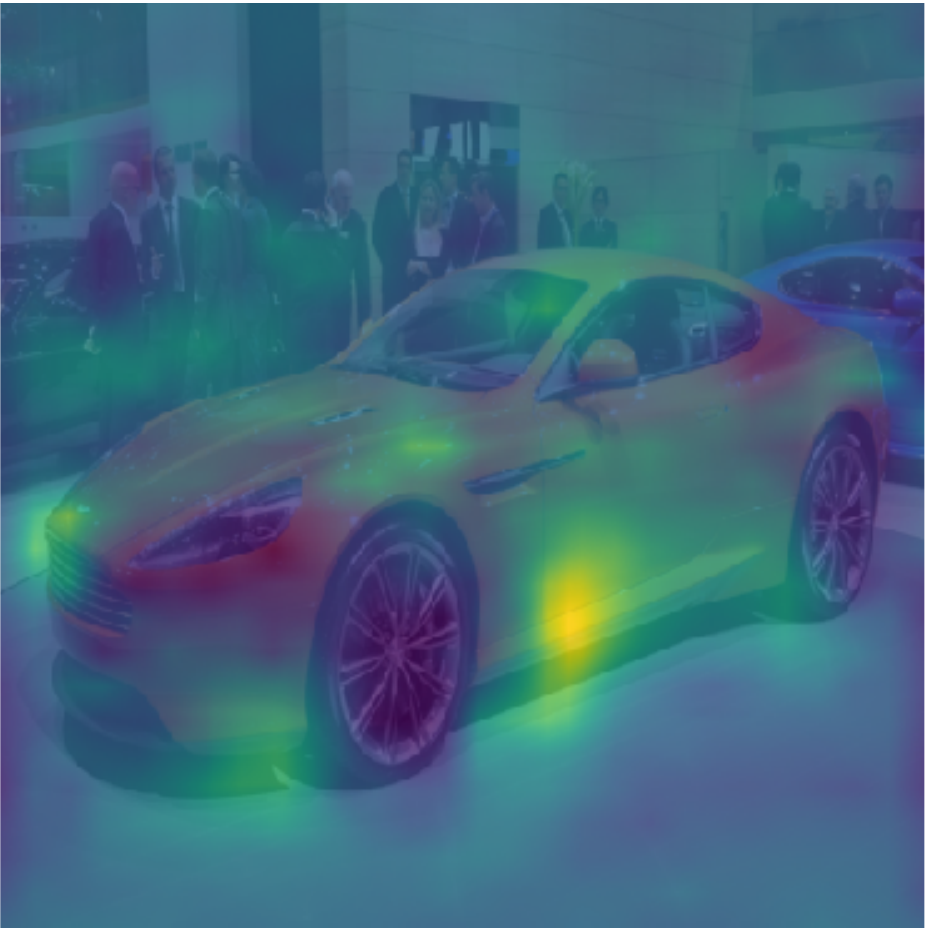}\\
\includegraphics[width=0.45\linewidth]{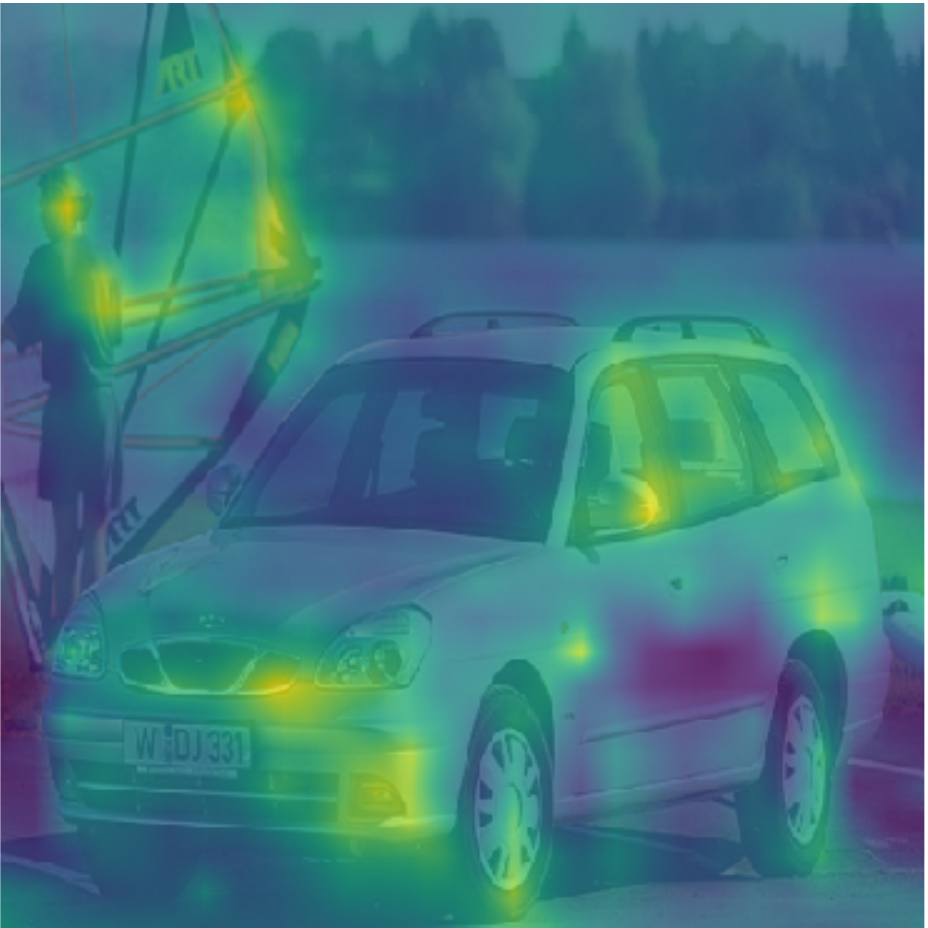}
\includegraphics[width=0.45\linewidth]{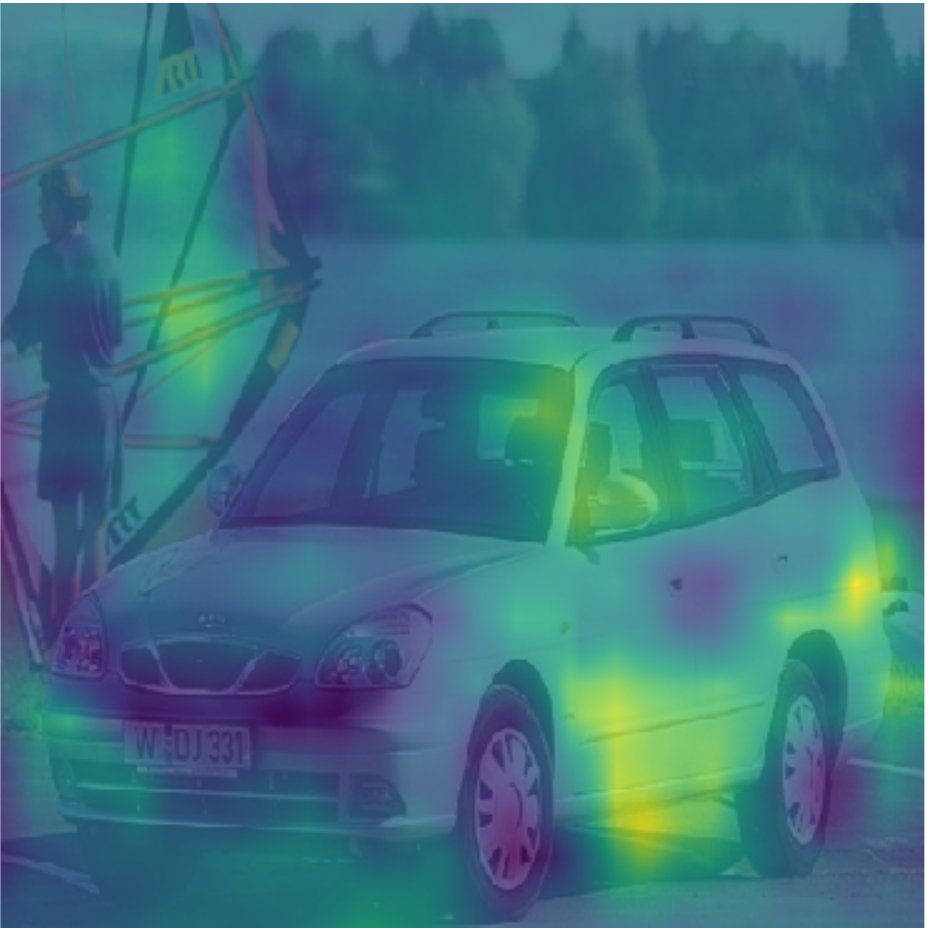}\\
\includegraphics[width=0.45\linewidth]{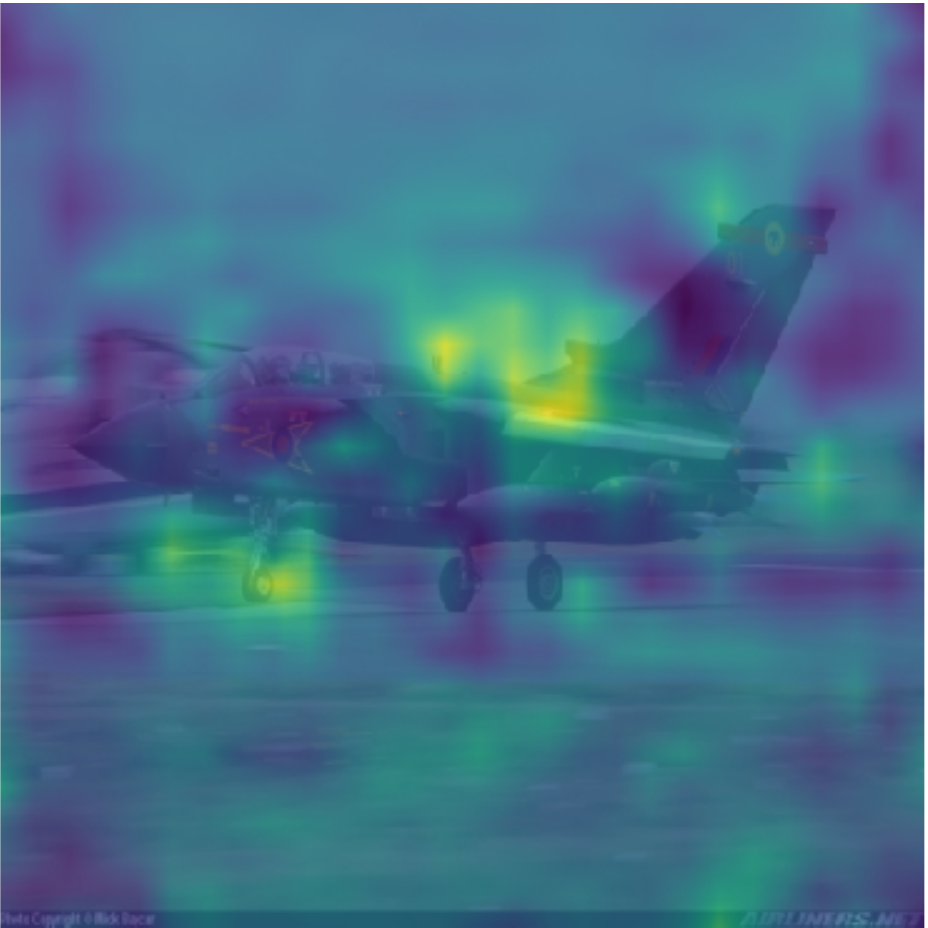}
\includegraphics[width=0.45\linewidth]{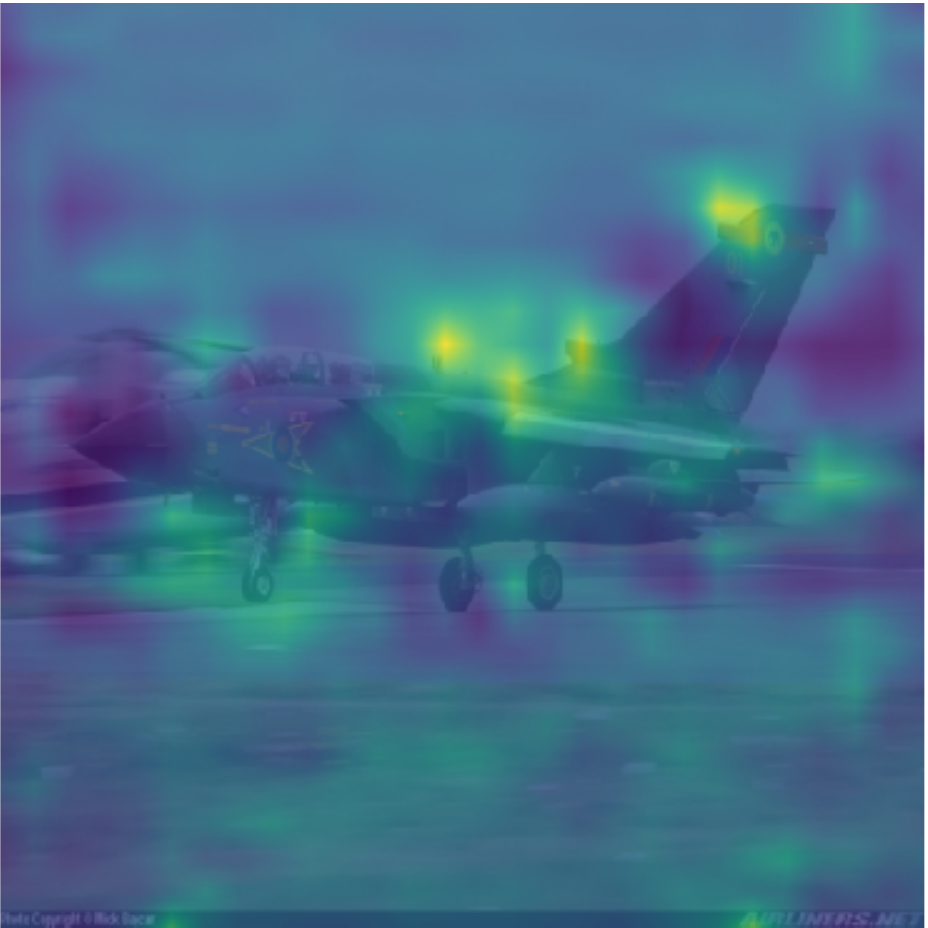}\\
\includegraphics[width=0.45\linewidth]{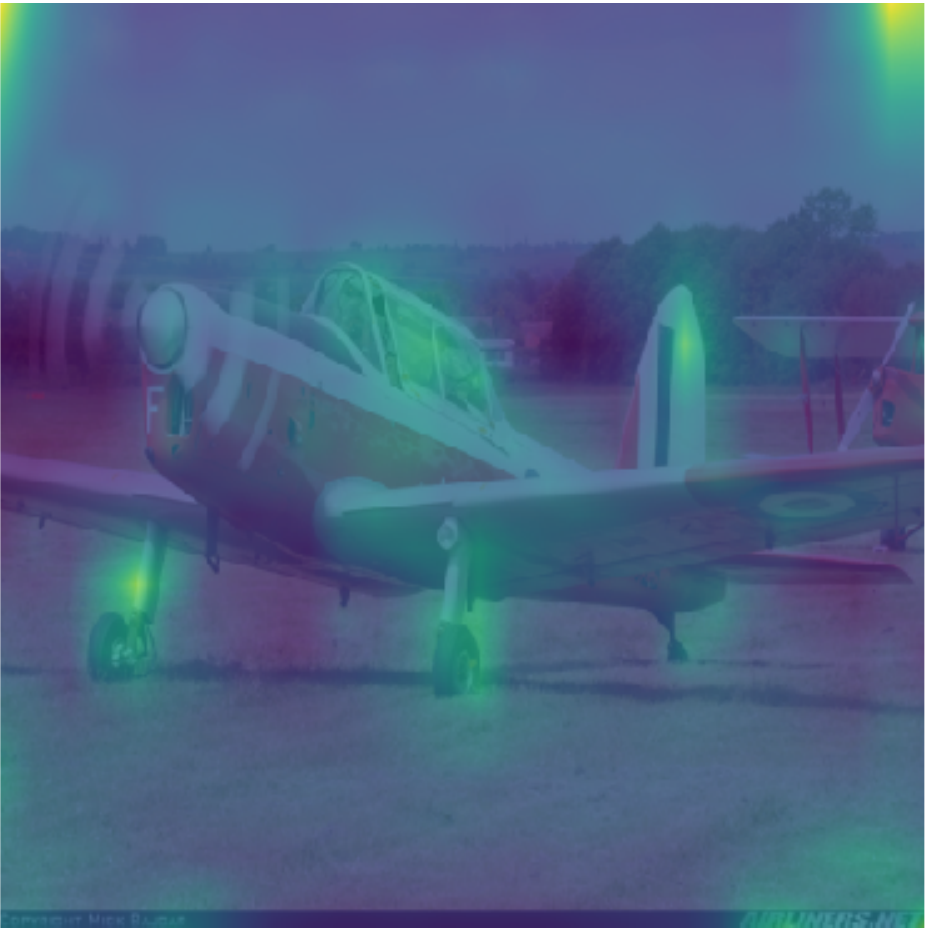}
\includegraphics[width=0.45\linewidth]{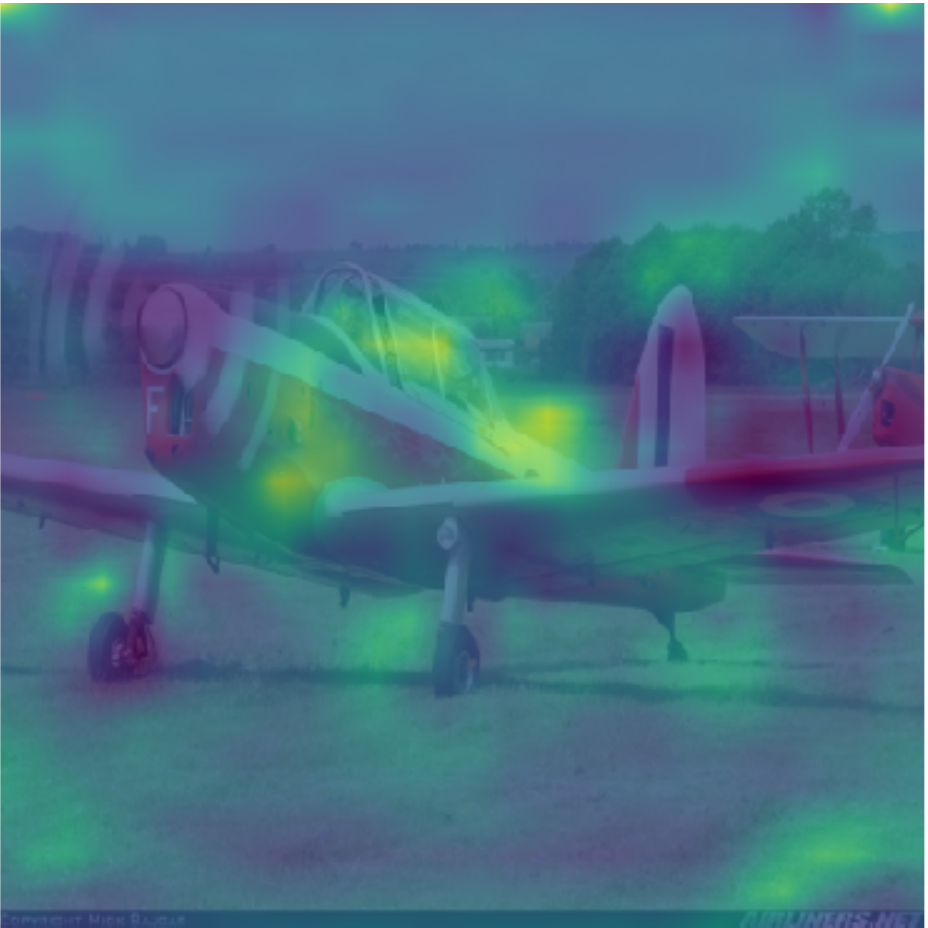}\\
\centering (e)
\end{minipage}
\end{center}
\caption{Superimposed display of activation maps (b) $\mathbf{U}_p^{L-1}$, (c) $\mathbf{U}_p^L$ and (d) $\mathbf{U}_p^G$  from CUB-Birds, Stanford Cars, and FGVC-Aircraft. The first column (a) shows original images and the last two columns (e) are combined activation maps from corresponding columns of $\mathbf{U}_p^{L-1}$, $\mathbf{U}_p^L$ and $\mathbf{U}_p^G$ . Each of (b)$\sim$(e) shows activations of two excitation modules in the corresponding layers. Best viewed in color.} 
\label{fig:acm}
\end{figure*}

\textbf{GMP vs. GAP:} As indicated in Fig.~\ref{fig:diagram}, GAP and global max pooling (GMP) can be alternatively adopted to pool feature maps. However, we only switch the pooling method from GAP to GMP in $\mathbf{U}^{L-1}$, since we initially thought the discriminative structure of FGVC is local and subtle, thus GMP should have advantage over GAP to capture the these structures, and provide a better feature representation. This is verified on almost all datasets (the top group of Tab.~\ref{tab:rslt-gmpgap}). The results indicate $CL$ can collaborate well with GMP to provide robust mid-level features. However, when the network is enhanced by the merged feature maps, which use GAP, but without $CL$, the results show different behaviour (the bottom group of Tab.~\ref{tab:rslt-gmpgap}). GAP{\scriptsize$+$}, where GAP is employed on $\mathbf{U}^{L-1}$, achieves the best performance on Cars, Dogs, and Aircraft but fails to surpass the performance of GMP{\scriptsize$-$}, where GMP is employed on $\mathbf{U}^{L-1}$, on Birds. This phenomenon indicates that GMP is necessary for ascertaining local and subtle structures in categories with fine-and-rich texture. The difference caused by employing GMP or GAP on $\mathbf{U}^{L-1}$ can also be observed in Fig.~\ref{fig:acm} (b) (see Sec.~\ref{sec:vis}).
% where GMP leads to single position activation (the first two rows) while GAP results in scattered activations (the last 4 rows).
Therefore, we report the final results on Cars, Dogs, and Aircraft with GAP employed on $\mathbf{U}^{L-1}$ while on Birds with GMP employed on $\mathbf{U}^{L-1}$ in Sec.~\ref{sec:rslt}    

%When GAP is used in both $\mathbf{U}^{L-1}$ and $\mathbf{U}^G$This is caused by the different pooling methods used in stage $L-1$ and $\mathbf{U}^G$

% We alternatively report our results with $\verb'GMP'$ and $\verb'GAP'$ in Tab.~\ref{tab:rslt-nabirds}-\ref{tab:rslt-stdogs}. This stems from the different pooling methods employed on $\mathbf{U}_p^{L-1}$ to get $\mathbf{f}_p^{L-1}$. We initially thought the mid-level features are sensitive to local structures, thus $\verb'GMP'$ should have advantages than $\verb'GAP'$ to provide a better feature representation. This viewpoint is verified on almost all datasets except for Stanford Dogs (the top group of Tab.~\ref{tab:rslt-gmpgap}). The results reveal that CLR can cope well with $\verb'GMP'$ to boost performance while it is not all the case for $\verb'GAP'$ (C3S+GxP+CLR vs. C3S+GxP in Fig~\ref{fig:gmpresults}-\ref{fig:gapresults}). However, when combined with the merged FP maps, which uses $\verb'GAP'$, the behaviour is changed as the results showed in the bottom group of Tab.~\ref{tab:rslt-gmpgap}. 

% This is because $\verb'GAP'$ is apt for objects with repeated structures while $\verb'GMP'$ is suitable for objects with fine-and-rich texture. Thus birds need $\verb'GMP'$ to extract local features for a better performance. 

%$\verb'GAP'$ in  when objects possess  with very similar structures, \ie birds, robust feature extracted from local region is critical to correctly recognition. 

%they show very different behavior --- the performance of C3S+GMP+CLR is better than that of C3S+GMP+FP (Fig.~\ref{fig:gmpresults}), while this phenomenon is reversed for C3S+GAP+CLR and C3S+GAP+FP (Fig.~\ref{fig:gapresults}).  Finally, the performance boosts in general when combining FP with CLR (C3S+GxP+FP+CLR), no matter what pooling method is used.
%
%The last two rows of Tab.~\ref{tab:rslt-gmpgap} compares the performance of our CX-CNN with $\verb'GMP'$ or $\verb'GAP'$. 

\subsection{Comparison with State-of-the-Art}
\label{sec:rslt}

\begin{table}
\small
\begin{center}
%\begin{tabularx}{\linewidth}{|c|c|c|c|c|}
\begin{tabular}{@{}@{\extracolsep{\fill}}|c|c|c|c|@{}}
\hline
Approach					&1-Stage 	& Sep. Init. 	& Accuracy\\
\hline\hline
AlexNet-fc6~\cite{nabirds15Perona} &$\mathtt{\surd}$ &$\mathtt{\times}$ &35.0\\
PN-CNN~\cite{nabirds15Perona} &$\mathtt{\times}$ &$\mathtt{\times}$ &74.0\\
MaxEnt-CNN~\cite{maxent@nips} &\multirow{2}{*}{\centering $\mathtt{\surd}$} 	&\multirow{2}{*}{\centering $\mathtt{\times}$} 	&\multirow{2}{*}{\centering 69.2}\\
(ResNet-50)& & &\\
SENet-50~\cite{senet17cvpr} &$\mathtt{\surd}$ &$\mathtt{\times}$ &82.1\\
ResNet-50~\cite{resnet16kaiming} &$\mathtt{\surd}$ &$\mathtt{\times}$ &82.2\\
MaxEnt-CNN~\cite{maxent@nips} &\multirow{2}{*}{\centering $\mathtt{\surd}$} 		&\multirow{2}{*}{\centering $\mathtt{\times}$} 						&\multirow{2}{*}{\centering 83.0}\\
(DenseNet-161)& & &\\
\hline
Cross-X (SENet) 			&$\mathtt{\surd}$ 	&$\mathtt{\times}$ 	&\textcolor{blue}{$\mathbf{86.4}$}\\
Cross-X (ResNet) 			&$\mathtt{\surd}$ 	&$\mathtt{\times}$ 	&$\mathbf{86.2}$\\
\hline
\end{tabular}
\end{center}
\caption{Performance on NABirds. The result of PN-CNN is implemented with part annotations based on AlexNet. 1-Stage means the network is trained end-to-end after initialization. Sep. Init. denotes separate initialization.}
\label{tab:rslt-nabirds}
\end{table}
% \textbf{Results on NABirds:} NABirds is a recently released dataset which is an order of magnitude larger than the other datasets and includes many more fine-grained categories. 
\textbf{Results on NABirds:} Most previous methods do not report results on this dataset because of the computational complexity of the multi-crop, multi-scale, and multi-stage optimization. Due to the simplicity of our approach, it scales well to big datasets. Tab.~\ref{tab:rslt-nabirds} compares results from methods that are all optimized on single-crop inputs. Our re-implementation of SENet/ResNet-50 is better than the more sophisticated posed-normalized PN-CNN~\cite{pncnn@perona} and the maximum entropy regularized MaxEnt-ResNet-50. The MaxEnt-CNN improves its performance to $83.0\%$ by employing the DenseNet-161 architecture~\cite{densnet@cvpr}. This shows the benefits brought by more advanced network architectures. However, our Cross-X learning can further outperform it by $3.2\%$ with a relatively simple ResNet-50 backbone, which signifies the effectiveness of our approach.

\begin{table}[t]
\small
\begin{center}
%\begin{tabularx}{\linewidth}{|c|c|c|c|c|}
\begin{tabular}{@{}@{\extracolsep{\fill}}|c|c|c|c|@{}}
\hline
Approach					&1-Stage 	& Sep. Init. 	& Accuracy\\
\hline\hline
FCAN~\cite{fcan@lin} 		&$\mathtt{\surd}$ 	&$\mathtt{\times}$	&84.3\\
RA-CNN~\cite{racnn@mei}		&$\mathtt{\times}$	&$\mathtt{\surd}$	&85.3\\
DT-RAM~\cite{ram@arxiv} 	&$\mathtt{\times}$ 	&$\mathtt{\times}$	&86.0\\
MA-CNN~\cite{macnn@mei} 	&$\mathtt{\surd}$ 	&$\mathtt{\surd}$ 	&86.5\\
NTS-Net~\cite{ntscnn@eccv} 	&$\mathtt{\surd}$ 	&$\mathtt{\times}$	&87.5\\
\hline
MaxEnt-CNN~\cite{maxent@nips} &$\mathtt{\surd}$ &$\mathtt{\times}$	&80.4\\
SENet-50~\cite{senet17cvpr} &$\mathtt{\surd}$ &$\mathtt{\times}$ &83.0\\
ResNet-50~\cite{resnet16kaiming} &$\mathtt{\surd}$ &$\mathtt{\times}$ &84.5\\
Kernel-Pooling~\cite{kp@cvpr}&$\mathtt{\surd}$	&$\mathtt{\surd}$	&84.7\\
MAMC-CNN~\cite{mamc18eccv} 	&$\mathtt{\surd}$	&$\mathtt{\times}$	&86.2\\
DFB-CNN~\cite{dfbnet18larry}&$\mathtt{\surd}$	&$\mathtt{\surd}$ &87.4\\
\hline
%Ours (avg pool) 			&$\mathtt{\surd}$ 	&$\mathtt{\times}$	&85.5\\
Cross-X (SENet)			&$\mathtt{\surd}$ 	&$\mathtt{\times}$ 	&$\mathbf{87.5}$\\
Cross-X (ResNet)			&$\mathtt{\surd}$ 	&$\mathtt{\times}$ 	&\textcolor{blue}{$\mathbf{87.7}$}\\
\hline
\end{tabular}
\end{center}
\caption{Performance on CUB-Birds. RA-CNN and MA-CNN are based on VGGNet. Multi-crop operations are employed in the first group while not in others.}
\label{tab:rslt-cubbirds}
\end{table}
\textbf{Results on CUB-Birds:} The classification results for CUB birds are presented in Tab.~\ref{tab:rslt-cubbirds}. Compared to previous methods, our approach achieves the state-of-the-art performance in a much easier experimental setting, in which only one feedforward operation on a single scale input is needed, without any specialized initialization. Notice that DFB-CNN~\cite{dfbnet18larry} needs a separate layer initialization to prevent the model learning from degeneration and NTS-Net~\cite{ntscnn@eccv} conducts feature combinations from multiple crops. MA-CNN~\cite{macnn@mei} obtains comparable results based on VGGNet with part localization pretraining and multi-crop inputs. MaxEnt-CNN~\cite{maxent@nips} can achieve $86.5\%$ when implemented with DenseNet-161, MAMC-CNN~\cite{mamc18eccv} improves to $86.5\%$ when using ResNet-101 and Kernel-Pooling reaches $86.2\%$ when combined with VGGNet as reported in their work; however, still clearly lower than ours.

\begin{table}
\small
\begin{center}
%\begin{tabularx}{\linewidth}{|c|c|c|c|c|}
\begin{tabular}{@{}@{\extracolsep{\fill}}|c|c|c|c|@{}}
\hline
Approach					&1-Stage 	& Sep. Init. 	& Accuracy\\
\hline\hline
RA-CNN~\cite{racnn@mei}		&$\mathtt{\times}$	&$\mathtt{\surd}$	&92.5\\
MA-CNN~\cite{macnn@mei} 	&$\mathtt{\surd}$ 	&$\mathtt{\surd}$ 	&92.8\\
FCAN~\cite{fcan@lin} 		&$\mathtt{\surd}$ 	&$\mathtt{\times}$	&93.1\\
DT-RAM~\cite{ram@arxiv} 	&$\mathtt{\times}$ 	&$\mathtt{\times}$	&93.1\\
NTS-Net~\cite{ntscnn@eccv} 	&$\mathtt{\surd}$ 	&$\mathtt{\times}$ 	&93.9\\
\hline
SENet-50~\cite{senet17cvpr} &$\mathtt{\surd}$ &$\mathtt{\times}$ &91.6\\
Kernel-Pooling~\cite{kp@cvpr}&$\mathtt{\surd}$	&$\mathtt{\surd}$	&92.4\\
ResNet-50~\cite{resnet16kaiming} &$\mathtt{\surd}$ &$\mathtt{\times}$ &92.9\\
MAMC-CNN~\cite{mamc18eccv} 	&$\mathtt{\surd}$	&$\mathtt{\times}$	&93.0\\
DFB-CNN~\cite{dfbnet18larry}&$\mathtt{\surd}$	&$\mathtt{\surd}$ 	&93.8\\
MaxEnt-CNN~\cite{maxent@nips} &$\mathtt{\surd}$ &$\mathtt{\times}$	&93.9\\
\hline
%Ours (avg pool) 			&$\mathtt{\surd}$ 	&$\mathtt{\times}$	&85.5\\
Cross-X (SENet)			&$\mathtt{\surd}$ 	&$\mathtt{\times}$ 	&$\mathbf{94.5}$\\
Cross-X (ResNet)			&$\mathtt{\surd}$ 	&$\mathtt{\times}$ 	&\textcolor{blue}{$\mathbf{94.6}$}\\
\hline
\end{tabular}
\end{center}
\caption{Performance on Stanford Cars. Kernel-Pooling, RA-CNN, and MA-CNN are based on VGGNet. Multi-crop training and testing are employed in the first group.}
\label{tab:rslt-stcars}
\end{table}

\textbf{Results on Stanford Cars:} Tab.~\ref{tab:rslt-stcars} shows the results on Stanford Cars. Our Cross-X learning also achieves the state-of-the-art performance on this dataset, even though DBF-CNN~\cite{dfbnet18larry} and NTS-Net~\cite{ntscnn@eccv} employ separate layer initialization and multi-scale crops, respectively. Kernel-Pooling attains a better result when coupled with VGGNet compared to that of ResNet-50 ($91.9\%$), thus we report the VGGNet-driven results in Tab.~\ref{tab:rslt-stcars}. Compared to MAMC-CNN~\cite{mamc18eccv}, which learns multiple feature maps by embedding the OSME block in a metric learning framework, our Cross-X learning outperforms it by $1.6\%$. The improvement indicates the effectiveness of our proposal to learn semantic part features by exploring the correlations between excitation modules and to extract robust features by bridging the relationship between features in different layers.

\begin{table}
\small
\begin{center}
%\begin{tabularx}{\linewidth}{|c|c|c|c|c|}
\begin{tabular}{@{}@{\extracolsep{\fill}}|c|c|c|c|@{}}
\hline
Approach					&1-Stage 	& Sep. Init. 	& Accuracy\\
\hline\hline
RA-CNN~\cite{racnn@mei}			&$\mathtt{\times}$	&$\mathtt{\surd}$	&87.3\\
FCAN~\cite{fcan@lin} 		&$\mathtt{\surd}$ 	&$\mathtt{\times}$	&$\mathbf{88.9}$\\
\hline
MaxEnt-CNN~\cite{maxent@nips}	&$\mathtt{\surd}$ &$\mathtt{\times}$	&73.6\\
MAMC-CNN~\cite{mamc18eccv} 		&$\mathtt{\surd}$	&$\mathtt{\times}$	&84.8\\
SENet-50~\cite{senet17cvpr} &$\mathtt{\surd}$ &$\mathtt{\times}$ 	&87.1\\
ResNet-50~\cite{resnet16kaiming} &$\mathtt{\surd}$ &$\mathtt{\times}$ 	&88.1\\
\hline
%Ours (avg pool) 			&$\mathtt{\surd}$ 	&$\mathtt{\times}$	&85.5\\
Cross-X (SENet)			&$\mathtt{\surd}$ 	&$\mathtt{\times}$ 	&88.2\\
Cross-X (ResNet)			&$\mathtt{\surd}$ 	&$\mathtt{\times}$ 	&\textcolor{blue}{$\mathbf{88.9}$}\\
\hline
\end{tabular}
\end{center}
\caption{Performance on Stanford Dogs. The first group uses multi-crop operations while the others are not.}
\label{tab:rslt-stdogs}
\end{table}
\textbf{Results on Stanford Dogs:} Classification results are presented in Tab.~\ref{tab:rslt-stdogs}. Surprisingly, the performance of our re-implementation of SENet/ResNet-50 surpasses many previous methods. Even though they can improve their performance by employing more advanced architectures, \eg, MAMC-CNN~\cite{mamc18eccv} with ResNet-101 ($85.2\%$) and MaxEnt-CNN~\cite{maxent@nips} with DenseNet-161 ($83.6\%$) as reported in their papers, still falling behind us. However, our Cross-X learning can beat ResNet-50 a bit and achieve the state-of-the-art performance by combining with SENet-50 and ResNet-50, respectively. FCAN~\cite{fcan@lin} also achieves the best performance, but it is more complicated than our approach and needs multi-scale multi-crops for model training and testing. In contrast, Cross-X learning is simple and effective.

%We observe the best performance with $P=3$ of our approach on this dataset and present the comparison in Tab~\ref{tab:rslt-stdogs}. Our Cross-X learning achieves a comparable result to the state-of-the-art performance. Comparing to FCAN~\cite{fcan@lin}, which achieves the best performance by using multi-scale multi-crop inputs, our approach is simple and only needs one single feedforward computation on a single scale. 
%MaxEnt-CNN~\cite{maxent@nips} and MAMC-CNN~\cite{mamc18eccv} respectively obtain $83.6\%$ and $85.2\%$ when coupled with DenseNet-161 and ResNet-101 as reported in their papers. Surprisingly, our reimplementation of SE-ResNet-50 obtains a comparable result to that of MaxEnt-CNN coupled with DenseNet-161.
%However, our Cross-X learning with ResNet-50 backbone further surpasses them in performance by a relative $4.6\%$ and $3.0\%$.

\begin{table}[t]
\small
\begin{center}
%\begin{tabularx}{\linewidth}{|c|c|c|c|c|}
\begin{tabular}{@{}@{\extracolsep{\fill}}|c|c|c|c|@{}}
\hline
Approach					&1-Stage 	& Sep. Init. 	& Accuracy\\
\hline\hline
MA-CNN~\cite{macnn@mei} 		&$\mathtt{\surd}$ 	&$\mathtt{\surd}$ 	&89.9\\
NTS-Net~\cite{ntscnn@eccv} 		&$\mathtt{\surd}$ 	&$\mathtt{\times}$ 	&91.4\\
\hline
Kernel-Pooling~\cite{kp@cvpr}	&$\mathtt{\surd}$	&$\mathtt{\surd}$	&86.9\\
MaxEnt-CNN~\cite{maxent@nips}	&$\mathtt{\surd}$ 	&$\mathtt{\times}$	&89.8\\
ResNet-50~\cite{resnet16kaiming} &$\mathtt{\surd}$ 	&$\mathtt{\times}$ 	&90.3\\
SENet-50~\cite{senet17cvpr} &$\mathtt{\surd}$ 	&$\mathtt{\times}$ 	&90.6\\
DFB-CNN~\cite{dfbnet18larry}	&$\mathtt{\surd}$	&$\mathtt{\surd}$ 	&92.0\\
\hline
Cross-X (SENet)			&$\mathtt{\surd}$ 	&$\mathtt{\times}$ 	&\textcolor{blue}{$\mathbf{92.7}$}\\
Cross-X (ResNet)			&$\mathtt{\surd}$ 	&$\mathtt{\times}$ 	&$\mathbf{92.6}$\\
\hline
\end{tabular}
\end{center}
\caption{Performance on FGVC-Aircraft. The first group uses multi-crop operations. Kernel-Pooling and DFB-CNN are based on VGGNet. MaxEnt-CNN is implemented with DenseNet-161.}
\label{tab:rslt-vggaircraft}
\end{table}
\textbf{Results on FGVC-Aircraft:} Tab.~\ref{tab:rslt-vggaircraft} reports the average class-prediction accuracy. 
%This dataset is with the fact that the structure of the aircraft changes with their design, varying across subcategories. Nevertheless, o
Our approach obtains the best result among methods reporting results on this dataset, even compared to those based on more advanced network architectures. 
As the main difference of the categories in this dataset results from the changes of aircraft structures, this result implies that our Cross-X learning is applicable to classification problems with relatively large inter-class structural variation. Notice that the performance of Kernel-Pooling~\cite{kp@cvpr}, MaxEnt~\cite{maxent@nips} and DFB-CNN~\cite{dfbnet18larry} methods drop to $83.9\%$, $85.7\%$ and $91.7\%$ respectively when supported by ResNet-50 instead of VGGNet.

\subsection{Visualization}
\label{sec:vis}
Fig.~\ref{fig:acm} displays the resized activation maps~\cite{cam@torralba} of 6 images from 3 datasets (see the supplementary material for more displays). Activation maps from the same layer complement each other --- they concentrate on different regions of the objects. In addition, we find the activations in corresponding columns of (b)$\sim$(d) cover the same object parts at different scales. %even though we do not constrain the corresponding relationship between OSMEs in different stages in our approach. 
Compared to the activation maps (c) $\mathbf{U}^L$, the highly-activated area in (b) $\mathbf{U}^{L-1}$ and (d) $\mathbf{U}^G$ have respectively a relatively small scale and a highlighted center. The activation maps of $\mathbf{U}^G$ can further be seen as the enhanced activation maps of $\mathbf{U}^L$ from that of $\mathbf{U}^{L-1}$, \eg head of birds, wings of planes.
This is consistent with the design of the fine spatial-resolution and rich high-level semantic feature in FPN~\cite{fpn17kaiming}. The difference caused by employing GMP or GAP on $\mathbf{U}^{L-1}$ can also be observed in (b) where GMP leads to consistent activation in a single region (the first two rows) while GAP results in scattered activations in multiple regions (the last 4 rows). 
We further present the combined activation maps in (e) to demonstrate the refined final maps taken as input in our approach for classification. %Last but not the least, the difference caused by employing $\verb'GMP'$ or $\verb'GAP'$ in $\mathbf{U}_p^{L-1}$ can be observed in (b) where $\verb'GMP'$ leads to single position activations (the first two rows) while $\verb'GAP'$ results in scattered activations (the last 4 rows). 

\section{Conclusion}
\label{sec:clus}
We proposed Cross-X learning to learn robust fine-grained feature by exploiting relationships between features from different images and different network layers. Our approach leverages the fact that
features for the same semantic parts, although coming from different images with different class labels, should be more correlated than those for different semantic parts. Experiments evaluated on five benchmark datasets, ranging from 100 to 555 categories, validate the effectiveness of our approach. Ablation studies further demonstrate the role of every component of Cross-X.%We evaluated our approach on five fine-grained benchmark datasets ranging from 100 to 555 categories. Experimental results validate the effectiveness of our approach. Ablation studies further explain the role of every component of our approach. %In the future, we plan to extend our learning strategy to more network architectures to further verify its effectiveness.

\noindent\textbf{Acknowledgement}
This work was supported by NSFC (No.61702197), NSFGD (No.2017A030310261), and Facebook AI.

{\small
\bibliographystyle{ieee_fullname}
\bibliography{main-arxiv}
}

\appendix
\section{Hyper-parameters}
\label{sec:hyperparams}

\begin{table}[th]
\begin{center}
%\begin{tabularx}{\linewidth}{|c|c|c|c|c|c|}
\begin{tabular}{@{}@{\extracolsep{\fill}}|c|c|c|c|c|c|@{}}
\hline
params			&NABirds &CUB-Birds &Cars	& Dogs	& Aircraft\\
\hline\hline
$\#P$ 		    &2		& 2 		&2		& 3 	&2\\
\hline
$\gamma_1$ 		&0.1 	&1		    & 1 	&1      &0.5\\
\hline
$\gamma_2$ 	    &0.25	& 0.25 		&0.25	&0.5 	&0.1\\
\hline
$\gamma_3$ 		&0.5	& 1 		& 1 	&1		&0.1\\
\hline
$\lambda_1$ 	&1		& 1 		& 1 	&1		&1\\
\hline
$\lambda_2$ 	&1		& 1 		& 1 	&1		&1\\
\hline
\end{tabular}
% \begin{tablenotes}
% $\#$parts is an alternative to $\#$excitations in this paper.
% \end{tablenotes}
\end{center}
\caption{Hyper-parameters of Cross-X with SENet-50 backbone.}
\label{tab:hyperparams-senet}
\end{table}

\begin{table}[th]
\begin{center}
%\begin{tabularx}{\linewidth}{|c|c|c|c|c|c|}
\begin{tabular}{@{}@{\extracolsep{\fill}}|c|c|c|c|c|c|@{}}
\hline
params	&NABirds 	& CUB-Birds 	& Cars 	& Dogs & Aircraft \\
\hline\hline
$\#P$ 		    & 2 	& 2 	& 2 	&2 		& 2 \\
\hline
$\gamma_1$ 		& 0.5 	& 0.5 	&1		&0.01	&0.5\\
\hline
$\gamma_2$ 	    & 0.25 	& 0.25 	&0.25 	&0.01	&0.1\\
\hline
$\gamma_3$ 		& 0.5 	& 0.5 	&1		&1	    &0.5\\
\hline
$\lambda_1$ 	& 1 	& 1 	&1		&1	    &1\\
\hline
$\lambda_2$ 	& 1 	& 1 	&1		&1	    &1\\
\hline
\end{tabular}
% \begin{tablenotes}
% $\#$parts is an alternative to $\#$excitations in this paper.
% \end{tablenotes}
\end{center}
\caption{Hyper-parameters of Cross-X with ResNet-50 backbone.}
\label{tab:hyperparams-resnet}
\end{table}

Cross-X learning involves 6 hyper-parameters --- $P$, $\gamma_1, \gamma_2, \gamma_3, \lambda_1, \lambda_2$. Among them, $P$ is the number of excitations employed in OSME; $\gamma_1, \gamma_2$ and $\gamma_3$ are used to balance the effects of $C^3S$ for different layers (see Eq.~(\textcolor{red}{10})); $\lambda_1$ and $\lambda_2$ are adopted to adjust the effects of $CL$ (see Eq.~(\textcolor{red}{11})). These hyper-parameters are determined by evaluating models on hold-out validation datasets. The hyper-parameters for various datasets are presented in Tab.~\ref{tab:hyperparams-senet} and~\ref{tab:hyperparams-resnet}.

\section{Training details}
All experiments in ablation studies are implemented on the SENet backbone (Sec.~\textcolor{red}{4.3}). On all datasets, images are resized to $448\times 448$ for training and testing. OSMEs with 2 excitations are used in all experiments on all datasets except that on Stanford Dogs where 3 excitations are employed. 

To present the state-of-the-art performance (Sec.~\textcolor{red}{4.4}), images on CUB-Birds, NABirds, and VGG-Aircraft are first resized to $600\times 600$, and then image patches of size $448\times 448$ from random cropping and center cropping are used for training and testing, respectively. We did not observe any advantage of this trick on Stanford Cars and Stanford Dogs, thus default operations as that implemented in the ablation study are employed on these two datasets. The re-implementation of SENet-50 and ResNet-50 in Sec.~\textcolor{red}{4.4} also obeys these operation rules.

\section{Visualization}
\label{sec:visual}
\begin{figure*}[h]
\begin{center}
\includegraphics[width=\linewidth]{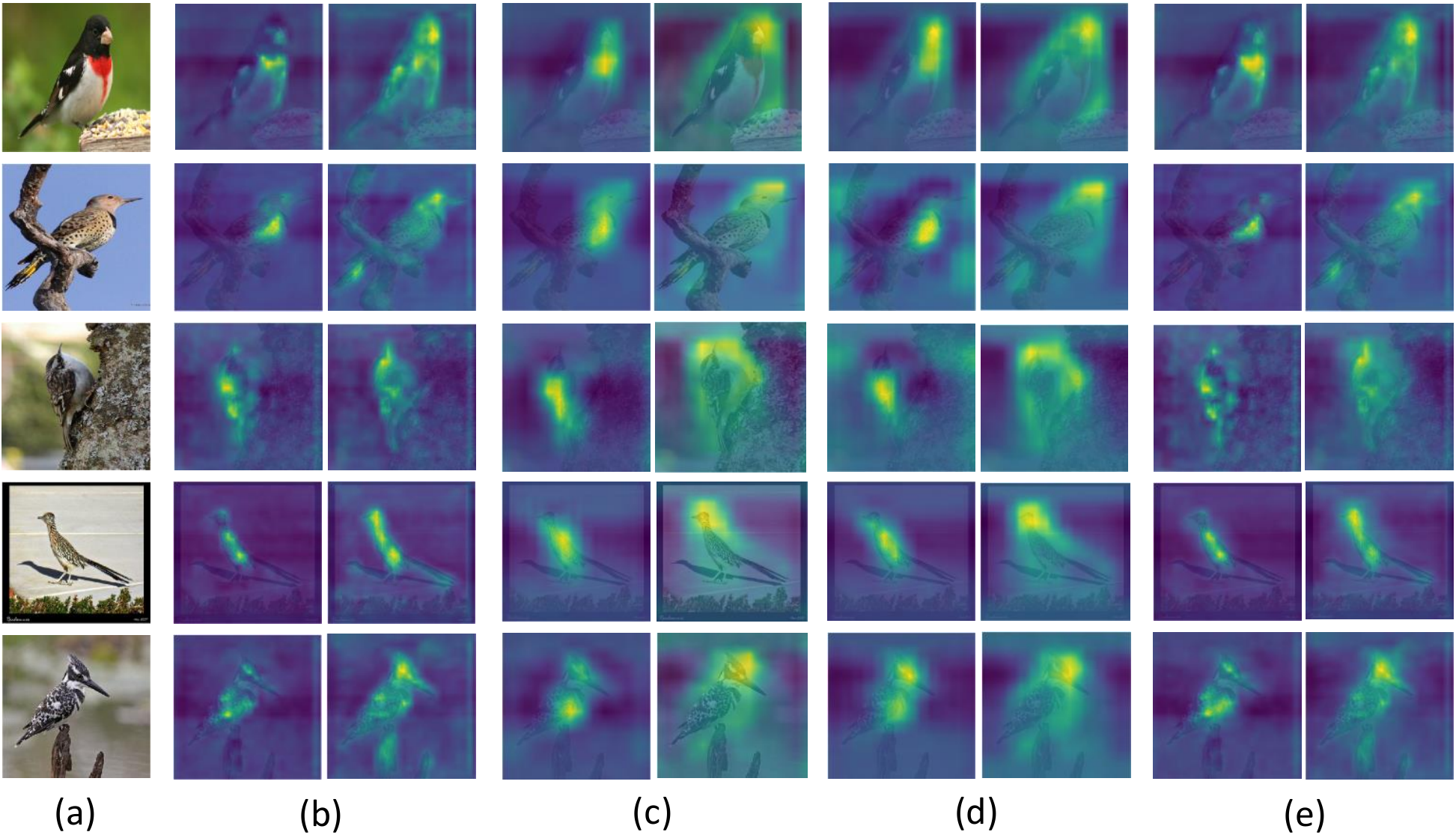}
\end{center}
\caption{Superimposed display of activation maps (b) $\mathbf{U}_p^{L-1}$, (c) $\mathbf{U}_p^L$ and (d) $\mathbf{U}_p^G$  for images from CUB-Birds. The first column (a) shows original images and the last two columns (e) are combined activation maps from corresponding columns of $\mathbf{U}_p^{L-1}$, $\mathbf{U}_p^L$ and $\mathbf{U}_p^G$ . Each of (b)$\sim$(e) shows the activations of two excitation modules in corresponding layers. Best viewed in color.} 
\label{fig:acm-birds}
\end{figure*}

\begin{figure*}[h]
\begin{center}
\includegraphics[width=\linewidth]{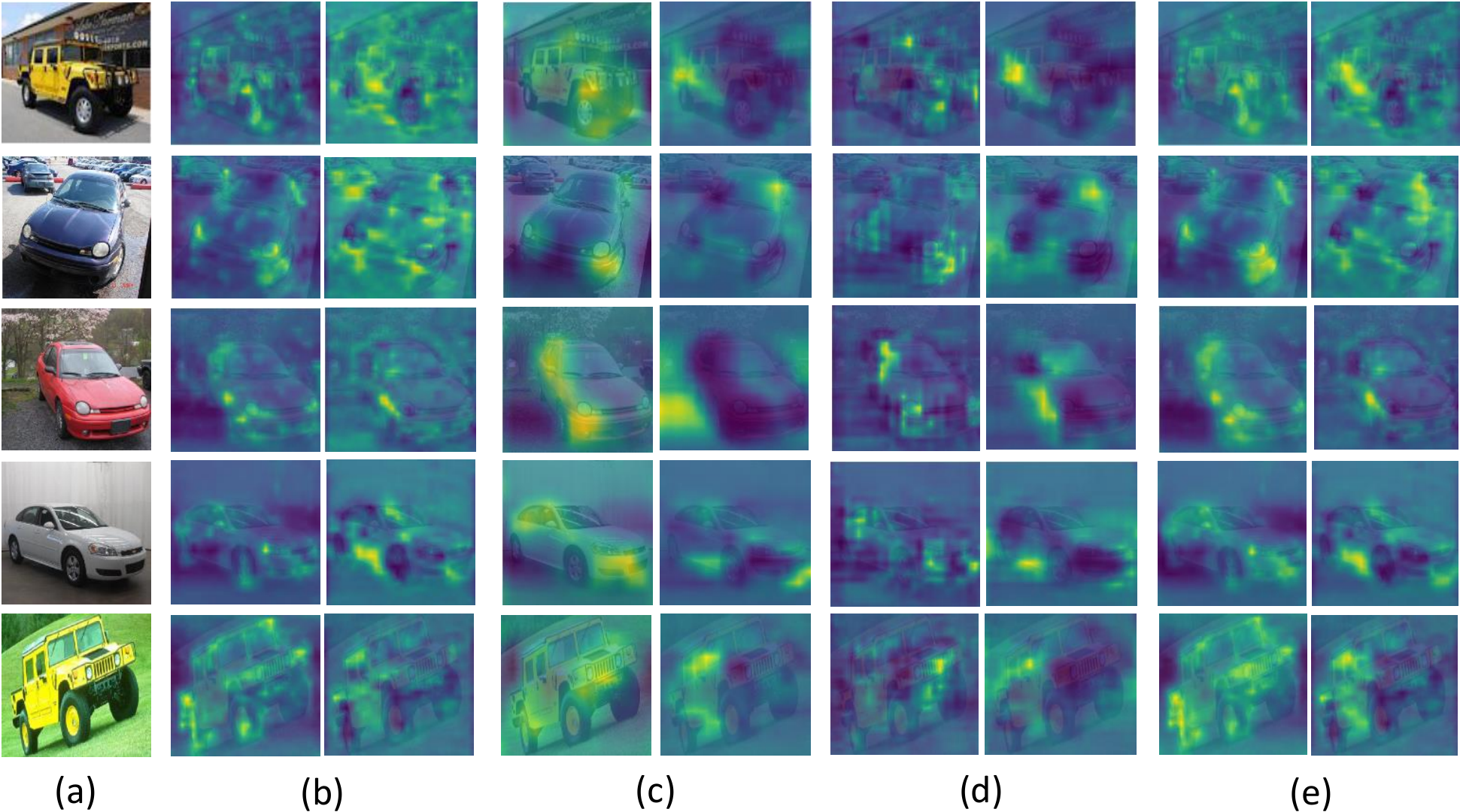}
\end{center}
\caption{Superimposed display of activation maps (b) $\mathbf{U}_p^{L-1}$, (c) $\mathbf{U}_p^L$ and (d) $\mathbf{U}_p^G$  for images from Stanford Cars. The first column (a) shows original images and the last two columns (e) are combined activation maps from corresponding columns of $\mathbf{U}_p^{L-1}$, $\mathbf{U}_p^L$ and $\mathbf{U}_p^G$ . Each of (b)$\sim$(e) shows the activations of two excitation modules in corresponding layers. Best viewed in color.} 
\label{fig:acm-cars}
\end{figure*}

\begin{figure*}[t]
\begin{center}
\includegraphics[width=\linewidth]{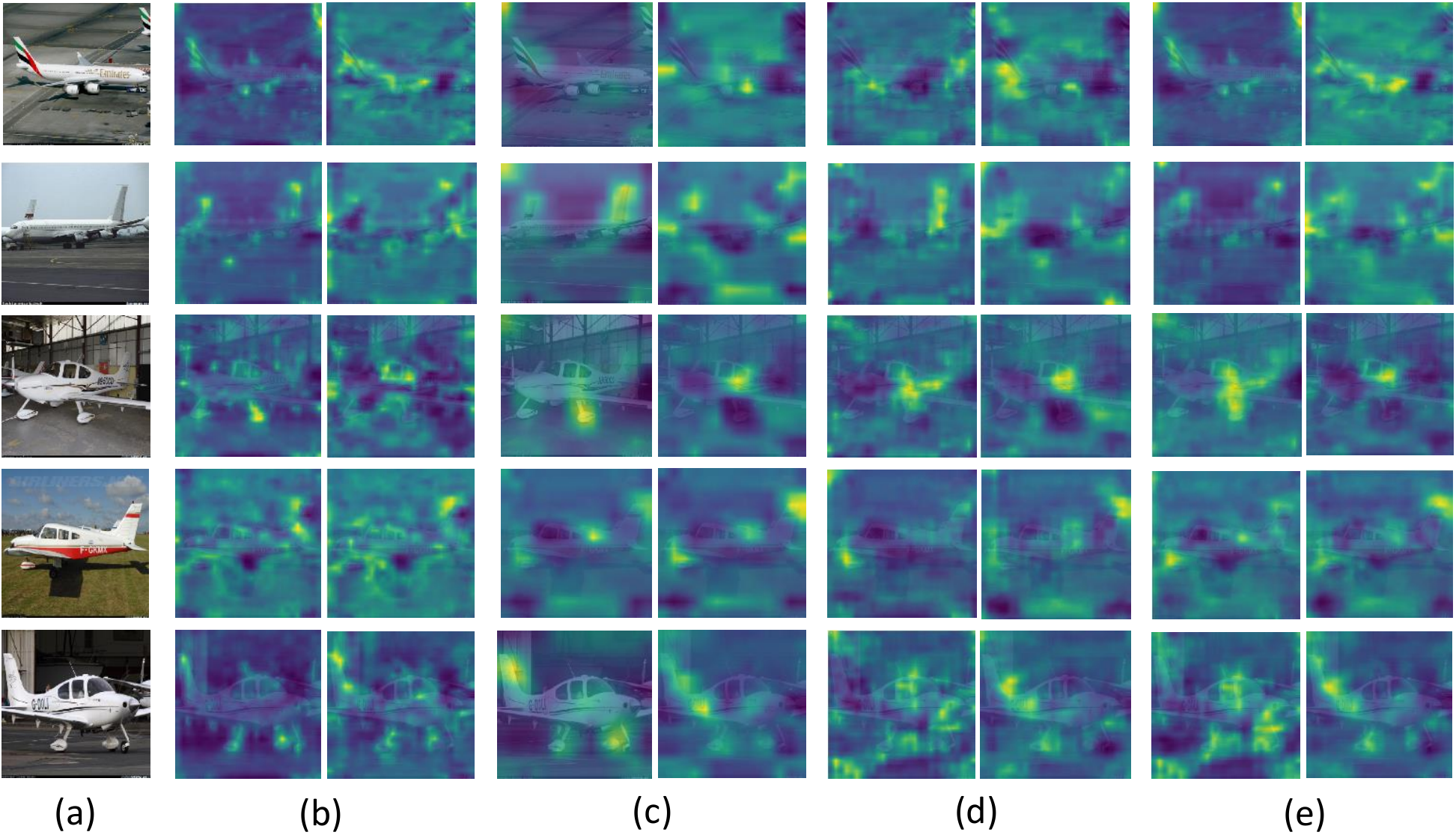}
\end{center}
\caption{Superimposed display of activation maps (b) $\mathbf{U}_p^{L-1}$, (c) $\mathbf{U}_p^L$ and (d) $\mathbf{U}_p^G$  for images from FGVC-Aircraft. The first column (a) shows original images and the last two columns (e) are combined activation maps from corresponding columns of $\mathbf{U}_p^{L-1}$, $\mathbf{U}_p^L$ and $\mathbf{U}_p^G$ . Each of (b)$\sim$(e) shows the activations of two excitation modules in corresponding layers. Best viewed in color.} 
\label{fig:acm-aircraft}
\end{figure*}

\begin{figure*}[t]
\begin{center}
\includegraphics[width=\linewidth]{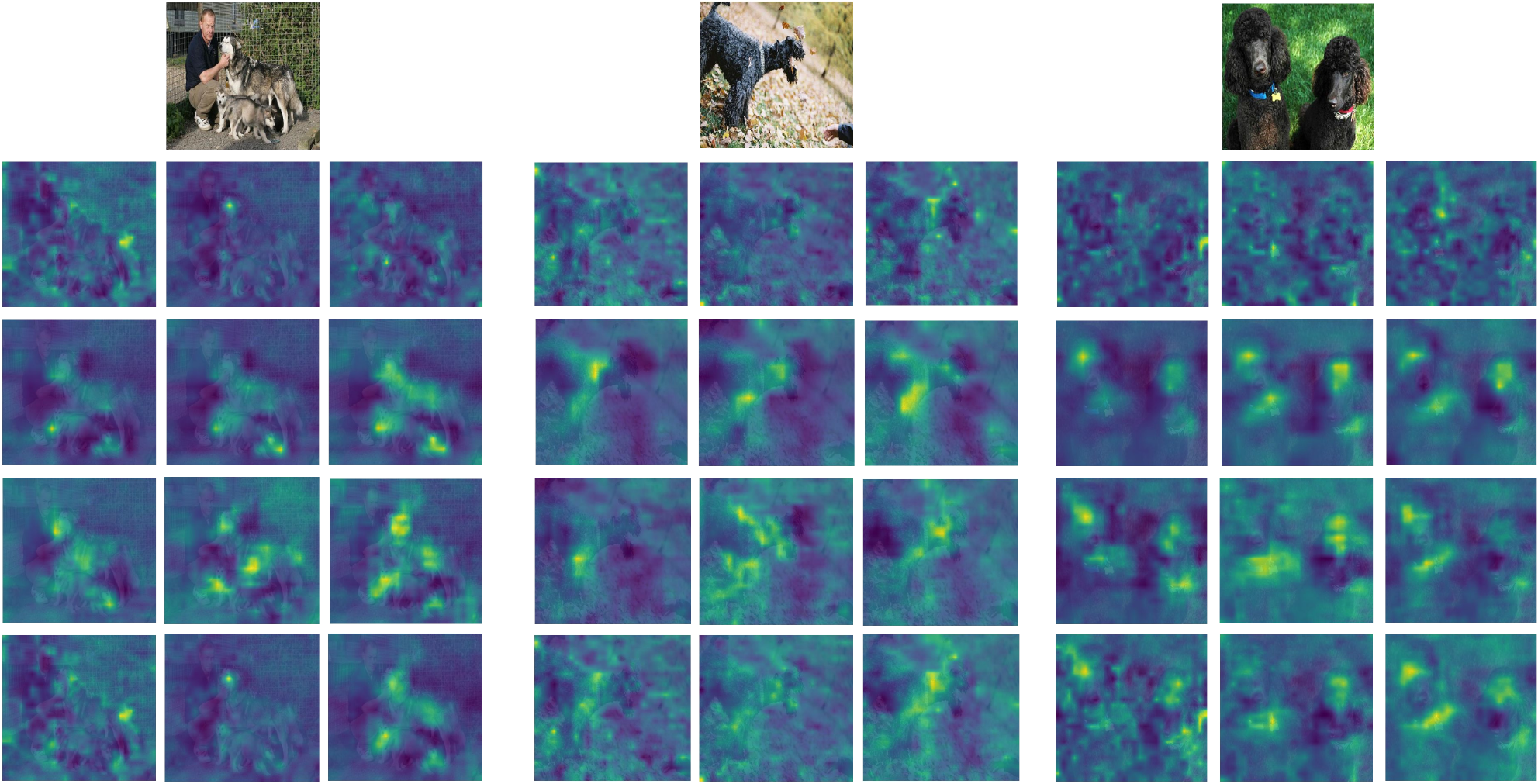}
\end{center}
\caption{Superimposed display of activation maps $\mathbf{U}_p^{L-1}$ (2nd row),  $\mathbf{U}_p^L$ (3rd row) and $\mathbf{U}_p^G$ (4th row) for images from Stanford Dogs. The first row shows original images and the last row are combined activation maps from corresponding rows of $\mathbf{U}_p^{L-1}$, $\mathbf{U}_p^L$ and $\mathbf{U}_p^G$. Each row shows the activations of three excitation modules in corresponding layers. Best viewed in color.} 
\label{fig:acm-dogs}
\end{figure*}

We display additional activation maps in this section for images from birds (Fig.~\ref{fig:acm-birds}), cars (Fig.~\ref{fig:acm-cars}), aircraft (Fig.~\ref{fig:acm-aircraft}) and dogs (Fig.~\ref{fig:acm-dogs}). The images shown here are consistent with the analysis presented in Section~\textcolor{red}{4.5} of the paper.

\end{document}